\documentclass[10pt]{article} 
\usepackage{natbib}

\usepackage{amsmath,amsthm,amssymb}
\usepackage{times}
\usepackage{algorithm}
\usepackage{graphicx}
\usepackage{color}       % seems to be necessary
\usepackage{epsfig}
\usepackage{amsfonts}
\usepackage{amssymb}
\usepackage{amsthm}

\newcommand{\q}{\qquad}
\newcommand{\rem}[1]{}

\newtheorem{proposition}{Proposition}
\newtheorem{lemma}{Lemma}

\newtheorem{definition}{Definition}

\newcommand{\HCRF}{HSCRF}
\newcommand{\DCRF}{DCRF}
\newcommand{\vertex}{\mathcal{V}}
\newcommand{\edges}{\mathcal{E}}

\newcommand{\G}{\mathcal{G}}

\newcommand{\f}{\mathbf{f}}
\newcommand{\F}{\mathbf{F}}
\newcommand{\w}{\mathbf{w}}

\newcommand{\BigO}{\mathcal{O}}

\begin{document}

\title{Hierarchical Semi-Markov Conditional Random Fields for Recursive Sequential Data}
%\date{}
\author{Tran The Truyen{\small~$^\dagger$}, Dinh Q. Phung{\small~$^\dagger$},
       	Hung H. Bui{\small~$^\ddagger$}\thanks{Hung Bui is supported by the Defense Advanced
					Research Projects Agency (DARPA) under Contract No.
					FA8750-07-D-0185/0004. Any opinions, findings and conclusions
					or recommendations expressed in this material are those of the authors
					and do not necessarily reflect the views of DARPA, or the Air Force Research
					Laboratory (AFRL).} , 
			and Svetha Venkatesh{\small~$^\dagger$}\\
       	{\small~$^\dagger$}Department of Computing, Curtin University of Technology\\
       		GPO Box U1987 Perth, WA 6845, Australia\\
    		\texttt{thetruyen.tran@postgrad.curtin.edu.au}\\
    		\texttt{\{D.Phung,S.Venkatesh\}@curtin.edu.au}\\
       		\\
       	{\small~$^\ddagger$}Artificial Intelligence Center,	SRI International\\
			333 Ravenswood Ave, Menlo Park, CA 94025, USA\\
			\texttt{bui@ai.sri.com}
}

\maketitle

\begin{abstract}%   <- trailing '%' for backward compatibility of .sty file
Inspired by the hierarchical hidden Markov models (HHMM),
we present the \emph{hierarchical semi-Markov conditional random field} ({\HCRF}),
a generalisation of embedded undirected Markov chains to model
complex hierarchical, nested Markov processes.
It is parameterised in a discriminative framework
and has polynomial time algorithms for learning
and inference.
Importantly, we consider partially-supervised learning and propose
algorithms for generalised partially-supervised learning and
constrained inference.
We demonstrate the {\HCRF} in two applications:
(i) recognising human activities of daily living (ADLs)
	from indoor surveillance cameras, and
(ii) noun-phrase chunking.
We show that the {\HCRF} is capable of learning rich
hierarchical models with reasonable accuracy in both fully and
partially observed data cases. 
\end{abstract}

%\begin{keywords}
%  Hierarchical Conditional Random Fields, Partially Supervised Learning,
%  Constrained Inference, Asymmetric Inside/Outside.
%\end{keywords}

\tableofcontents

\section{Introduction}
\label{sec:intro}
% Introduction

% Intro for Hierarchical modelling
Modelling hierarchical aspects in complex stochastic processes is an important
research issue in many application domains. In an hierarchy,
each level is an \emph{abstraction} of lower level details.
Consider, for example, a frequent activity performed by human like `eat-breakfast'
may include a series of more specific activities like 
`enter-kitchen', `go-to-cupboard', `take-cereal', `wash-dishes' and
`leave-kitchen'. Each specific activity can be decomposed into finer details.
Similarly, in natural language processing (NLP) syntax trees are inherently
hierarchical. In a partial parsing task known as
noun-phrase (NP) chunking~\citep{Sang-Buchholz-CoNLL00}, 
there are three semantic levels: the sentence, noun-phrases and part-of-speech (POS) tags.
In this setting, the sentence is a sequence of NPs and non-NPs
and each phrase is a sub-sequence of POS tags. 

A popular approach to deal with hierarchical data is to build a \emph{cascaded} model
where each level is modelled separately, and the output of the lower level
is used as the input of the level right above it
(e.g. see \citep{oliver2004lrl}). For instance,
in NP chunking this approach first builds a POS tagger and then constructs
a chunker that incorporates the output of the tagger. 
This approach is clearly sub-optimal because the 
POS tagger takes no information of the NPs and the chunker is not aware of the
reasoning of the tagger. In contrast, a noun-phrase is often very informative
to infer the POS tags belonging to the phrase. As a result, this layered approach
often suffers from the so-called \emph{cascading error} problem as
the error introduced from the lower layer will propagate to
higher levels. 

A more holistic approach is to build a joint representation of all
the levels. Formally, we are given a data observation $z$ and we need
to model and infer about the joint semantic $x$. The main problem is
to choose an appropriate representation of $x$ so that inference
can be efficient. In this paper, we are interested in a specific class
of hierarchical models that supports both joint modelling
and efficient inference. More specifically,
the models of interest are \emph{recursive} and \emph{sequential}, 
in that each level is a sequence and each
node in a sequence can be decomposed further into a sub-sequence
of finer grain at the lower level. 

There has been substantial investigation of these types of model, especially
in the area of probabilistic context-free grammars (e.g. see \citep[Chapter 11]{ManningSchutze}). 
However, grammars are 
often unbounded in depth and thus difficult to represent by graphical models. 
A more restricted version
known as hierarchical hidden Markov model (HHMM) \citep{fine98hierarchical} offers
clearer representation in that the depth is fixed and the semantic levels
are well defined. Essentially, the HHMM
is a nested hidden Markov network (HMM) in the sense that each state is a sub HMM by itself.

%--- Discriminative arguements ------------
These models share a common property in that they are
\emph{generative}, i.e. they assume that the data observation is generated by
the hierarchical semantics. The generative models try to construct the the joint 
distribution $\Pr(x,z) = \Pr(z|x)\Pr(x)$. 
However, there are some drawbacks associated with this approach. First,
the generative process modelled by $\Pr(z|x)$ is typically unknown and 
complicated. Second, given an observation $z$, we are more often interested
in inferring $\Pr(x|z)$. Since $\Pr(x,z)=\Pr(x|z)\Pr(z)$, 
modelling $\Pr(z)$ may be unnecessary. 

% discriminative - CRF, DCRF, C-PCFG, and why {\HCRF}
An attractive alternative is to model the
distribution $\Pr(x | z)$ directly, avoiding the modelling of $z$. This line of
research has recently attracted much interest, largely
triggered by the introduction of the {\em conditional random field}
(CRF)~\citep{lafferty01conditional}.
The advantages of the CRF is largely attributed to its {\em discriminative} nature that
allows arbitrary and long-range interdependent features.
%Early work in CRFs is limited to flat structures
%for efficient inference. Recent extensions to include hierarchical
%structures include dynamic CRFs (DCRF)~\citep{Sutton-et-alJMLR07},
%and hierarchical CRFs~\citep{Liao-et-al-ISRR05}). However, DCRFs
%are not recursive and not efficient, while hierarchical CRFs are
%not flexible to infer tree structures from data.

In this paper we follow the HMM/HHMM path to
generalise from chain-structured CRFs to nested CRFs. 
As a result, we construct a novel model called  
{\em Hierarchical Semi-Markov Conditional Random Field} ({\HCRF}), which
is  an undirected conditional graphical model of nested Markov chains.
Thus {\HCRF} is the combination of the discriminative nature
of CRFs and the nested modelling of the HHMM.

To be more concrete let us return to the NP chunking example.
The problem can be modelled as a three-level {\HCRF}, 
where the root represents the sentence, the second level the
NP process, and the bottom level the POS process. The root and the two processes 
are conditioned on the sequence of words in the sentence.
Under the discriminative modelling of the {\HCRF}, 
rich contextual information such as starting and ending of the phrase, the phrase length,
and the distribution of words falling inside the phrase
can be effectively encoded. On the other hand,
such encoding is much more difficult for HHMMs.

We then proceed to address important issues.
First, we show how to represent {\HCRF}s using a dynamic graphical model (e.g.
see \citep{Lauritzen96}) which effectively encodes hierarchical and temporal semantics. 
For parameter learning, an efficient algorithm
based on the Asymmetric Inside-Outside of \citep{Bui-et-al04} is introduced.
For inference, we generalise the Viterbi algorithm to decode the semantics
from an observational sequence. 

The common assumptions in discriminative learning and inference are that
the training data in learning is fully labelled, and the test data
during inference is not labelled. We propose to relax these assumptions
in that training labels may only be partially available, and
we term the learning as \em partial-supervision\em. Likewise, 
when some labels are given during inference, the algorithm should automatically
adjust to meet the new constraints.

We demonstrate the effectiveness of {\HCRF}s in two applications:
(i) segmenting and labelling activities of daily living (ADLs) in an indoor
environment and (ii) jointly modeling noun-phrases and part-of-speeches
in shallow parsing.  Our experimental results in the first application
show that the {\HCRF}s are  capable of learning rich, hierarchical activities
with good accuracy and exhibit better performance when
compared to {\DCRF}s and flat-CRFs.
Results for the partially observable case also demonstrate that
significant reduction of training labels still results in models
that perform reasonably well. We also show that observing a small amount of
labels can significantly increase the accuracy during decoding.
In shallow parsing, the {\HCRF}s can achieve higher accuracy
than standard CRF-based techniques and the recent DCRFs.

To summarise, in this paper we claim the following contributions: 
\begin{itemize}
\item Introducing a novel Hierarchical Semi-Markov
		Conditional Random Field ({\HCRF}) to model complex hierarchical and nested
		Markovian processes in a discriminative framework,
\item Developing an efficient generalised Asymmetric Inside-Outside (AIO) algorithm
		for full-supervised learning.
\item Generalising the Viterbi algorithm for decoding the most probable
		semantic labels and structure given an observational sequence.
\item Addressing the problem of partially-supervised 
		learning and constrained inference.
\item Demonstration of the applicability of the {\HCRF}s for
		modeling human activities in the domain of home video surveillance
		and shallow parsing of English.
\end{itemize}

%----------------------
\subsubsection*{Notations and Organisation}
%--
This paper makes use of a number of mathematical notations
which we include in Table~\ref{tab:hcrf-notation} for reference.

\begin{table}[htb]
\begin{center}
\begin{tabular}{rl} 
\hline
\textbf{Notation}			& \textbf{Description}\\\hline\hline
$x^{d:d'}_{i:j}$			& Subset of state variables from level $d$ down to level $d'$ \\
							& \q and starting from time $i$ and ending at time $j$, inclusive.\\
$e^{d:d'}_{i:j}$			& Subset of ending indicators from level $d$ down to level $d'$ \\
							& \q and starting from time $i$ and ending at time $j$, inclusive.\\
$\zeta^{d,s}_{i:j}$			& Set of state variables and ending indicators of a \\
							& \q sub model rooted at $s^d$, level $d$, spanning a sub-string $[i,j]$ \\
$\sigma$					& Contextual clique\\
$i,j,t$						& Time indices \\
$\tau^d$ 					& Set of all ending time indices, e.g. if $i \in \tau^d$ then $e^d_{i} = 1$\\
$r,s,u,v,w$ 				& State \\
$R^{d,s,z}_{i:j}$			& State-persistence potential
							 	of state $s$, level $d$, spanning $[i,j]$ \\
$\pi^{d,s}_{u,i}$			& Initialisation potential of
								state $s$ at level $d$, time $i$
								initialising sub-state $u$\\
$A^{d,s,z}_{u,v,i}$			& Transition at level $d$, time $i$ from state
								$u$ to $v$ under the same parent $s$ \\
$E^{d,s,z}_{u,i}$			& Ending potential of state $z$ at level $d$ and time $i$, and receiving \\
							& \q the return control from the child $u$\\
$\Phi[\zeta,z]$				& The global potential of a particular configuration $\zeta$ \\
$S^d$						& The number of state symbols at level $d$\\
$\Delta^{d,s}_{i:j}$		& The symmetric inside mass for a state $s$ at level $d$, \\
							& \q spanning	a substring $[i,j]$\\
$\hat{\Delta}^{d,s}_{i:j}$	& The full symmetric inside mass for a state $s$ at level $d$, \\
							& \q spanning a substring $[i,j]$\\
$\Lambda^{d,s}_{i:j}$		& The symmetric outside mass for a state $s$ at level $d$, \\
							& \q spanning a substring $[i,j]$\\
$\hat{\Lambda}^{d,s}_{i:j}$	& The full symmetric outside mass for a state $s$ at level $d$, \\
							& \q spanning a substring $[i,j]$\\
$\alpha^{d,s}_{i:j}(u)$		& The asymmetric inside mass for a parent state $s$ at level $d$, starting at $i$  \\
							& \q and having a child-state $u$ which returns control \\
							& \q to parent or transits to new child-state at $j$\\
$\lambda^{d,s}_{i:j}(u)$	& The asymmetric outside mass, as a counterpart of \\
							& \q asymmetric inside mass  $\alpha^{d,s}_{i:j}(u)$ \\
$\psi(.), \varphi(.)$		& Potential functions.\\
\hline
\end{tabular}
\end{center}
\caption{Notations used in this paper.}
\label{tab:hcrf-notation}
\end{table}

The rest of the paper is organised as follows.
Section~\ref{sec:bg} reviews Conditional Random Fields. 
Section~\ref{sec:model} continues with the {\HCRF} model definition and parameterisation. 
Section~\ref{sec:AIO-all} defines building blocks 
required for common inference tasks. These blocks are computed
in Section~\ref{sec:IF} and~\ref{sec:OB}.
Section~\ref{sec:MAP} presents the generalised Viterbi algorithm.
Parameter estimation follows in Section~\ref{sec:learning}.
Learning and inference with partially available labels
are addressed in Section~\ref{sec:partial}. Section~\ref{sec:scaling}
presents a method for numerical scaling to prevent numerical overflow.
Section~\ref{sec:exp} documents experimental results.
Section~\ref{sec:con} concludes the paper.

\section{Related Work}
\label{sec:bg}
%Previous work

%--------------------------
\subsection{Hierarchical Modelling of Stochastic Processes}
%--
Hierarchical modelling of stochastic processes can be largely categorised as either
graphical models extending the flat hidden Markov models (HMM)
(e.g., the layered HMM~\citep{oliver2004lrl}, the abstract HMM~\citep{Bui-et-al02},
hierarchical HMM (HHMM)~\citep{fine98hierarchical,Bui-et-al04}, DBN~\citep{Murphy02}) or
grammar-based models (e.g., PCFG~\citep{pereira92insideoutside}).
These models are all generative. 

% discriminative - CRF, DCRF, C-PCFG, and why HierCRF
Recent development in discriminative, hierarchical structures include extension of the flat CRFs 
(e.g. dynamic CRFs (DCRF)~\citep{Sutton-et-alJMLR07},
hierarchical CRFs~\citep{liao2007epa,Kumar-Hebert-ICCV05})
and conditional learning of the grammars (e.g. see \citep{Miyao-TsujiiHLT02,Clark-Curran-EMNLP03}).
The main problem of the DCRFs is that they are not scalable due to inference intractability.
The hierarchical CRFs, on the other hand, are tractable but assume fixed
tree structures, and therefore are not flexible to adapt to
complex data. For example, in the noun-phrase chunking
problem no prior tree structures are known. Rather, if such a
structure exists, it can only be discovered after
the model has been successfully built and learned.

The conditional probabilistic context-free grammar (C-PCFG) appears
to address both tractability and dynamic structure issues. 
More precisely, in C-PCFGs it takes cubic time in sequence length
to parse a sentence. However, the context-free grammar does not
limit the depth of semantic hierarchy, thus making it unnecessarily
difficult to map many hierarchical problems into its form.
Secondly, it lacks a graphical model representation and thus
does not enjoy the rich set of approximate inference techniques
available in graphical models.

%-- Details of HHMM ---
\subsection{Hierarchical Hidden Markov Models}
\label{sec:bgr-HHMM}
%--
Hierarchical HMMs are generalisations of HMMs \citep{RabinerIEEE-89}
in the way that a state in an HHMM may be a sub-HHMM.
Thus, an HHMM is a nested Markov chain.
In the model temporal evolution,
when a child Markov chain terminates, it returns the control to its parent.
Nothing from the terminated child chain is carried forward.
Thus, the parent state abstracts out everything belonging to it.
Upon receiving the return control
the parent then either transits to a new parent, (given that the grand parent
has not finished), or terminates. 

\begin{figure}[htb]
\begin{center}
\begin{tabular}{c}
\includegraphics[width=0.80\linewidth]{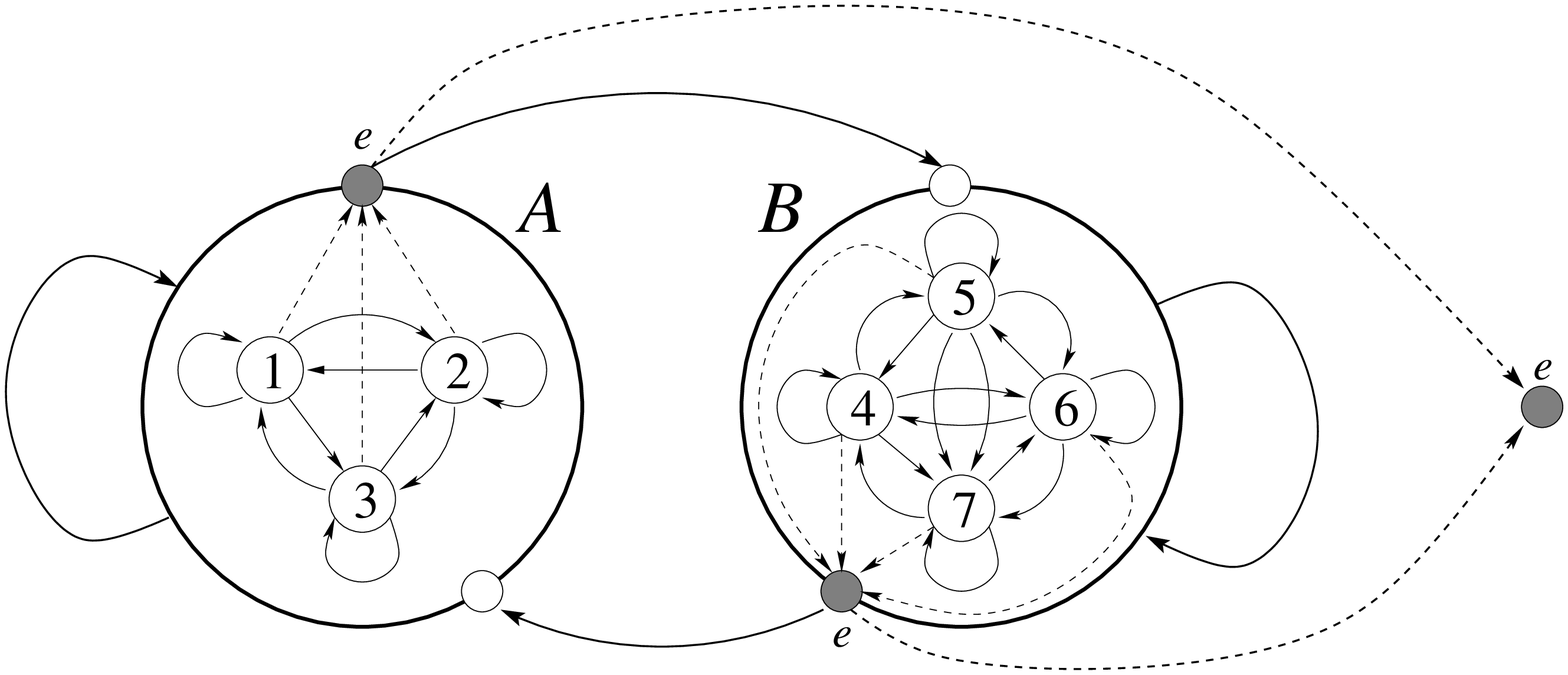} \\
\end{tabular}
\end{center}
\caption{The state transition diagram of an HHMM.}
\label{fig:HHMM-state-transit}
\end{figure}

Figure~\ref{fig:HHMM-state-transit} illustrates the state transition
diagram of a two-level HHMM. At the top level there are two parent states
$\{A,B\}$. The parent $A$ has three children, i.e. $ch(A) = \{1,2,3\}$
and $B$ has four, i.e. $ch(B) = \{4,5,6,7\}$. At the top level the transitions
are between $A$ and $B$, as in a normal directed Markov chain.
Under each parent there are also transitions between child states,
which only depend on the direct parent (either $A$ or $B$).
There are special ending states (represented as shaded nodes in Figure~\ref{fig:HHMM-state-transit})
to signify the termination of the Markov chains. 
At each time step of the child Markov chain, a
child will emit an observational symbol (not shown here).

%-------------
%-------------

\begin{figure}[htb]
\begin{center}
\begin{tabular}{c}
\includegraphics[width=0.60\linewidth]{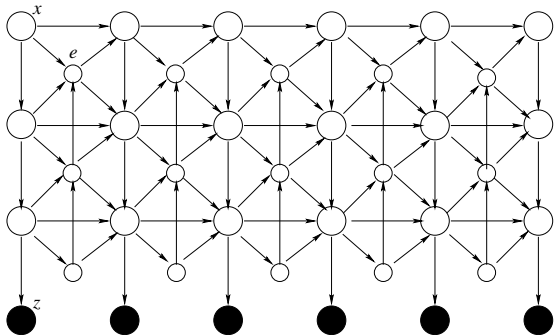} \\
\end{tabular}
\end{center}
\caption{Dynamic Bayesian network representation of HHMMs.}
\label{fig:hhmm-dbn}
\end{figure}

The temporal evolution of the HHMM can be represented
as a dynamic Bayesian network, which was first done in \citep{Murphy-Paskin01}.
Figure~\ref{fig:hhmm-dbn} depicts a DBN structure of 3 levels. 
The bottom level is often referred to as \emph{production level}.
Associated with each state is an ending indicator to signify the termination
of the state. Denote by $x^d_t$ and $e^d_t$ the state and ending indicator
at level $d$ and time $t$, respectively. When $e^d_t = 0$, the state $x^d_t$
continues, i.e. $x^d_t = x^d_{t+1}$. And when $e^d_t = 1$, the state $x^d_t$
transits to a new state, or transits to itself. 
There are hierarchical consistency rules that must be ensured.
Whenever a state persists (i.e. $e^d_t=0$), all of the states above
it must also persist (i.e. $e^{d'}_t=0$ for all $d' < d$).
Similarly, whenever a state ends  (i.e $e^d_t=1$), all of the states
below it must also end (i.e. $e^{d'}_t=1$ for all $d' > d$).

Inference and learning in HHMMs follow the Inside-Outside algorithm
of the probabilistic context-free grammars. Overall, the algorithm has
$\BigO(|S|^3DT^3)$ time complexity where $|S|$ is the maximum size of the state space
at each level, $D$ is the depth of the model and $T$ is the model length.

When representing as a DBN, the whole stack of states $x^{1:D}_t$ can be collapsed
into a `mega-state' of a big HMM, and therefore inference
can be carried out in $\BigO(|S|^{2D}T)$ time. 
This is efficient for a shallow model (i.e. $D$ is small), but problematic for
a deep model (i.e. $D$ is large).

%-------------
%-------------

%----------------------------
\subsection{Conditional Random Fields}
\label{sec:bgr-crf}
%---

Denote by $\G=(\vertex,\edges)$ the graph where $\vertex$ is the set
of vertices, and $\edges$ is the set of edges. 
Associated with each vertex $i$ is a state variable $x_i$
Let $x$ be joint state variable, i.e. $x=(x_i)_{i \in \vertex}$.
\em Conditional random fields \em (CRFs)~\citep{lafferty01conditional}
define a conditional distribution given the observation $z$ as follows
\begin{eqnarray}
	\label{CRF-def}
	\Pr(x|z) = \frac{1}{Z(z)}\prod_c\phi_c(x_c,z)
\end{eqnarray}
where $c$ is the index of cliques in the graph,
$\phi_c(x_c,z)$ is a non-negative potential function
defined over the clique $c$, and $Z(z) = \sum_x\prod_c\phi_c(x_c,z)$ is
the partition function.

Let $\{\tilde{x}\}$ be the set of observed state variables
with the empirical distribution $Q(\tilde{x})$, and $\w$ be
the parameter vector. Learning in CRFs is typically
by maximising the (log) likelihood
\begin{eqnarray}
	\label{log-ll-full}
	\w^* = \arg\max_{\w} \mathcal{L}(\w) =
		\arg\max_{\w} \sum_{\tilde{x}}Q(\tilde{x})\log \Pr(\tilde{x}|z;\w)
\end{eqnarray}

The gradient of the log-likelihood can be computed as
\begin{eqnarray}
	\label{CRF-grad}
	\nabla \mathcal{L}(\w) = \sum_{\tilde{x}}Q(\tilde{x})\sum_c \left(\nabla \log \phi_c(\tilde{x}_c,z)
					- \sum_{x_c}\Pr(x_c|z) \nabla \log \phi_c(x_c,z) \right)
\end{eqnarray}
Thus, the inference needed in CRF parameter estimation
is the computation of clique marginals $\Pr(x_c|z)$.

Typically, CRFs are parameterised using log-linear models (also known
as exponential family, Gibbs distribution or Maximum Entropy model), i.e.
$\phi_c(x_c,z) = \exp(\w^{\top}\f(x_c,z))$, where
$\f(.)$ is the feature vector and $\w$ is the vector
of feature weights. The features are also known as sufficient statistics in
the exponential family setting. 
Let $\F(x,z) = \sum_c\f(x_c,z)$ be the global feature.
Equation~\ref{CRF-grad} can be written as follows
\begin{eqnarray}
	\label{ll-grad-full}
	\nabla \mathcal{L} &=& \sum_{\tilde{x}}Q(\tilde{x})\sum_c\left( \f(\tilde{x}_c,z) - \sum_{x_c}\Pr(x_c|z)\f(x_c,z)\right) \\
					&=& \mathbb{E}_{Q(\tilde{x})}[\F] - \mathbb{E}_{\Pr(x|z)}[\F]
\end{eqnarray}
Thus gradient-based maximum likelihood learning 
in the log-linear setting boils down to estimating the feature expectations, also known as
expected sufficient statistics (ESS). 

The probabilistic nature of CRFs allows incorporating hidden variables
in a disciplined manner. Let $\tilde{x}=(\vartheta,h)$,
where $\vartheta$ is the set of visible variables,
and $h$ is the set of hidden variables. The incomplete log-likelihood
and its gradient are given as
\begin{eqnarray}
	\label{log-ll-hidden}
	\mathcal{L} &=& \sum_{\tilde{x}}Q(\tilde{x})\log \Pr(\vartheta|z) = \sum_{\tilde{x}}Q(\tilde{x})\log \sum_h \Pr(\vartheta,h|z) \nonumber\\
				&=& \sum_{\tilde{x}}Q(\tilde{x})(\log Z(\vartheta,z) - \log Z(z))
\end{eqnarray}
where $Z(\vartheta,z) = \sum_{h}\prod_c\phi_c(\vartheta_c,h_c,z)$. The gradient
reads
\begin{eqnarray}
	\label{log-ll-hidden-grad}
	\nabla \mathcal{L} &=& \mathbb{E}_{h|\vartheta,z}[\F(\vartheta,h,z)] - \mathbb{E}_{x|z}[\F(x,z)] \nonumber\\
					  	&=& \sum_{\tilde{x}}Q(\tilde{x})\sum_c\left( \sum_{h_c}\Pr(h_c|\vartheta,z)\f(\vartheta_c,h_c,z) - \sum_{x_c}\Pr(x_c|z)\f(x_c,z)\right)
\end{eqnarray}

%---
%\subsection{Extensions to CRFs}
%--
There have been extensions to CRFs, which can be broadly grouped 
into two categories.
The first category involves generalisation of model representation,
for example by extending CRFs for complex temporal structures
as in Dynamic CRFs ({\DCRF}s)~\citep{Sutton-et-alJMLR07},
segmental sequences as in Semi-Markov CRFs (SemiCRFs)~\citep{sarawagi04}, 
and relational data~\citep{Taskar-et-alUAI02}. 
The second category investigates learning schemes other than maximum likelihood,
for example perceptron~\citep{Collins-voted2002} and SVM~\citep{NIPS2003_AA04}.

{\DCRF}s and SemiCRFs are the most closely related to our {\HCRF}s.
{\DCRF}s are basically the conditional, undirected version of the
Dynamic Bayesian Networks \citep{Murphy02}.
The {\DCRF}s introduce multi-level of semantics, which help to represent
more complex sequential data. The main drawback
of the {\DCRF}s is the intractability of inference, except for
shallow models with small state space.

Similarly, the SemiCRFs are the conditional, undirected version
of the Semi-Markov HMMs. These allows
non-Markovian processes to be embedded in the chain CRFs, and thus
giving a possibility of modelling process duration.
Appendix~\ref{apdx:SemiCRF} analyses the SemiCRFs in more details.

Our {\HCRF}s deal with the inference problem of {\DCRF}s by limiting
to recursive processes, and thus obtaining efficient inference
via dynamic programming in the Inside-Outside family of algorithms.
Furthermore, it generalises the SemiCRFs to model multilevel of semantics.
It also addresses partial labels by introducing
appropriate constraints to the Inside-Outside algorithms.

%----------------------------
\section{Model Definition of {\HCRF}}
\label{sec:model}
%{\HCRF}- the model definitions

%----------------------------------
%\subsection{Model Definition}
%\label{sec:mode_def}

\begin{figure}[htb]
\begin{center}
\begin{tabular}{c}
\includegraphics[width=0.30\linewidth]{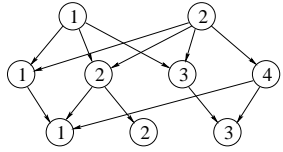} \\
\end{tabular}
\end{center}
\caption{The shared topological structure.}
\label{fig:shared-topo}
\end{figure}

Consider a hierarchically nested Markov process
with $D$ levels. Then as in the HHMMs (see Section~\ref{sec:bgr-HHMM}), the parent 
state embeds a child Markov chain whose states may in turn
contain child Markov chains. The family relation is defined in the \em model topology\em,
which is a state hierarchy of depth $D$. The model
has a set of states $S^d$ at each level $d \in [1,D]$, i.e. $S^d = \{1...|S^d|\}$, 
where $|S^d|$ is the number of states at level $d$. 
For each state $s^d \in S^d$ where $1\le d < D$,  the topological structure
also defines a set of children
$ch(s^d) \subset S^{d+1}$. Conversely, each child
$s^{d+1}$ has a set of parents $pa(s^{d+1}) \subset S^{d}$.
Unlike the original HHMMs where the child states belong
exclusively to the parent, the {\HCRF}s allow
arbitrary sharing of children between parents. 
For example, in Figure \ref{fig:shared-topo}, $ch(s^1=1) = \{1,2,3\}$,
and $pa(s^3=1) = \{1,2,4\}$. This helps to avoid an explosive number of
sub-states when $D$ is large, leading to fewer parameters
and possibly less training data and time. The shared topology 
has been investigated in the context of HHMMs in \citep{Bui-et-al04}.

The temporal evolution in the nested Markov processes with 
sequence length of $T$ operates as follows:
\begin{itemize}
\item As soon as a state is created 
	at level $d < D$, it
	\em initialises \em a child state at level $d+1$. The initialisation
	continues downward until reaching the bottom level. 
\item As soon as a child process at level $d+1$ \em ends\em, it
	returns control to its parent at level $d$, and in the case of $d > 1$, the parent
	either \em transits \em to a new parent state or returns to
	the grand-parent at level $d-1$.
\end{itemize}
The main requirement for the hierarchical nesting is that
the life-span of the child process belongs exclusively to
the life-span of the parent. For example, consider
a parent process at level $d$ starts a new state $s^d_{i:j}$ at time $i$ and 
persists until time $j$. At time $i$ the parent 
initialises a child state $s^{d+1}_i$ which
continues until it ends at time $k < j$, at which the child state
transits to a new child state $s^{d+1}_{k+1}$. The child process
exits at time $j$, at which the control from the child level is returned
to the parent $s^d_{i:j}$. Upon receiving the control the parent state 
$s^d_{i:j}$ may transit to a new parent state $s^d_{j+1:l}$, or
end at $j$, returning the control to the grand-parent at level
$d-1$. 

%-------------
%-------------

%-------------
\begin{figure}[htb]
\begin{center}
\begin{tabular}{c}
\input{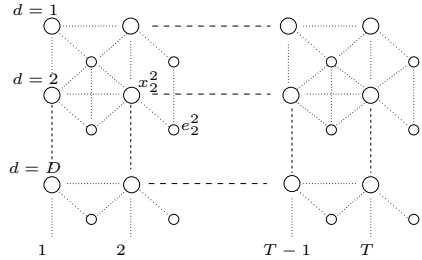}
\end{tabular}
\end{center}
\caption{The multi-level temporal model.}
\label{fig:DCRF}
\end{figure}
%-------------

We are now in a position to specify the nested Markov processes 
in a more formal way. Let us introduce a multi-level temporal graphical 
model of length $T$ with $D$ levels, starting from the top as 1 and the bottom
as $D$ (Figure~\ref{fig:DCRF}). At each level $d \in [1,D]$ and time index $i \in [1,T]$,
there is a node representing a state variable $x^d_i \in S^d = \{1,2,...,|S^d|\}$.
Associated with each $x^d_i$ is an ending indicator $e^d_i$
which can be either $1$ or $0$ to signify whether the state
$x^d_i$ ends or persists at $i$. The nesting nature of 
the {\HCRF}s is now realised by imposing the specific constraints
on the value assignment of ending indicators (Figure~\ref{fig:consistency}).

\begin{figure}[htb]
\begin{center}
\begin{tabular}{l}\hline
\q 	$\bullet$ The top state persists during the course of evolution, i.e.
		$e^1_{1:T-1} = 0$.\\
\q 	$\bullet$ When a state finishes, all of its descendants must also finish, \\
\q\q	i.e. $e^d_i = 1$ implies $e^{d+1:D}_i = 1$. \\
\q 	$\bullet$ When a state persists, all of its ancestors must also persist, \\
\q\q	i.e. $e^d_i = 0$ implies $e^{1:d-1}_i =0$.\\
\q 	$\bullet$ When a state transits, its parent must remain unchanged,
		i.e. $e^d_{i} = 1$, $e^{d-1}_i = 0$.\\
\q 	$\bullet$ The bottom states do not persists, i.e. $e^D_i = 1$ for all $i \in [1,T]$. \\
\q 	$\bullet$ All states end at $T$, i.e. $e^{1:D}_T = 1$.\\
\hline
\end{tabular}
\end{center}
\caption{Hierarchical constraints.}
\label{fig:consistency}
\end{figure}

\begin{figure}[htb]
\centering
%\begin{center}
%\begin{tabular}{c}
\input{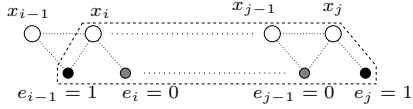}
%\end{tabular}
%\end{center}
\caption{An example of a state-persistence sub-graph.}
\label{fig:contemp-persist}
\end{figure}

\begin{figure}[htb]
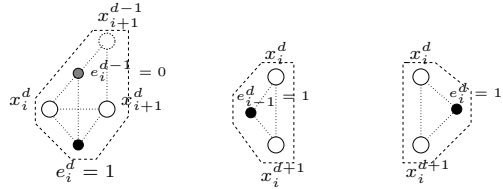

\begin{center}
\begin{tabular}{ccc}
\input{state-transition.pstex_t} & \input{state-init.pstex_t} & \input{state-ending.pstex_t}
%\\
%(a) transition & (b) initialisation & (c) ending
\end{tabular}
\end{center}
\caption{Sub-graphs for state transition (left), initialisation (middle) and ending (right).}
\label{fig:contemp-transit-init-end}
\end{figure}

Thus, specific value assignments of ending indicators provide
\em contexts \em that realise the evolution of the model states
in both hierarchical (vertical) and temporal (horizontal) directions.
Each context at a level and associated state variables 
form a \em contextual clique\em, and we identify four 
contextual clique types:
\begin{itemize}
\item \em State-persistence \em: This corresponds to 
	the life time of a state at a given level (see Figure~\ref{fig:contemp-persist}).
	Specifically, given a context $c = (e^d_{i-1:j}=(1,0,..,0,1))$,
	then $\sigma^{persist,d}_{i:j} = (x^d_{i:j},c)$, is a contextual clique that specifies
	the life-span $[i,j]$ of any state $s = x^d_{i:j}$.
\item \em State-transition \em: This corresponds to a state at level 
	$d \in [2,D]$ at time $i$ transiting to a new
	state (see Figure~\ref{fig:contemp-transit-init-end}a).
	Specifically, given a context $c = (e^{d-1}_i = 0, e^d_i = 1)$
	then $\sigma^{transit,d}_i = (x^{d-1}_{i+1},x^d_{i:i+1},c)$ is a contextual clique that
	specifies the transition of $x^d_i$ to $x^d_{i+1}$ at time
	$i$ under the 	same parent $x^{d-1}_{i+1}$.
\item \em State-initialisation \em: This corresponds to a 
	state at level $d \in [1,D-1]$ initialising a new
	child state at level $d+1$ at time $i$ (see Figure~\ref{fig:contemp-transit-init-end}b). 
	Specifically, given a context $c = (e^d_{i-1} = 1)$,
	then $\sigma^{init,d}_i = (x^d_i,x^{d+1}_i,c)$ is a contextual
	clique that specifies the initialisation 
	at time $i$ from the parent $x^d_i$ to the child $x^{d+1}_i$.
\item \em State-ending \em: This corresponds to 
	a state at level $d \in [1,D-1]$ to end at time $i$ (see Figure~\ref{fig:contemp-transit-init-end}c). 
	Specifically, given a context $c = (e^d_{i} = 1)$,
	then $\sigma^{end,d}_i = (x^{d}_{i},x^{d+1}_i,c)$ is a contextual
	clique that specifies the ending of $x^{d}_i$ at time $i$
	with the last child $x^{d+1}_i$.
\end{itemize}

In the {\HCRF} we are interested in the \em conditional \em setting
in which the entire state variables and ending indicators $(x^{1:D}_{1:T},e^{1:D}_{1:T})$ are
conditioned on observational sequences $z$. For example,
in computational linguistics, the observation is often the sequence
of words and the state variables might be the part-of-speech tags
and the phrases.

To capture the correlation between variables and such conditioning,
we define a non-negative potential function $\psi(\sigma,z)$
over each contextual clique $\sigma$. Figure~\ref{fig:poten}
shows the notations for potentials that correspond to the four contextual clique types
we have identified above. Details of potential specification
are described in the Section~\ref{sec:param}.

\begin{figure}[htb]
\begin{center}
\begin{tabular}{l}\hline
\q $\bullet$ $R^{d,s,z}_{i:j} = \psi(\sigma^{persist,d}_{i:j},z)$
	where $s = x^d_{i:j}$. \\
\q $\bullet$ $A^{d,s,z}_{u,v,i} = \psi(\sigma^{transit,d}_{i},z)$ 
	where $s=x^{d-1}_{i+1}$ and $u = x^d_i, v = x^d_{i+1}$. \\
\q $\bullet$ $\pi^{d,s,z}_{u,i} = \psi(\sigma^{init,d}_i,z)$
	where $s = x^d_i, u = x^{d+1}_i$. \\
\q $\bullet$ $E^{d,s,z}_{u,i} = \psi(\sigma^{end,d}_{i},z)$ 
	where $s = x^d_i, u = x^{d+1}_i$.\\
\hline
\end{tabular}
\end{center}
\caption{Shorthands for contextual clique potentials.}
\label{fig:poten}
\end{figure}

%Let $\mathcal{V} = (x^{1:D}_{1:T},e^{1:D}_{1:T})$ denote the set
%of all variables and let $\tau^d = \{i_k\}_{k=1}^m$ denote the set of all time indices
%where $e^d_{i_k} = 1$.  A \em configuration \em $\zeta$ of the model is a complete 
%assignment of all the states and ending indicators
%$(x^{1:D}_{1:i},e^{1:D}_{1:T})$ which satisfies the set of
%hierarchical constraints in Figure~\ref{fig:consistency}.

Let $\zeta = (x^{1:D}_{1:T},e^{1:D}_{1:T})$ denote the set
of all variables that satisfies the set of
hierarchical constraints in Figure~\ref{fig:consistency}.
%Let $\tau^d = \{i_k\}_{k=1}^m$ denote the set of all time indices
%where $e^d_{i_k} = 1$. 
Let $\tau^d$ denote ordered set of all ending time indices
at level $d$, i.e. if $i \in \tau^d$ then $e^d_{i} = 1$.
The joint potential defined for each configuration
is the product of all contextual clique potentials 
over all ending time indices $i \in [1,T]$ and all semantic levels $d \in [1,D]$:
{\small\begin{eqnarray}
	\label{joint-potential}
	\Phi[\zeta,z] &=& \bigg[\prod_{d \in [1,D]}\prod_{i_k,i_{k+1}\in\tau^d} R^{d,s,z}_{i_k+1:i_{k+1}}\bigg]\times \nonumber\\
					& &\times\prod_{d \in [1,D-1]}\left\{
						\bigg[\prod_{i_k\in \tau^{d+1},i_k\notin\tau^{d}} A^{d+1,s,z}_{u,v,i_k}\bigg]
						\bigg[\prod_{i_k \in \tau^{d+1}} \pi^{d,s,z}_{u,i_k+1}\bigg]
						\bigg[\prod_{i_k \in \tau^{d+1}}E^{d,s,z}_{u,i_k}\bigg]\right\}
\end{eqnarray}}

The conditional distribution is given as
\begin{eqnarray}
	\label{model-dis}
	\Pr(\zeta|z) = \frac{1}{Z(z)}\Phi[\zeta,z]
\end{eqnarray}
where $Z(z) = \sum_{\zeta}\Phi[\zeta,z]$ is the partition function
for normalisation. 

In what follows we omit $z$ for clarity, and implicitly use
it as part of the partition function $Z$ and the potential $\Phi[.]$. 
It should be noted that in the unconditional formulation, there is only
a single $Z$ for all data instances. In conditional setting
there is a $Z(z)$ for each data instance $z$.\\

\textbf{Remarks}: The temporal model of {\HCRF}s presented here is not a standard graphical model
\citep{Lauritzen96} since the connectivity (and therefore
the clique structures) is not fixed. The potentials are
defined on-the-fly depending on the context of assignments of
ending indicators. Although the model topology is identical
to that of shared structure HHMMs \citep{Bui-et-al04},
the unrolled temporal representation is an undirected graph
and the model distribution is formulated in a discriminative way.
Furthermore, the state persistence potentials capture
duration information that is not available in
the dynamic DBN representation of the HHMMs in \citep{Murphy-Paskin01}.

In the way the potentials are introduced it may first appear to
resemble the clique templates in the discriminative relational Markov 
networks (RMNs) \citep{Taskar-et-alUAI02}. It is, however, different
because cliques in the {\HCRF}s are dynamic and context-dependent.

%----------------------------
\section{Asymmetric Inside-Outside Algorithm}
\label{sec:AIO-all}
This section describes a core inference engine called
Asymmetric Inside-Outside (AIO) algorithm, which is partly
adapted from the generative, directed counter part of HHMMs in \citep{Bui-et-al04}.
We now show how to compute the building blocks that are
needed in most inference and learning tasks.

%--------------
\subsection{Building Blocks and Conditional Independence}
\label{sec:hcrf-defs}

\begin{figure}[htb]
\begin{center}
 \begin{tabular}{ccc}
\includegraphics[width=0.45\linewidth]{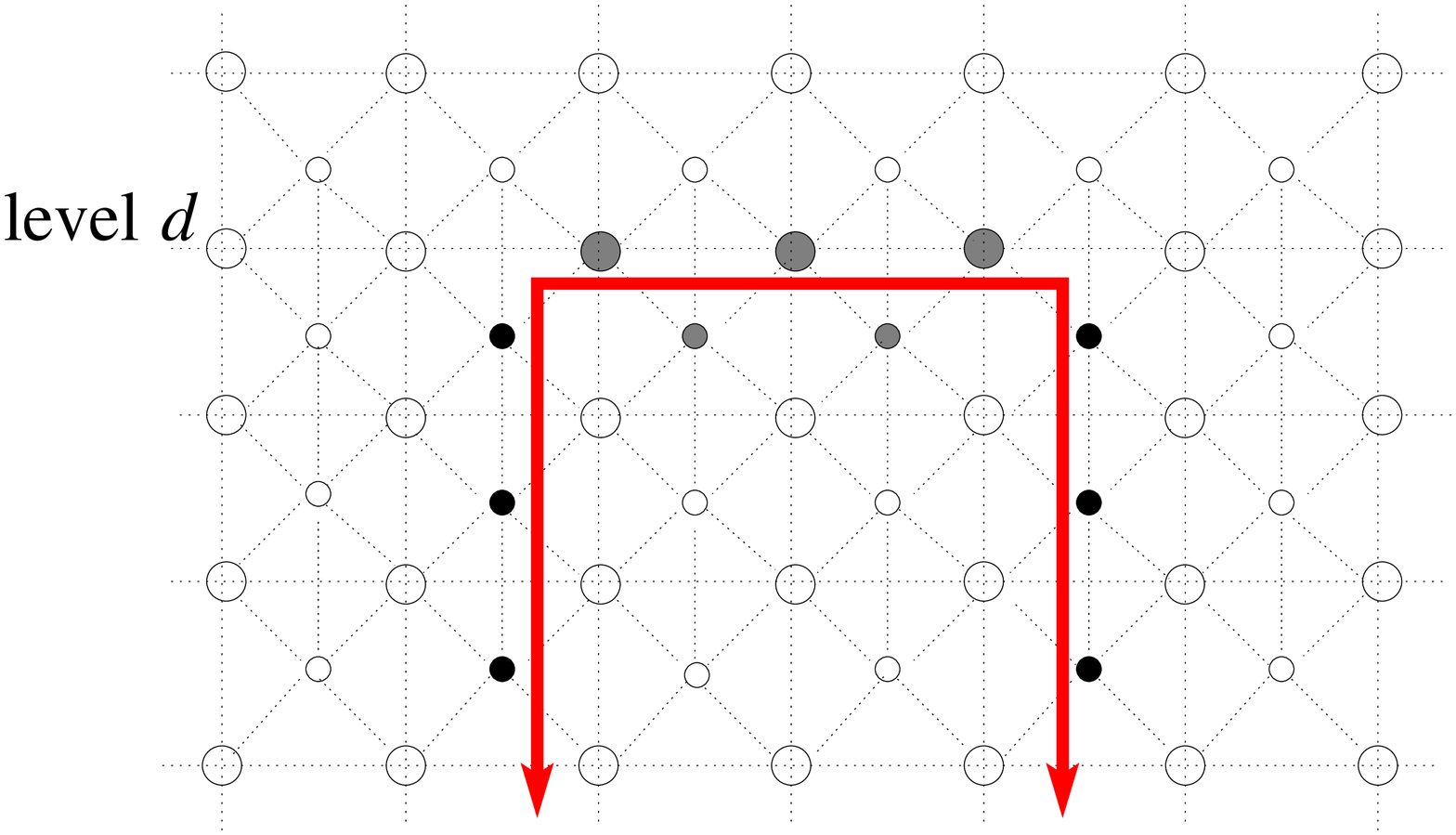} & &
\includegraphics[width=0.45\linewidth]{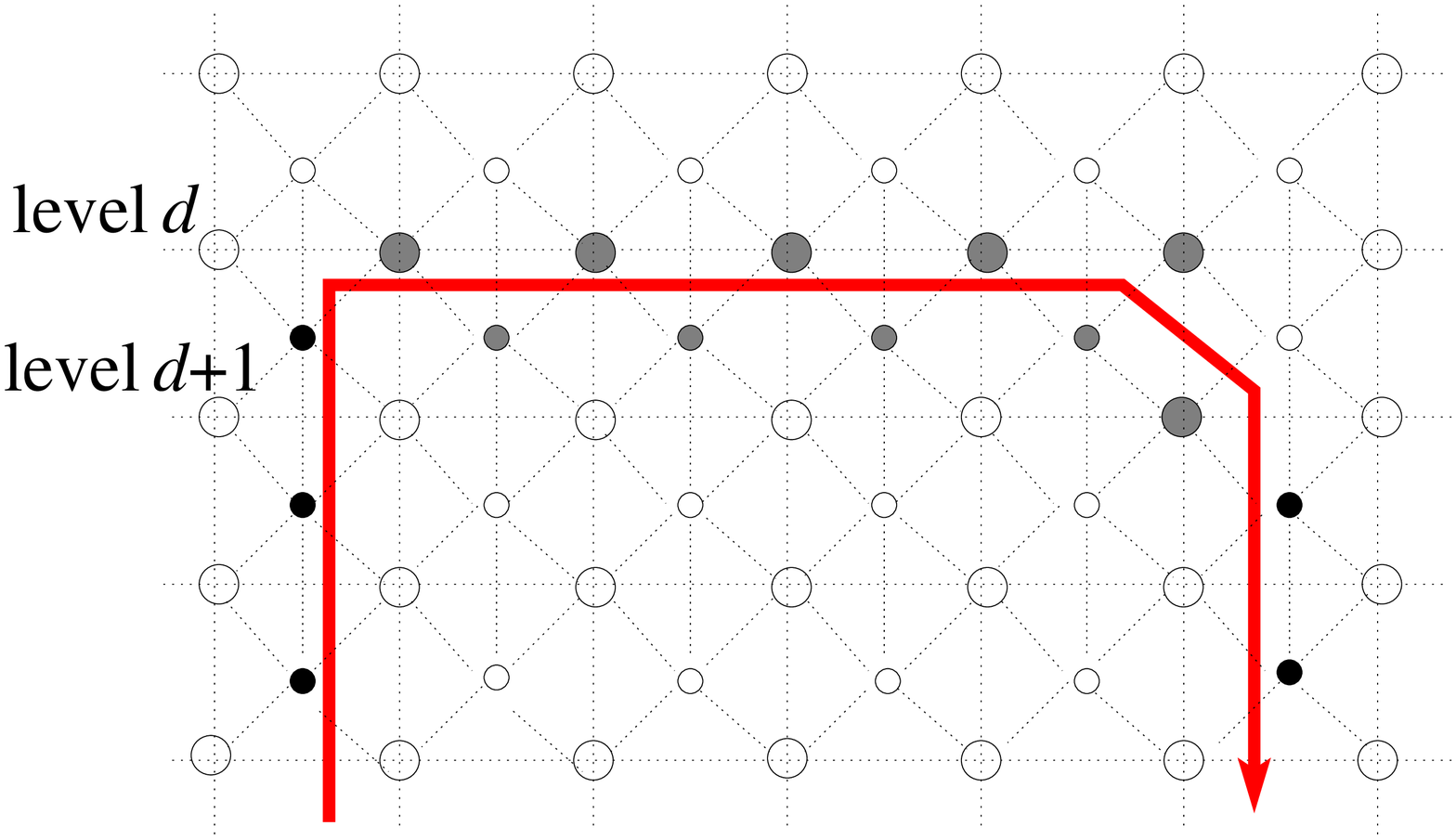} \\
(a) & & (b)
\end{tabular}
\end{center}
\caption{(a) Symmetric Markov blanket, and (b) Asymmetric Markov blanket.}
\label{fig:Markov-blanket}
\end{figure}

%--------------
\subsubsection{Contextual Markov blankets}
In this subsection we define elements that are building blocks for
inference and learning. These building blocks are identified
given the corresponding boundaries. Let us introduce
two types of boundaries:  the contextual \emph{symmetric}
and \emph{asymmetric Markov blankets}.
%and their corresponding
%subsets of variables separated by the boundaries.

%-----------
\begin{definition}
	A symmetric Markov blanket at level $d$ for
	a state $s$ starting at $i$ and ending at $j$
	is the following set
	\begin{eqnarray}
		\Pi^{d,s}_{i:j} = (x^d_{i:j}=s,e^{d:D}_{i-1}=1,e^{d:D}_{j}=1,e^{d}_{i:j-1} = 0)
	\end{eqnarray}
\end{definition}
%--------
%-----------
\begin{definition}
	\label{def:sym-variables}
	Let $\Pi^{d,s}_{i:j}$ be a symmetric Markov blanket, we define
	$\zeta^{d,s}_{i:j}$ and $\underline{\zeta}^{d,s}_{i:j}$ as follows
	\begin{eqnarray}
		\label{inside-var}
		\zeta^{d,s}_{i:j} &=& (x^{d+1:D}_{i:j},e^{d+1:D}_{i:j-1})\\
		\label{outside-var}
		\underline{\zeta}^{d,s}_{i:j} &=& \zeta\backslash (\zeta^{d,s}_{i:j},\Pi^{d,s}_{i:j})
	\end{eqnarray}
	subject to $x^d_{i:j} = s$. Further, we define
	\begin{eqnarray}
			\hat{\zeta}^{d,s}_{i:j} &=& (\zeta^{d,s}_{i:j},\Pi^{d,s}_{i:j}) \\
			\hat{\underline{\zeta}}^{d,s}_{i:j} &=& (\underline{\zeta}^{d,s}_{i:j},\Pi^{d,s}_{i:j})
	\end{eqnarray}
\end{definition}
Figure~\ref{fig:Markov-blanket}a shows an example of a symmetric Markov blanket
(represented by a double-arrowed line).

%------
\begin{definition}
	A asymmetric Markov blanket at level $d$
	for a parent state $s$ starting at $i$ and a child state
	$u$ ending at $j$ is the following set
	\begin{eqnarray}
		\Gamma^{d,s}_{i:j}(u) = (x^d_{i:j} = s, x^{d+1}_j = u,
					e^{d:D}_{i-1}=1,e^{d+1:D}_{j}=1,e^{d}_{i:j-1} = 0)
	\end{eqnarray}
\end{definition}
%--------

%------
\begin{definition}
	\label{def:asym-variables}
	Let $\Gamma^{d,s}_{i:j}(u)$ be an asymmetric Markov blanket, we define
	${\zeta}^{d,s}_{i:j}(u)$ and ${\underline{\zeta}}^{d,s}_{i:j}(u)$ as follows
	\begin{eqnarray}
		{\zeta}^{d,s}_{i:j}(u) &=& (x^{d+1:D}_{i:j-1},x^{d+2:D}_j,e^{d+1:D}_{i:j-1})\\
		{\underline{\zeta}}^{d,s}_{i:j}(u) &=& \zeta\backslash ({\zeta}^{d,s}_{i:j}(u),\Gamma^{d,s}_{i:j}(u))
	\end{eqnarray}
	subject to $x^d_{i:j} = s$ and $x^{d+1}_j = u$. Further, we define
	\begin{eqnarray}
		\label{asym-inside-var}
		\hat{\zeta}^{d,s}_{i:j}(u) &=& (\zeta^{d,s}_{i:j}(u),\Gamma^{d,s}_{i:j}(u)) \\
		\label{asym-outside-var}
		\hat{\underline{\zeta}}^{d,s}_{i:j}(u) &=& (\underline{\zeta}^{d,s}_{i:j}(u),\Gamma^{d,s}_{i:j}(u))
	\end{eqnarray}
	
\end{definition}

Figure~\ref{fig:Markov-blanket}b shows an example of asymmetric Markov blanket
(represented by an arrowed line).

\textbf{Remark}: The concepts of contextual Markov blankets (or Markov blankets for short)
are different from those in traditional Markov random fields and Bayesian networks because they are
specific assignments of a subset of variables, rather than a collection of variables.

%--------------
\subsubsection{Conditional independence}
Given these two definitions we have the following
propositions of conditional independence.

%------
\begin{proposition}
\label{conj:sym-CI}
	$\zeta^{d,s}_{i:j}$ and $\underline{\zeta}^{d,s}_{i:j}$ 
	are conditionally independent given $\Pi^{d,s}_{i:j}$
	\begin{eqnarray}
		\Pr(\zeta^{d,s}_{i:j},\underline{\zeta}^{d,s}_{i:j}|\Pi^{d,s}_{i:j})
			= \Pr(\zeta^{d,s}_{i:j}|\Pi^{d,s}_{i:j})
				\Pr(\underline{\zeta}^{d,s}_{i:j}|\Pi^{d,s}_{i:j})
	\end{eqnarray}
\end{proposition}

This proposition gives rise to the following factorisation
\begin{eqnarray}
	\label{sym-CI-factorise}
	\Pr(\zeta) = \Pr(\Pi^{d,s}_{i:j})\Pr(\zeta^{d,s}_{i:j},\underline{\zeta}^{d,s}_{i:j}|\Pi^{d,s}_{i:j})
					= \Pr(\Pi^{d,s}_{i:j})\Pr(\zeta^{d,s}_{i:j}|\Pi^{d,s}_{i:j})\Pr(\underline{\zeta}^{d,s}_{i:j}|\Pi^{d,s}_{i:j})
\end{eqnarray}

%------
\begin{proposition}
\label{conj:asym-CI}
	${\zeta}^{d,s}_{i:j}(u)$ and ${\underline{\zeta}}^{d,s}_{i:j}(u)$
	are conditionally independent given $\Gamma^{d,s}_{i:j}(u)$
	\begin{eqnarray}
		\Pr({\zeta}^{d,s}_{i:j}(u),\underline{\zeta}^{d,s}_{i:j}(u)|\Gamma^{d,s}_{i:j}(u))
			= \Pr({\zeta}^{d,s}_{i:j}(u)|\Gamma^{d,s}_{i:j}(u))
				\Pr(\underline{\zeta}^{d,s}_{i:j}(u)|\Gamma^{d,s}_{i:j}(u))
	\end{eqnarray}
\end{proposition}

The following factorisation is a consequence of Proposition~\ref{conj:asym-CI}
\begin{eqnarray}
	\Pr(\zeta) &=& \Pr(\Gamma^{d,s}_{i:j}(u))\Pr({\zeta}^{d,s}_{i:j}(u),{\underline{\zeta}}^{d,s}_{i:j}(u)|\Gamma^{d,s}_{i:j}(u)) \nonumber\\
				&=& \Pr(\Gamma^{d,s}_{i:j}(u))\Pr({\zeta}^{d,s}_{i:j}(u)|\Gamma^{d,s}_{i:j}(u))
								\Pr({\underline{\zeta}}^{d,s}_{i:j}(u)|\Gamma^{d,s}_{i:j}(u))
\end{eqnarray}

The proof of Propositions~\ref{conj:sym-CI} and~\ref{conj:asym-CI} is given in
Appendix~\ref{apdx:proof-CI}. 

%--------------
\subsubsection{Symmetric Inside/Outside Masses}
From Equation~\ref{outside-var} we have
$\zeta = (\zeta^{d,s}_{i:j},\Pi^{d,s}_{i:j},\underline{\zeta}^{d,s}_{i:j})$.
Since $\Pi^{d,s}_{i:j}$ separates $\zeta^{d,s}_{i:j}$
from $\underline{\zeta}^{d,s}_{i:j}$,
we can group local potentials in Equation~\ref{joint-potential} into three parts:
$\Phi[\hat{\zeta}^{d,s}_{i:j}[$, $\Phi[\hat{\underline{\zeta}}^{d,s}_{i:j}[$, and 
$\Phi[\Pi^{d,s}_{i:j}]$. By `grouping' we mean to multiply all
the local potentials belonging to a certain part, 
in the same way that we group all the local potentials belonging
to the model in Equation~\ref{joint-potential}.
Note that although $\hat{\zeta}^{d,s}_{i:j}$ contains
$\Pi^{d,s}_{i:j}$ we do not group 
$\Phi[\Pi^{d,s}_{i:j}]$ into $\Phi[\hat{\zeta}^{d,s}_{i:j}]$.
The same holds for  $\Phi[\hat{\underline{\zeta}}^{d,s}_{i:j}]$.

By definition of the state-persistence clique potential (Figure~\ref{fig:poten}),
we have $\Phi[\Pi^{d,s}_{i:j}] = R^{d,s}_{i:j}$.
Thus Equation~\ref{joint-potential} can be replaced by
\begin{eqnarray}
	\label{factorise}
	\Phi[\zeta] = \Phi[\hat{\zeta}^{d,s}_{i:j}]R^{d,s}_{i:j}\Phi[\hat{\underline{\zeta}}^{d,s}_{i:j}]
\end{eqnarray}
There are two special cases: (1) when $d=1$, $\Phi[\hat{\underline{\zeta}}^{1,s}_{1:T}]=1$
for $s \in S^1$, and (2) when $d=D$, $\Phi[\hat{\zeta}^{D,s}_{i:i}]=1$
for $s \in S^D$ and $i \in [1,T]$.
This factorisation plays an important role in efficient inference.

We know define a quantity called \emph{symmetric inside mass} $\Delta^{d,s}_{i:j}$, 
and another called \emph{symmetric outside mass} $\Lambda^{d,s}_{i:j}$.
%-------------
\begin{definition}
	Given a symmetric Markov blanket $\Pi^{d,s}_{i:j}$, the symmetric inside mass $\Delta^{d,s}_{i:j}$
	and the symmetric outside mass $\Lambda^{d,s}_{i:j}$ are defined as
	\begin{eqnarray}
		\label{inside-def}
		\Delta^{d,s}_{i:j} &=& \sum_{\zeta^{d,s}_{i:j}}\Phi[\hat{\zeta}^{d,s}_{i:j}] \\
		\label{outside-def}
		\Lambda^{d,s}_{i:j} &=& \sum_{\underline{\zeta}^{d,s}_{i:j}}\Phi[\hat{\underline{\zeta}}^{d,s}_{i:j}]
	\end{eqnarray}
As special cases we have 
$\Lambda^{1,s}_{1:T}=1$ and $s \in S^1$, and
$\Delta^{D,s}_{i:i}=1$ for $i\in [1,T]$, $s\in S^D$.
For later use let us introduce the `full' symmetric inside mass
$\hat{\Delta}^{d,s}_{i:j}$ and the `full' symmetric outside mass $\hat{\Lambda}^{d,s}_{i:j}$
as
\begin{eqnarray}
	\hat{\Delta}^{d,s}_{i:j} &=& R^{d,s}_{i:j}\Delta^{d,s}_{i:j} \\
	\hat{\Lambda}^{d,s}_{i:j} &=& R^{d,s}_{i:j}\Lambda^{d,s}_{i:j}
\end{eqnarray}
\end{definition}
%-------------

In the rest of the thesis, when it is clear in the context, 
we will use \emph{inside mass} as a shorthand for symmetric inside mass,
\emph{outside mass} for symmetric outside mass, \emph{full-inside mass} for full-symmetric inside mass,
and \emph{full-outside mass} for full-symmetric outside mass.

Thus, from Equation~\ref{factorise} the partition function can be computed 
from the full-inside mass at the top level ($d=1$)
\begin{eqnarray}
	\label{Z-I}
	Z &=& \sum_{\zeta}\Phi[\zeta]  \nonumber\\
		&=& \sum_{\zeta^{1,s}_{1:T}}\sum_{s \in S^1}\Phi[\hat{\zeta}^{1,s}_{1:T}]R^{1,s}_{1:T} \nonumber\\
		&=&  \sum_{s \in S^1} \Delta^{d,s}_{1:T} R^{d,s}_{1:T} \nonumber \\
		&=& \sum_{s \in S^1} \hat{\Delta}^{1,s}_{1:T}
\end{eqnarray}
With the similar derivation the partition function can also be computed
from the full-outside mass at the bottom level ($d=D$)
\begin{eqnarray}
	\label{Z-O}
	Z = \sum_{s \in S^D} \hat{\Lambda}^{D,s}_{i:i}, \mbox{ for any } i \in [1,T]
\end{eqnarray}
In fact, we will prove a more general way to compute $Z$ in Appendix~\ref{apdx:margin}
\begin{eqnarray}
	\label{Z-general}
	Z = \sum_{s \in S^d}\sum_{i \in [1,t]}\sum_{j \in [t,T]}\Delta^{d,s}_{i:j}\Lambda^{d,s}_{i:j}R^{d,s}_{i:j}
\end{eqnarray}
for any $t \in [1,T]$ and $d \in [2,D-1]$.
These relations are summarised in Figure~\ref{fig:Z-inside-outside}.

%---------
\begin{figure}[htb]
\begin{center}
\begin{tabular}{l}\hline
$\bullet$ $Z	= \sum_{s \in S^1} \hat{\Delta}^{1,s}_{1:T}$ \\
$\bullet$ $Z = \sum_{s \in S^D} \hat{\Lambda}^{D,s}_{i:i}$ for any  $i \in [1,T]$ \\
$\bullet$ $Z = \sum_{s \in S^d}\sum_{i \in [1,t]}\sum_{j \in [t,T]}\Delta^{d,s}_{i:j}\Lambda^{d,s}_{i:j}R^{d,s}_{i:j}$
			for any $t \in [1,T]$ and $d \in [2,D-1]$\\
\hline
\end{tabular}
\end{center}
\caption{Computing the partition function from the full-inside mass and full-outside mass.}
\label{fig:Z-inside-outside}
\end{figure}
%---------

Given the fact that $\zeta^{d,s}_{i:j}$ is separated from the rest
of variables by the symmetric Markov blanket $\Pi^{d,s}_{i:j}$, we have
Proposition~\ref{conj:SIO-conds}.
\begin{proposition}
\label{conj:SIO-conds}
	The following relations hold
	\begin{eqnarray}
		\label{SIO-cond}
		\Pr(\zeta^{d,s}_{i:j}|\Pi^{d,s}_{i:j}) &=& \frac{1}{\Delta^{d,s}_{i:j}}\Phi[\hat{\zeta}^{d,s}_{i:j}]\\
		\label{SIO-cond2}
		\Pr(\underline{\zeta}^{d,s}_{i:j}|\Pi^{d,s}_{i:j}) &=& \frac{1}{\Lambda^{d,s}_{i:j}}\Phi[\hat{\underline{\zeta}}^{d,s}_{i:j}] \\
		\label{sym-blanket-prob}
		 \Pr(\Pi^{d,s}_{i:j}) &=& \frac{1}{Z}\Delta^{d,s}_{i:j}R^{d,s}_{i:j}\Lambda^{d,s}_{i:j}
	\end{eqnarray}
\end{proposition}
The proof of this proposition is given in Appendix~\ref{apdx:proof-SIO-conds}.

%--------------
\subsubsection{Asymmetric Inside/Outside Masses}
Recall that we have introduced the concept of asymmetric Markov blanket
$\Gamma^{d,s}_{i:j}(u)$ which separates $\zeta^{d,s}_{i:j}(u)$ and $\underline{\zeta}^{d,s}_{i:j}(u)$.
Let us group all the local contextual clique potentials associated with 
$\zeta^{d,s}_{i:j}(u)$ and $\Gamma^{d,s}_{i:j}(u)$
into a joint potential $\Phi[\hat{{\zeta}}^{d,s}_{i:j}(u)]$.
Similarly, we group all local potentials associated with $\underline{\zeta}^{d,s}_{i:j}(u)$
and $\Gamma^{d,s}_{i:j}(u)$ into a joint potential $\Phi[\hat{\underline{\zeta}}^{d,s}_{i:j}(u)]$.
Note that $\Phi[\hat{\underline{\zeta}}^{d,s}_{i:j}(u)])$
includes the state-persistence potential $R^{d,s}_{i:j}$.

\begin{definition}
	Given the asymmetric Markov blanket $\Gamma^{d,s}_{i:j}(u)$,
	the asymmetric inside mass $\alpha^{d,s}_{i:j}(u)$
	and the asymmetric outside mass $\lambda^{d,s}_{i:j}(u)$
	are defined as follows
	\begin{eqnarray}
		\label{IF-def}
		\alpha^{d,s}_{i:j}(u) &=& \sum_{{\zeta}^{d,s}_{i:j}(u)}\Phi[\hat{{\zeta}}^{d,s}_{i:j}(u)]\\
		\label{OB-def}
		\lambda^{d,s}_{i:j}(u) &=& \sum_{{\underline{\zeta}}^{d,s}_{i:j}(u)}\Phi[\hat{{\underline{\zeta}}}^{d,s}_{i:j}(u)]
	\end{eqnarray}
\end{definition}

The relationship between the asymmetric outside mass and asymmetric inside mass is
analogous to that between the outside and inside masses. 
However, there is a small difference, that is,
the asymmetric outside mass `owns' the segment $x^{d}_{i:j}=s$ and the associated 
state-persistence potential $R^{d,s}_{i:j}$, whilst
the outside mass $\Lambda^d_{i:j}(s)$ does not.

%--
\subsection{Computing Inside Masses}
\label{sec:IF}
%--------------
%
\begin{figure}[htb]
\begin{center}
 \begin{tabular}{c}
\includegraphics[width=0.45\linewidth]{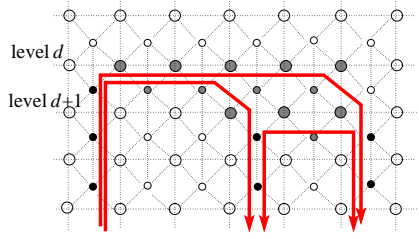}
\end{tabular}
\end{center}
\caption{Decomposition with respect to symmetric/asymmetric Markov blankets.}
\label{fig:DCRF-asym2}
\end{figure}

In this subsection we show how to recursively compute
the pair: inside mass and asymmetric inside mass. The key idea here is to exploit
the decomposition within the asymmetric Markov blanket.
As shown in Figure~\ref{fig:DCRF-asym2}, an outer asymmetric Markov blanket
can be decomposed into a sub-asymmetric Markov blanket
and a symmetric blanket.

%-----------
\subsubsection{Computing asymmetric inside mass from inside mass}
Assume that within the asymmetric Markov blanket $\Gamma^{d,s}_{i:j}(u)$,
the child $u$ starts somewhere at $t \in [i,j]$ and ends at $j$, i.e.
$x^{d+1}_{t:j} = u$, $e^{d+1}_{t:j-1}=0$ and $e^{d+1:D-1}_{t-1} = 1$. Let us consider
two cases: $t > i$ and $t = i$.

\textbf{Case 1}. For $t > i$, denote by $v = x^{d+1}_{t-1}$. 
We have two smaller blankets within $\Gamma^{d,s}_{i:j}(u)$: 
the symmetric blanket $\Pi^{d+1,u}_{t:j}$ associated with the child $u=x^{d+1}_{t:j}$,
and the asymmetric blanket $\Gamma^{d,s}_{i:t-1}(v)$ associated with
the child $v$ ending at $t-1$ under the parent $s$.
Figure~\ref{fig:DCRF-asym2} illustrates the blanket decomposition.
The assignment ${\zeta}^{d,s}_{i:j}(u)$
can be decomposed as 
\begin{eqnarray}
	{\zeta}^{d,s}_{i:j}(u) = 
		({\zeta}^{d,s}_{i:t-1}(v),\zeta^{d+1,u}_{t:j},u=x^{d+1}_{t:j},e^{d}_{t-1:j-1}=0,e^{d+1:D}_{t-1}=1)
\end{eqnarray}

Thus, the joint potential 
$\Phi[\hat{{\zeta}}^{d,s}_{i:j}(u)]$ can be factorised as
follows
\begin{eqnarray}
	\label{IF-factorise}
	\Phi[\hat{{\zeta}}^{d,s}_{i:j}(u)]
		 = \Phi[\hat{{\zeta}}^{d,s}_{i:t-1}(v)]
		 	\Phi[\hat{\zeta}^{d+1,u}_{t:j}]A^{d+1,s}_{v,u,t-1}R^{d+1,u}_{t:j}
\end{eqnarray}
The transition potential $A^{d+1,s}_{v,u,t-1}$ is enabled in
the context $c = (e^d_{t-1}=0,e^{d+1}_{t-1}=1,x^{d}_{t}=s,x^{d+1}_{t-1}=v,x^{d+1}_{t}=u)$, and
the state-persistence potential  $R^{d+1,u}_{t:j}$
in the context $c = (e^{d+1}_{t:j-1}=0,e^{d+1:D}_{t-1}=1,e^{d+1:D}_{j}=1,x^{d+1}_{t:j}=u)$.

\textbf{Case 2}. For $t=i$, the asymmetric blanket $\Gamma^{d,s}_{i:t-1}(v)$ does not exist since
$i > t-1$. We have the following decompositions of assignment
$\hat{\zeta}^{d,s}_{i:j}(u)	= (\hat{\zeta}^{d+1,u}_{i:j},e^{d}_{i-1} = 1, e^{d}_{i:j-1} = 0)$.
In the context $c=(e^{d}_{i-1} = 1)$, the state-initialisation
potential $\pi^{d,s}_{u,i}$ is activated. Thus we have
\begin{eqnarray}
	\label{IF-factorise2}
	\Phi[\hat{{\zeta}}^{d,s}_{i:j}(u)]
		 = \pi^{d,s}_{u,i}\Phi[\hat{\zeta}^{d+1,u}_{i:j}]R^{d+1,u}_{i:j}
\end{eqnarray}

Substituting Equations~\ref{IF-factorise} and \ref{IF-factorise2}
into Equation~\ref{IF-def}, and together with the fact that
$t$ can take any value in the interval $[i,j]$,
and $v$ can take any value in $S^{d+1}$,
we have the following relation
\begin{eqnarray}
	\label{IF-detail}
	\alpha^{d,s}_{i:j}(u) &=& \sum_{t \in [i+1,j]}\sum_{v \in S^{d+1}}\sum_{{{\zeta}}^{d,s}_{i:t-1}(v)}\sum_{{\zeta}^{d+1,u}_{t:j}}
							\Phi[\hat{{\zeta}}^{d,s}_{i:t-1}(v)]
						 	\Phi[\hat{\zeta}^{d+1,u}_{t:j}]A^{d+1,s}_{v,u,t-1}R^{d+1,u}_{t:j} + \nonumber\\
						 	&& \q\q\q\q\q+ \sum_{{\zeta}^{d+1,u}_{i:j}}\pi^{d,s}_{u,i}\Phi[\hat{\zeta}^{d+1,u}_{i:j}]R^{d+1,u}_{i:j} \nonumber\\
						&=& \sum_{t \in [i+1,j]}\sum_{v \in S^{d+1}}\alpha^{d,s}_{i:t-1}(v)
							\hat{\Delta}^{d+1,u}_{t:j}A^{d+1,s}_{v,u,t-1} + \hat{\Delta}^{d+1,u}_{i:j}\pi^{d,s}_{u,i}					
\end{eqnarray}
As we can see, the asymmetric inside mass $\alpha$ plays
the role of a \emph{forward message} starting from the starting
time $i$ to the ending time $j$. There is a recursion
where the asymmetric inside mass ending at time $j$ is computed
from all the asymmetric inside masses ending at time $t-1$, for
$t \in [i+1,j$.

There are special cases for the asymmetric inside mass: (1) when
$i=j$, we only have
\begin{eqnarray}
	\label{IF-special}
	\alpha^{d,s}_{i:i}(u)= \hat{\Delta}^{d+1,s}_{i:i}\pi^{d,s}_{u,i}
\end{eqnarray}
and (2) when $d=D-1$, the sum over the index $t$ 
as in Equation~\ref{IF-detail} is not allowed since
at level $D$ the inside mass only spans a single index. 
We have the following instead
\begin{eqnarray}
	\label{IF-special2}
	\alpha^{D-1,s}_{i:j}(u) 
			&=& \sum_{v \in S^{d+1}}\alpha^{D-1,s}_{i:j-1}(v)\hat{\Delta}^{D,u}_{j:j}A^{D,s}_{v,u,j-1} \nonumber\\
			&=& \sum_{v \in S^{d+1}}\alpha^{D-1,s}_{i:j-1}(v)R^{D,u}_{j:j}A^{D,s}_{v,u,j-1}
\end{eqnarray}

%-----------
\subsubsection{Computing inside mass from asymmetric inside mass}
Notice the relationship between the asymmetric Markov blanket $\Gamma^{d,s}_{i:j}(u)$
and the symmetric blanket $\Pi^{d,s}_{i:j}$, where $d < D$. When
$e^d_j = 1$, i.e. the parent $s$ ends at $j$, and
$\Gamma^{d,s}_{i:j}(u)$ will become $\Pi^{d,s}_{i:j}$ with $u = x^{d+1}_j$. 
Then  we have decompositions $\zeta^{d,s}_{i:j} = ({\zeta}^{d,s}_{i:j}(u),u= x^{d+1}_j)$
and $\hat{\zeta}^{d,s}_{i:j} = (\hat{{\zeta}}^{d,s}_{i:j}(u),e^d_j=1,u= x^{d+1}_j)$.
These lead to the factorisation
\begin{eqnarray}
	\label{inside-factorise}
	\Phi[\hat{\zeta}^{d,s}_{i:j}] = \Phi[\hat{{\zeta}}^{d,s}_{i:j}(u)]E^{d,s}_{u,j}
\end{eqnarray}
where the state-ending potential $E^{d,s}_{u,j}$ is activated in
the context $c = (e^d_j=1)$. 
Thus, the inside mass in Equation~\ref{inside-def} can be rewritten as
\begin{eqnarray}
	\label{I-IF}
	\Delta^{d,s}_{i:j} 
		&=& \sum_{u \in S^{d+1}}\sum_{{\zeta}^{d,s}_{i:j}(u)}\Phi[\hat{{\zeta}}^{d,s}_{i:j}(u)]E^{d,s}_{u,j} \nonumber\\
		&=& \sum_{u \in S^{d+1}}E^{d,s}_{u,j}\sum_{{\zeta}^{d,s}_{i:j}(u)}\Phi[\hat{{\zeta}}^{d,s}_{i:j}(u)] \nonumber\\
			&=& \sum_{u \in S^{d+1}}E^{d,s}_{u,j}\alpha^{d,s}_{i:j}(u)
\end{eqnarray}
This equation holds for $d < D$. When $d=D$, we set $\Delta^{D,s}_{i:i}=1$ for all $s \in S^D$ and $i \in [1,T]$,
and when $d=1$, we must ensure that $i=1$ and $j=T$.

\textbf{Remark}: Equations~\ref{IF-detail}, \ref{IF-special}, \ref{IF-special2} and~\ref{I-IF}
specify a \emph{left-right} and \emph{bottom-up} algorithm to compute both 
the inside and asymmetric inside masses.
Initially, at the bottom level $\Delta^{D,s}_{i:i}=1$ for $i \in [1,T]$ and $s \in S^D$. 
A pseudo-code of the dynamic programming algorithm
to compute all the inside and asymmetric inside masses
and the partition function is given in Figure~\ref{alg:inside}.

%----------
\begin{figure}[htb]
\begin{center}
\begin{tabular}{l}\hline
\textbf{Input}: $D,T$, all the potential function values. \\
\textbf{Output}: partition function $Z$; \\
\q\q		$\Delta^{1,s}_{1:T}$, for $s \in S^1$;\\
\q\q		$\Delta^{d,s}_{i:j}$, for $d\in [2,D-1]$, $s\in S^d$ and  $ 1 \le i \le j \le T$; \\
\q\q		$\Delta^{D,s}_{i:i}$ for $s \in S^D$ and $i \in [1,T]$;\\
\q\q		$\alpha^{d,s}_{i:j}(u)$ for $d \in [1,D-1]$, $u \in S^{d+1}$
		and $ 1 \le i \le j \le T$\\\hline
\em/* Initialisation */ \em \\
$\Delta^{D,s}_{i:i}=1$ for all $i \in [1,T]$ and $s \in S^D$\\ 
\em/* At the level d=D-1 */\em \\
\textbf{For} $i=1,2,...,T$ \\
\q	\textbf{For} $j=i,i+1,...,T$ \\
\q\q		Compute $\alpha^{D-1,s}_{i:j}(u)$ using Equation~\ref{IF-special2} \\
\q\q		Compute $\Delta^{D-1,s}_{i:j}$ using Equation~\ref{I-IF} \\
\q	\textbf{EndFor} \\
\textbf{EndFor} \\
\em/* The main recursion loops: bottom-up and forward */\em \\
\textbf{For} $d=D-2,D-3,...,1$\\
\q	\textbf{For} $i=1,2,...,T$\\
\q\q		\textbf{For} $j=i,i+1,...,T$\\
\q\q\q			Compute $\alpha^{d,s}_{i:i}(u)$ using Equation~\ref{IF-special} \textbf{If} $j=i$\\
\q\q\q			Compute $\alpha^{d,s}_{i:j}(u)$ using Equation~\ref{IF-detail} \textbf{If $j > i$}\\
\q\q\q			Compute $\Delta^{d,s}_{i:j}$ using Equation~\ref{I-IF} \textbf{If} $d > 1$\\
\q\q		\textbf{EndFor}\\
\q	\textbf{EndFor}\\
\textbf{EndFor}\\
Compute $Z$ using Equation~\ref{Z-I}.\\
\hline
\end{tabular}
\end{center}
\caption{Computing the set of inside/asymmetric inside masses and
	the partition function.} 
\label{alg:inside}
\end{figure}

%--
\subsection{Computing Outside Masses}
\label{sec:OB}
%--
In this subsection we show how to recursively compute
the symmetric outside mass and the asymmetric outside mass.
We use the same blanket decomposition as in Section~\ref{sec:IF}.
However, this time the view is reversed as we are interested in
quantities outside the blankets. For example,
outside the inner symmetric Markov blanket
in Figure~\ref{fig:DCRF-asym2}, there exists an outer asymmetric blanket 
and another sub-asymmetric blanket on the left.

%------------
\subsubsection{Computing asymmetric outside mass from outside mass}
Let us examine the variables $\underline{\zeta}^{d,s}_{i:j}(u)$
associated with the asymmetric Markov blanket $\Gamma^{d,s}_{i:j}(u)$,
for $d \in [1,D-1]$ and $1 \le i \le j \le T$ (see Definition~\ref{def:asym-variables}).  
For $j < T$, assume that there exists an outer asymmetric Markov blanket
$\Gamma^{d,s}_{i:t}(v)$ for some $v \in S^{d+1}$ and $ t \in [j+1,T]$,
and a symmetric Markov blanket $\Pi^{d+1,v}_{j+1:t}$ right next
to $\Gamma^{d,s}_{i:j}(u)$.  Given these blankets we have the decomposition
$\hat{{\underline{\zeta}}}^{d,s}_{i:j}(u) = 
	(\hat{{\underline{\zeta}}}^{d,s}_{i:t}(v),\hat{\zeta}^{d+1,v}_{j+1:t}, x^{d+1}_j=u$),
which leads to the following factorisation
\begin{eqnarray}
	\Phi[\hat{{\underline{\zeta}}}^{d,s}_{i:j}(u)] =
		\Phi[\hat{{\underline{\zeta}}}^{d,s}_{i:t}(v)]\Phi[\hat{\zeta}^{d+1,v}_{j+1:t}]
		R^{d+1,v}_{j+1:t}A^{d+1,s}_{u,v,j}
\end{eqnarray}
The state transition potential $A^{d+1,s}_{u,v,j}$ is enabled in the context
$c = (e^d_j=0,e^{d+1}_j=1)$, and the state persistence potential
$R^{d+1,v}_{j+1:t}$ in the context $c = (e^{d+1}_j=1,e^{d+1}_{j+1:t-1}=0,e^{d+1}_t=1)$.

In addition, there exists a special case 
where the state $s$ ends at $j$. We have the decomposition 
$\hat{{\underline{\zeta}}}^{d,s}_{i:j}(u) =
		(\hat{\underline{\zeta}}^{d,s}_{i:j},u=x^{d+1}_j)$
and the following factorisation
\begin{eqnarray}
	\Phi[\hat{{\underline{\zeta}}}^{d,s}_{i:j}(u)] =
		\Phi[\hat{\underline{\zeta}}^{d,s}_{i:j}]R^{d,s}_{i:j}E^{d,s}_{u,j}
\end{eqnarray}
The ending potential $E^{d,s}_{u,j}$ appears here because of
the context $c=(e^d_j=1)$, i.e. $s$ ends at $j$.

Now we relax the assumption of $t,v$ and allow
them to receive all possible values, i.e. $t \in [j,T]$
and $v \in S^{d+1}$. Thus we can replace Equation~\ref{OB-def}
by
\begin{eqnarray}
	\lambda^{d,s}_{i:j}(u) &=& \sum_{v \in S^{d+1}}\sum_{t \in [j+1,T]}\sum_{{{\underline{\zeta}}}^{d,s}_{i:t}(v)}
				\sum_{{\zeta}^{d+1,v}_{j+1:t}}\Phi[\hat{{\underline{\zeta}}}^{d,s}_{i:t}(v)]
												\Phi[\hat{\zeta}^{d+1,v}_{j+1:t}]
					R^{d+1,v}_{j+1:t}A^{d+1,s}_{u,v,j} \nonumber\\
				 && \q\q\q\q\q\q\q+ \sum_{{\underline{\zeta}}^{d,s}_{i:j}(u)}\Phi[\hat{\underline{\zeta}}^{d,s}_{i:j}]R^{d,s}_{i:j}E^{d,s}_{u,j} \nonumber\\
	\label{OB-OB}
			&=& \sum_{v \in S^{d+1}}\sum_{t \in [j+1,T]}\lambda^{d,s}_{i:t}(v)\hat{\Delta}^{d+1,v}_{j+1:t}A^{d+1,s}_{u,v,j}
				 + \hat{\Lambda}^{d,s}_{i:j}E^{d,s}_{u,j}
\end{eqnarray}
for $d \in [2,D-2]$, and $1 \le i \le j \le T$. 
Thus, the $\lambda^{d,s}_{i:j}(u)$ can be thought as
a message passed \emph{backward} from $j=T$ to $j=i$. 
Here, the asymmetric outside mass ending at $j$ is computed by using
all the asymmetric outside masses ending at $t$ for $t \in [j+1,T]$.

There are two special cases. At the top level, i.e. $d=1$, 
then $\lambda^{d,s}_{i:j}(u)$ is only defined at $i=1$, and
the second term of the RHS of Equation~\ref{OB-OB} is included only if $i=1,j=T$.
At the second lowest level, i.e. $d=D-1$, we cannot sum over $t$ as in Equation~\ref{OB-OB}
since $\hat{\Delta}^{D,v}_{j+1:t}$ is only defined for $t = j+1$. 
We have the following relation instead
\begin{eqnarray}
	\label{OB-OB2}
	\lambda^{D-1,s}_{i:j}(u)	= \sum_{v \in S^{D}}\lambda^{D-1,s}_{i:j+1}(v)\hat{\Delta}^{D,v}_{j+1:j+1}A^{D,s}_{u,v,j}
				 + \hat{\Lambda}^{D-1,s}_{i:j}E^{D-1,s}_{u,j}
\end{eqnarray}

%------------
\subsubsection{Computing outside mass from asymmetric outside mass}
Given a symmetric Markov blanket $\Pi^{d+1,u}_{i:j}$ for $d \in [1,D-1]$,
assume that there exists
an asymmetric Markov blanket $\Gamma^{d,s}_{t:j}(u)$ at the
parent level $d$, where $t \in [1,i]$. Clearly,
for $t \in [1,i-1]$ there exists some sub-asymmetric Markov blanket 
$\Gamma^{d,s}_{t:i-1}(v)$. See Figure~\ref{fig:DCRF-asym2}
for an illustration. 

Let us consider two cases: $t < i$ and $t=i$.

\textbf{Case 1}. For $t < i$, this enables the decomposition
$\hat{\underline{\zeta}}^{d+1,u}_{i:j} = (\hat{{\underline{\zeta}}}^{d,s}_{t:j}(u),
		\hat{{\zeta}}^{d,s}_{t:i-1}(v),u=x^{d+1}_{i:j})$,
which leads to the following factorisation
\begin{eqnarray}
	\Phi[\hat{\underline{\zeta}}^{d+1,u}_{i:j}] = 
				\Phi[\hat{{\underline{\zeta}}}^{d,s}_{t:j}(u)]
				\Phi[\hat{{\zeta}}^{d,s}_{t:i-1}(v)]
				A^{d,s}_{v,u,i-1}
\end{eqnarray}
The state transition potential $A^{d,s}_{v,u,i-1}$
is activated in the context $c = (e^{d}_{i-1}=0,e^{d+1}_{i-1}=1)$.

\textbf{Case 2}. For $t=i$, the decomposition reduces to
$\hat{\underline{\zeta}}^{d+1,u}_{i:j} = 
	(\hat{{\underline{\zeta}}}^{d,s}_{i:j}(u),u=x^{d+1}_{i:j})$,
which leads to the following factorisation
\begin{eqnarray}
	\Phi[\hat{\underline{\zeta}}^{d+1,u}_{i:j}] = 
		\Phi[\hat{{\underline{\zeta}}}^{d,s}_{i:j}(u)]\pi^{d,s}_{u,i}
\end{eqnarray}
The state-initialisation potential $\pi^{d,s}_{u,i}$ plays the role in the context $c=(e^{d}_{i-1}=1)$

However, these decompositions and factorisations only hold given the assumption
of specific values of $s \in S^{d}$, $v \in S^{d+1}$, and
$t \in [1,i]$. Without further information we have to take
all possibilities into account. Substituting these relations 
into Equation~\ref{outside-def}, we have
\begin{eqnarray}
	\Lambda^{d+1,u}_{i:j} &=& \sum_{s \in S^{d}}\sum_{v \in S^{d+1}}\sum_{t\in [1,i-1]}
				\sum_{\underline{\zeta}^{d,s}_{t:j}(u)}
				\sum_{{\zeta}^{d,s}_{t:i-1}(v)}
				\Phi[\underline{\hat{\zeta}}^{d,s}_{t:j}(u)]
				\Phi[\hat{{\zeta}}^{d,s}_{t:i-1}(v)]
				A^{d+1,s}_{v,u,i-1} +\nonumber\\
			& &	\q\q\q\q\q\q+ \sum_{s \in S^{d}}\sum_{{{\underline{\zeta}}}^{d,s}_{i:j}(u)}
				\Phi[\hat{{\underline{\zeta}}}^{d,s}_{i:j}(u)]\pi^{d,s}_{u,i} \nonumber\\
	\label{O-OB}
			&=& \sum_{s \in S^{d}}\sum_{t\in [1,i-1]}\lambda^{d,s}_{t:j}(u)
				\sum_{v \in S^{d+1}}\alpha^{d,s}_{t:i-1}(v)A^{d+1,s}_{v,u,i-1} 
				+ \sum_{s \in S^{d}}\lambda^{d,s}_{i:j}(u)\pi^{d,s}_{u,i}	
\end{eqnarray}
for $d \in [2,D-2]$. 

There are three special cases. The first is the base case 
where $d=0$ and $\Lambda^{1,s}_{1:T} = 1$ for all $s \in S^1$.
In the second case, for $d=1$, we must fix the
index $t=1$ since the asymmetric inside mass
$\alpha^{d,s}_{t:i-1}$ is only defined at $t=1$.
Also the second term in the RHS is included only if
$i=1$ for the asymmetric outside mass $\lambda^{d,s}_{i:j}(u)$ to make sense. 
In the second case, for $d+1 = D$, we  only have $i=j$.

%-----------
\textbf{Remark}: Equations~\ref{OB-OB}, \ref{OB-OB2} and \ref{O-OB}  show
a recursive \emph{top-down} and \emph{outside-in} approach to compute the symmetric/asymmetric 
outside masses. 
We start from the top with $d=1$ and $\Lambda^{1,s}_{1:T} = 1$ for all
$s \in S^1$ and proceed downward until $d=D$. The pseudo-code
is given in Figure~\ref{alg:outside}. 
Figure~\ref{fig:IF-OB-summary} summarises the quantities
computed in Section~\ref{sec:IF} and \ref{sec:OB}.

Figure~\ref{alg:AIO} summarises the AIO algorithm
for computing all building blocks and the partition function.

\begin{figure}[htb]
\begin{center}
\begin{tabular}{l}\hline
\textbf{Input}: $D,T$, all the potential function values, all inside/asymmetric inside masses. \\
\textbf{Output}: all outside/asymmetric outside masses\\
\hline
Initialise: $\Lambda^{1,s}_{1:T} = 1$,\\
\q\q		$\lambda^{1,s}_{1:T}(u) = E^{1,s}_{u,T}$ for $s \in S^1, u \in S^2$\\
\em /* the main recursive loops: top-down and inside-out */\em \\
\textbf{For} $d=1,2,...,D-1$ \\
\q	\textbf{For} $i = 1,2,...,T$ \\
\q\q		\textbf{For} $j = T,T-1,...,i$ \\
\q\q\q			Compute the asymmetric outside mass $\lambda^{d,s}_{i:j}(u)$ using Equations~\ref{OB-OB},\ref{OB-OB2}\\
\q\q\q			Compute the outside mass $\Lambda^{d,s}_{i:j}$ using Equation~\ref{O-OB}\\
\q\q		\textbf{EndFor} \\
\q	\textbf{EndFor} \\
\textbf{EndFor} \\
\hline
\end{tabular}
\end{center}
\caption{Computing the set of outside/asymmetric outside masses.} 
\label{alg:outside}
\end{figure}

%\begin{figure}[htb]
%\begin{center}
%\begin{tabular}{cc}
%\includegraphics[width=0.45\linewidth]{OB-mass.eps} &
%\includegraphics[width=0.45\linewidth]{O-bw.eps} \\
%(a) & (b)
%\end{tabular}
%\end{center}
%\caption{(a) Computing outside masses from asymmetric outside,
%	and (b) asymmetric outside from asymmetric outside.}
%\label{fig:O-OB-OB}
%\end{figure}

\begin{figure}[htb]
\begin{center}
\begin{tabular}{l}\hline
$\bullet$ $\Delta^{1,s}_{1:T},\Lambda^{1,s}_{1:T} \mbox{ for } s \in S^1$ \\
$\bullet$ $\Delta^{d,s}_{i:j}, \Lambda^{d,s}_{i:j} \mbox{ for } d \in [2,D-1], s \in S^d, 
		1 \le i \le j \le T$ \\
$\bullet$ $\Delta^{D,s}_{i:i},\Lambda^{D,s}_{i:i} \mbox{ for } i \in [1,T], s \in S^D $\\
$\bullet$ $\alpha^{d,s}_{1:j}(u), \lambda^{d,s}_{1:j}(u) 
		\mbox{ for } d =1, s \in S^1, u \in S^{2},
		j \in [1,T]$ \\
$\bullet$ $\alpha^{d,s}_{i:j}(u), \lambda^{d,s}_{i:j}(u) 
		\mbox{ for } d \in [2,D-1], s \in S^d, u \in S^{d+1},
		1 \le i \le j \le T$ \\
\hline
\end{tabular}
\end{center}
\caption{Summary of basic building blocks computed in Section~\ref{sec:IF} and \ref{sec:OB}.}
\label{fig:IF-OB-summary}
\end{figure}

%---
\begin{figure}[htb]
\begin{center}
\begin{tabular}{l}\hline
\textbf{Input}: $D,T$, all the potential function values \\
\textbf{Output}: all building blocks and partition function\\
\hline
Compute all inside/asymmetric inside masses using the algorithm in Figure~\ref{alg:inside} \\
Compute all outside/asymmetric outside masses using the algorithm in Figure~\ref{alg:outside} \\
\hline
\end{tabular}
\end{center}
\caption{The AIO algorithm.} 
\label{alg:AIO}
\end{figure}

%--
\section{The Generalised Viterbi Algorithm}
\label{sec:MAP}
By definition the MAP assignment is the maximiser of
the conditional distribution given an observation sequence $z$
\begin{eqnarray}
	\zeta^{MAP} &=& \arg\max_{\zeta} \Pr(\zeta|z) \nonumber\\
				&=&  \arg\max_{\zeta} \Phi[\zeta,z]
\end{eqnarray}
For clarity, let us drop the notation $z$ and assume that it
is implicitly there.

The process of computing the MAP assignment
is very similar to that of computing the partition function.
This similarity comes from the relation between the
sum-product and max-product algorithm (a generalisation of the Viterbi algorithm)
of \cite{Pearl88}, and from the
fact that inside/asymmetric inside procedures described in Section~\ref{sec:IF} 
are essentially a sum-product version. What we need
to do is to just convert all the summations into
corresponding maximisations. The algorithm
is a two-step procedure:
\begin{itemize}
\item In the first step the maximum joint potential is computed and
		local maximum states and ending indicators are saved along the way.
		These states and ending indicators are maintained in
		a \emph{bookkeeper}.
\item In the second step we decode the best assignment
		by \emph{backtracking} through saved local maximum states.
\end{itemize}

We make use of the contextual decompositions
and factorisations from Section~\ref{sec:IF}.

%----------------------
\subsubsection*{Notations}
%--
This section, with some abuse, uses some
slight modifications to the notations used in the rest of the paper.
See Table~\ref{tab:hcrf-notation-MAP} for reference.

%-----------------------------
\begin{table}[htb]
\begin{center}
\begin{tabular}{rl} 
\hline
\textbf{Notation}				& \textbf{Description}\\\hline\hline
$\Delta^{\max,d,s}_{i:j}$		& The optimal potential function
									of the subset of variables $\zeta^{d,s}_{i:j}$\\
$\hat{\Delta}^{\max,d,s}_{i:j}$		& The `full' version of  $\Delta^{\max,d,s}_{i:j}$\\
$\alpha^{\max,d,s}_{i:j}(u)$	& The optimal potential function
									of the subset of variables $\zeta^{d,s}_{i:j}(u)$\\
$\Delta^{\arg,d,s}_{i:j}$		& The optimal child $u^{d+1}_j$ of $s$\\
$\alpha^{\arg,d,s}_{i:j}(u)$	& The optimal child $v^{d+1}_{t-1}$ that
										transits to $u^{d+1}_{t:j}$
										and the	time index $t$.\\
$\mathcal{I}^{d}$				& The set of optimal `segments' at each level $d$.\\
\hline
\end{tabular}
\end{center}
\caption{Notations used in this section.}
\label{tab:hcrf-notation-MAP}
\end{table}
%-----------------------------

We now describe the first step.

%-----------------------------
\subsection{Computing the Maximum Joint Potential, Maximal States and Time Indices}
%--------
As $\Phi[\zeta] = \Phi[\hat{\zeta}^{1,s}_{1:T}]R^{1,s}_{1:T}$ for $s \in S^1$ we have
\begin{eqnarray}
	\label{max-top-factorise}
	\max_{\zeta} \Phi[\zeta] = \max_{s \in S^1}R^{1,s}_{1:T}\max_{\zeta^{1,s}_{1:T}}\Phi[\hat{\zeta}^{1,s}_{1:T}]
\end{eqnarray}
Now, for a sub-assignment $\zeta^{d,s}_{i:j}$ for $1 \in [1,D-1]$, 
Equation~\ref{inside-factorise} leads to
\begin{eqnarray}
	\label{max-inside-factorise}
	\max_{\zeta^{d,s}_{i:j}}\Phi[\hat{\zeta}^{d,s}_{i:j}]
		= \max_{u \in S^{d+1}}E^{d,s}_{u,j}\max_{{\zeta}^{d,s}_{i:j}(u)}\Phi[{\hat{\zeta}}^{d,s}_{i:j}(u)]
\end{eqnarray}

With some slight abuse of notation we introduce
$\Delta^{\max,d,s}_{i:j}$ as the optimal potential function	of the subset of 
variables $\zeta^{d,s}_{i:j}$, and $\alpha^{\max,d,s}_{i:j}(u)$ as the optimal potential function
of the subset of variables $\zeta^{d,s}_{i:j}(u)$.

\begin{definition}
	We define $\Delta^{\max,d,s}_{i:j}$ and $\alpha^{\max,d,s}_{i:j}(u)$ as follows
	\begin{eqnarray}
		\Delta^{\max,d,s}_{i:j} &=&  \max_{\zeta^{d,s}_{i:j}}\Phi[\hat{\zeta}^{d,s}_{i:j}] \\
		\hat{\Delta}^{\max,d,s}_{i:j} &=& \Delta^{\max,d,s}_{i:j}R^{d,s}_{i:j}\\
		\alpha^{\max,d,s}_{i:j}(u) &=& \max_{{\zeta}^{d,s}_{i:j}(u)} 
												\Phi[{\hat{\zeta}}^{d,s}_{i:j}(u)]
	\end{eqnarray}
\end{definition}

The Equations~\ref{max-top-factorise} and \ref{max-inside-factorise} can be 
rewritten more compactly as
\begin{eqnarray}
	\label{max-top-I}
	\Phi[\zeta^{MAP}] &=& \max_{s \in S^1}\hat{\Delta}^{\max,1,s}_{1:T} \\
	\label{max-I-IF}
	\Delta^{\max,d,s}_{i:j} &=& \max_{u \in S^{d+1}}E^{d,s}_{u,j}\alpha^{\max,d,s}_{i:j}(u)
\end{eqnarray}
for $d \in [1,D-1]$. When $d=D$, we simply set $\Delta^{\max,D,s}_{i:i}=1$
for all $s \in S^D$ and $i \in [1,T]$.

From the factorisation in Equation~\ref{IF-factorise} and \ref{IF-factorise2}, we have
\begin{eqnarray}
	\label{max-IF-factorise}
	\max_{{{\zeta}}^{d,s}_{i:j}(u)}\Phi[\hat{{\zeta}}^{d,s}_{i:j}(u)]
		 &=& \max\left\{\left(\max_{v\in S^{d+1}}\max_{t\in[i+1,j]}
		 	R^{d+1,u}_{t:j}
		 	A^{d+1,s}_{v,u,t-1} 
		 	\max_{{{\zeta}}^{d,s}_{i:t-1}(v)}\Phi[\hat{{\zeta}}^{d,s}_{i:t-1}(v)] \times \right.\right. \nonumber\\
		 & &\q	\times \left.\max_{\zeta^{d+1,u}_{t:j}}\Phi[\hat{\zeta}^{d+1,u}_{t:j}]\right);
		 			\left.\left(R^{d+1,u}_{i:j}\max_{{\zeta}^{d+1,u}_{i:j}}\pi^{d,s}_{u,i}\Phi[\hat{\zeta}^{d+1,u}_{i:j}]\right)\right\}
\end{eqnarray}
and
\begin{eqnarray}
	\label{max-IF-detail}
	\alpha^{\max,d,s}_{i:j}(u) &=& \max\Bigg\{\left(\max_{v \in S^{d+1}}\max_{t \in [i+1,j]}\alpha^{\max,d,s}_{i:t-1}(v)
									\hat{\Delta}^{\max,d+1,u}_{t:j}A^{d,s}_{v,u,t-1}\right); \nonumber \\
								&&	\q\q\q\q\q\q\q\left(\hat{\Delta}^{\max,d+1,u}_{i:j}\pi^{d+1,s}_{u,i}\right)\Bigg\}					
\end{eqnarray}
for $d \in [1,D-2]$ and $i < j$. For $d=D-1$, we cannot scan the index $t$ in the interval
$[i+1,j]$ because the maximum inside ${\Delta}^{\max,D,u}_{t:j}$
is only defined at $t=j$. We have the following instead
\begin{eqnarray}
	\label{max-IF-special}
	\alpha^{\max,D-1,s}_{i:j}(u) = \max_{v \in S^{D}}\alpha^{\max,D-1,s}_{i:j-1}(v)
				\hat{\Delta}^{\max,D,u}_{j:j}A^{D,s}_{v,u,j-1}
\end{eqnarray}

There is a base case for $i=j$, where the context $c = (e^{d}_{i-1}=1)$ is active, then
\begin{eqnarray}
	\label{max-IF-special2}
	\alpha^{\max,d,s}_{i:i}(u) = \hat{\Delta}^{\max,d+1,u}_{i:i}\pi^{d,s}_{u,i}
\end{eqnarray}

Of course, what we are really interested in is not
the maximum joint potentials but the optimal states and time indices
(or ending indicators). We need some bookkeepers
to hold these quantities along the way.
With some abuse of notation
let us introduce the symmetric inside bookkeeper $\Delta^{\arg,d,s}_{i:j}$
associated with Equation~\ref{max-I-IF}, 
and the asymmetric inside bookkeeper $\alpha^{\arg,d,s}_{i:j}(u)$
associated with Equations~\ref{max-IF-detail}, \ref{max-IF-special}
and~\ref{max-IF-special2}.

\begin{definition}
	We define the symmetric inside bookkeeper $\Delta^{\arg,d,s}_{i:j}$ as follows
	\begin{eqnarray}
		\label{arg-I-IF}
		\Delta^{\arg,d,s}_{i:j} &=& u^* = {\arg\max}_{u \in S^{d+1}}E^{d,s}_{u,j}\alpha^{\max,d,s}_{i:j}(u)
	\end{eqnarray}
	Similarly, we define the asymmetric inside bookkeeper 
	$\alpha^{\arg,d,s}_{i:j}(u)$ associated with Equation~\ref{max-IF-detail}
	for $d \in [1,D-2]$ as
	\begin{eqnarray}
		\label{arg-IF-detail}
		\alpha^{\arg,d,s}_{i:j}(u) = (v,t)^* = {\arg\max}_{t \in [i+1,j],v \in S^{d+1}}\alpha^{\max,d,s}_{i:t-1}(v)
						\hat{\Delta}^{\max,d+1,u}_{t:j}A^{d,s}_{v,u,t-1}
	\end{eqnarray}
	if $\max_{v \in S^{d+1},t \in [i+1,j]}\alpha^{\max,d,s}_{i:t-1}(v)\hat{\Delta}^{\max,d+1,u}_{t:j}A^{d,s}_{v,u,t-1} 
		> \hat{\Delta}^{\max,d+1,u}_{i:j}\pi^{d+1,s}_{u,i}$ and $i < j$;
	and	
	\begin{eqnarray}
		\alpha^{\arg,d,s}_{i:j}(u) = \mbox{ undefined }
	\end{eqnarray}
	otherwise.	
	For $d=D-1$, the $\alpha^{\arg,d,s}_{i:j}(u)$ is associated with 
	Equation~\ref{max-IF-special}
	\begin{eqnarray}
		\label{arg-IF-special}		
		\alpha^{\arg,D-1,s}_{i:j}(u) = {\arg\max}_{v \in S^{D}}\alpha^{\max,d,s}_{i:j-1}(v)
					\hat{\Delta}^{\max,D,u}_{j:j}A^{d,s}_{v,u,j-1}
	\end{eqnarray}
\end{definition}

The Equations~\ref{max-top-I},\ref{max-I-IF},\ref{max-IF-detail},\ref{max-IF-special} 
and \ref{max-IF-special2} 
provide a recursive procedure to compute maximum joint potential 
in a bottom-up and left-right manner. Initially we just
set $\Delta^{\max,D,s}_{i:i}=1$ for all $s \in S^D$ and $i \in [1,T]$.
The procedure is summarised in Figure~\ref{alg:inside-max}.
%--------

%----------
\begin{figure}[htb]
\begin{center}
\begin{tabular}{l}\hline
\textbf{Input}: $D,T$, all the potential function values. \\
\textbf{Output}: the bookkeepers; \\
\q\q		$\Delta^{\arg,1,s}_{1:T}$, for $s \in S^1$ and  $ 1 \le i \le j \le T$;\\
\q\q		$\Delta^{\arg,d,s}_{i:j}$, for $d\in [2,D-1]$, $s\in S^d$; \\
\q\q		$\Delta^{\arg,D,s}_{i:i}$ for $s \in S^D$ and $i \in [1,T]$;\\
\q\q		$\alpha^{\arg,d,s}_{i:j}(u)$ for $d \in [1,D-1]$, $u \in S^{d+1}$
		and $ 1 \le i \le j \le T$\\\hline
\em/* Initialisation */ \em \\
$\Delta^{\max,D,s}_{i:i}=1$ for all $i \in [1,T]$ and $s \in S^D$\\ 
\em/* At the level d=D-1 */\em \\
\textbf{For} $i=1,2,...,T$ \\
\q	\textbf{For} $j=i,i+1,...,T$ \\
\q\q		Compute $\alpha^{\max,D-1,s}_{i:j}(u)$ using Equation~\ref{max-IF-special} and \\
\q\q\q			$\alpha^{\arg,D-1,s}_{i:j}(u)$ using Equation~\ref{arg-IF-special}\\
\q\q		Compute $\Delta^{\max,D-1,s}_{i:j}$ using Equation~\ref{max-I-IF} and \\
\q\q\q			$\Delta^{\arg,D-1,s}_{i:j}$ using Equation~\ref{arg-I-IF}\\
\q	\textbf{EndFor} \\
\textbf{EndFor} \\
\em/* The main recursion loops: bottom-up and forward */\em \\
\textbf{For} $d=D-2,D-3,...,1$\\
\q	\textbf{For} $i=1,2,...,T$\\
\q\q		\textbf{For} $j=i,i+1,...,T$\\
\q\q\q			\textbf{If} $j=i$ \\
\q\q\q\q				Compute $\alpha^{\max,d,s}_{i:i}(u)$ using Equation~\ref{max-IF-special2}\\
\q\q\q			\textbf{Else}\\
\q\q\q\q			Compute $\alpha^{\max,d,s}_{i:j}(u)$ using Equation~\ref{max-IF-detail} and\\
\q\q\q\q\q				$\alpha^{\arg,d,s}_{i:i}(u)$ using Equation~\ref{arg-IF-detail}\\
\q\q\q			\textbf{EndIf} \\
\q\q\q			\textbf{If} $d > 1$\\
\q\q\q\q			Compute $\Delta^{\max,d,s}_{i:j}$ using Equation~\ref{max-I-IF} and\\
\q\q\q\q\q				$\Delta^{\arg,d,s}_{i:j}$ using Equation~\ref{arg-I-IF}\\
\q\q\q			\textbf{EndIf} \\
\q\q		\textbf{EndFor}\\
\q	\textbf{EndFor}\\
\textbf{EndFor}\\
Compute $\Delta^{\max,1,s}_{1:T}$ using Equation~\ref{max-I-IF} and\\
\q		$\Delta^{\arg,1,s}_{1:T}$ using Equation~\ref{arg-I-IF}\\
\hline
\end{tabular}
\end{center}
\caption{Computing the bookkeepers.} 
\label{alg:inside-max}
\end{figure}

\subsection{Decoding the MAP Assignment}
The proceeding of the backtracking process is opposite
to that of the max-product. Specifically, we start from the root and proceed
in a \emph{top-down} and \emph{right-left} manner. The goal is to identify the right-most 
segment at each level. Formally, a segment is a triple $(s,i,j)$ where
$s$ is the segment label, and $i$ and $j$ are start and end time indices, respectively.
From the maximum inside $\Delta^{\max,d,s}_{i:j}$ at level $d$, we identify the best child 
$u$ and its ending time $j$ from Equation~\ref{max-I-IF}. 
This gives rise to the maximum asymmetric inside $\alpha^{\max,d,s}_{i:j}(u)$.
Then we seek for the best child $v$ that transits to $u$ under the
same parent $s$ using Equation~\ref{max-IF-detail}. Since the
starting time $t$ for $u$ has been identified the ending time
for $v$ is $t-1$. We now have a right-most segment $(u,t,j)$ at level $d+1$.
The procedure is repeated until we reach the starting time $i$ of the parent $s$.
The backtracking algorithm is summarised in  Figure~\ref{alg:backtrack-map}.

%--------------------
\begin{figure}[htb]
\begin{center}
\begin{tabular}{l}
\hline
\textbf{Input}: $D,T$, all the filled bookkeepers. \\
\textbf{Output}: the optimal assignment $\zeta^{MAP}$\\
\hline
$s^* = {\arg\max}_{s \in S^1}\hat{\Delta}^{\max,1,s}_{1:T}$ \\
Initialise triple buckets $\mathcal{I}^{1}=\{(s^*,1,T)\}$ and $\mathcal{I}^{d}=\{\}$ for $d\in [2,D]$\\
\textbf{For} $d=1,2,...,D-1$ \\
\q	\textbf{For} each triple $(s^*,i,j)$ in $\mathcal{I}^d$ \\
\q\q		Let $u^* = \Delta^{\arg,d,s^*}_{i:j}$\\	
\q\q		\textbf{For} $i \le j$ \\
\q\q\q			\textbf{If} $\alpha^{\arg,d,s^*}_{i:j}(u^*)$ is defined \textbf{Then}\\
\q\q\q\q			$(t^*,v^*) = \alpha^{\arg,d,s^*}_{i:j}(u^*)$ \\
\q\q\q\q			Add the triple $(v^*,t^*,j)$ to $\mathcal{I}^{d+1}$ and Set $j = t^*-1$ and $u^* = v^*$\\
\q\q\q			\textbf{Else} \\
\q\q\q\q			Add the triple $(u^*,i,j)$ to $\mathcal{I}^{d+1}$ and Break this loop\\
\q\q\q			\textbf{EndIf} \\
\q\q		\textbf{EndFor} \\
%\q\q		$(u^*) = \Delta^{\arg,d,s^*}_{i:j}$ \\
%\q\q		\textbf{If} $\alpha^{\arg,d,s^*}_{i:j}(u^*)$ is undefined \textbf{Then}\\
\q	\textbf{EndFor} \\
\textbf{EndFor} \\
For each stored triple $(s^*,i,j)$ in the bucket $\mathcal{I}^d$, for $d \in [1,D]$, \\
create a corresponding set of variables $(x^d_{i:j}=s^*,e^d_{i-1}=1,e^d_{j}=1,e^d_{i:j-1}=0)$.\\
The joining of these sets is the optimal assignment $\zeta^{MAP}$\\
\hline
\end{tabular}
\end{center}
\caption{Backtracking for optimal assignment (nested Markov blankets).} 
\label{alg:backtrack-map}
\end{figure}
%----------

Finally, the generalised Viterbi algorithm is given in
Figure~\ref{alg:generalised-Viterbi}.

%----------
\begin{figure}[htb]
\begin{center}
\begin{tabular}{l}\hline
\textbf{Input}: $D,T$, all the potential function values. \\
\textbf{Output}: the optimal assignment $\zeta^{MAP}$\\
\hline
Run the bottom-up discrete optimisation procedure described in Figure~\ref{alg:inside-max}.\\
Run the top-down backtracking procedure described in Figure~\ref{alg:backtrack-map}.\\
\hline
\end{tabular}
\end{center}
\caption{The generalised Viterbi algorithm.} 
\label{alg:generalised-Viterbi}
\end{figure}

%\rem{
%--------
\subsection*{Working in log-space to avoid numerical overflow}
With long sequence and complex topology we may run into the problem
of numerical overflow, i.e. when the numerical value of the maximum joint potential
is beyond the number representation of the machine. To avoid this, we can
work in the log-space instead, using the monotonic property of
the log function. The equations in the log-space
are summarised in Table~\ref{tab:map-log}.

%--------------------
\begin{table}[htb]
\begin{center}
\begin{tabular}{|l|c|}
\hline
Log-space equations & Equations. \\\hline\hline
$\log \Delta^{\max,d,s}_{i:j} = \max_{u \in S^{d+1}}\{\log E^{d,s}_{u,j} + \log \alpha^{\max,d,s}_{i:j}(u)\}$
	& Eq.~\ref{max-I-IF} \\\hline
$\log \alpha^{\max,d,s}_{i:j}(u) = \max\left\{ \max_{t \in [i+1,j]}\max_{v \in S^{d+1}}
		\{\log\alpha^{\max,d,s}_{i:t-1}(v) +  \right.$ & \\
		 $ + \log\hat{\Delta}^{\max,d+1,u}_{t:j} + \log A^{d,s}_{v,u,t-1}\};
		 \left.\log \hat{\Delta}^{\max,d+1,u}_{i:j} + \log \pi^{d+1,s}_{u,i}\right\}$					
	& Eq.~\ref{max-IF-detail} \\\hline
$\log \alpha^{\max,D-1,s}_{i:j}(u) = \max_{v \in S^{D}}\{\log\alpha^{\max,D-1,s}_{i:j-1}(v)+$ & \\
								$\q\q\q\q+\log\hat{\Delta}^{\max,D,u}_{j:j}+ \log A^{D,s}_{v,u,j-1}\}$
	& Eq.~\ref{max-IF-special} \\\hline
$\log\alpha^{\max,d,s}_{i:i}(u) = \log \hat{\Delta}^{\max,d+1,u}_{i:i} + \log \pi^{d,s}_{u,i}$
	& Eq.~\ref{max-IF-special2} \\\hline
\end{tabular}
\end{center}
\caption{MAP equations in the log-space.} 
\label{tab:map-log}
\end{table}
%----------
%}
%%}
%

%----------------------------
\section{Parameter Estimation}
\label{sec:learning}
In this section, we tackle the problem of parameter estimation by maximising
the (conditional) data likelihood. Typically we need some parametric form 
to be defined for a particular problem and we need 
some numerical method to do the optimisation task.

Here we employ the log-linear parameterisation, which is
commonly used in the CRF setting. 
Recall from Section~\ref{sec:bgr-crf} that estimating parameters
of the log-linear models using gradient-based methods requires
the computation of feature expectation, or expected sufficient
statistics (ESS). For our {\HCRF}s we need
to compute four types of ESS corresponding to
the state-persistence, state-transition, state-initialisation
and state-ending. 

%----------------------------------
\subsection{Log-Linear Parameterisation}
\label{sec:param}

In our {\HCRF} setting there is a feature vector
$\mathbf{f}^d_{\sigma}(\sigma,z)$ associated with each type
of contextual clique $\sigma$, in that
$\phi(\sigma^{d},z) = \exp(\w_{\sigma^d}^{\top}\mathbf{f}^d_{\sigma}(\sigma,z))$.
Thus, the features are active only in the context in which
the corresponding contextual cliques appear.

For the state-persistence contextual clique, the features
incorporate \em state-duration\em, start time $i$ and
end time $j$ of the state. Other feature types incorporate
the time index in which the features are triggered.
Specifically, 
\begin{eqnarray}
	R^{d,s,z}_{i:j} &=& \exp(\w_{\sigma^{persist,d}}^{\top} \mathbf{f}^{d,s}_{\sigma^{persist}}(i,j,z)) \\
	A^{d,s,z}_{u,v,i} &=& \exp(\w_{\sigma^{transit,d}}^{\top} \mathbf{f}^{d,s}_{\sigma^{transit},u,v}(i,z) \\ 
	\pi^{d,s,z}_{u,i} &=& \exp(\w_{\sigma^{init,d}}^{\top} \mathbf{f}^{d,s}_{\sigma^{init},u}(i,z) \\ 
	E^{d,s,z}_{u,i} &=& \exp(\w_{\sigma^{end,d}}^{\top} \mathbf{f}^{d,s}_{\sigma^{end},u}(i,z)
\end{eqnarray}

Denote by $\mathbf{F}^d_{\sigma}(\zeta,z)$ 
the global feature, which is
the sum of all active features  $\mathbf{f}^d_{\sigma}(z)$ at level $d$
in the duration $[1,T]$ for a given assignment of $\zeta$ and a
clique type $\sigma$. 
Recall that $\tau^d = \{i_k\}_{k=1}^m$ is the set of ending time indices
(i.e. $e^d_{i_k} = 1$). The four feature types are given in
Equations~\ref{glob-feature-persist}-\ref{glob-feature-exit}.
\begin{eqnarray}
	\label{glob-feature-persist}
	\mathbf{F}^{d,s}_{\sigma^{persist}}(\zeta,z) &=& \mathbf{f}^{d,s}_{\sigma^{persist}}(1,i_{1},z) + \sum_{i_k \in \tau^d, k > 1}\mathbf{f}^{d,s}_{\sigma^{persist}}(i_k+1,i_{k+1},z) \\
	\label{glob-feature-transit}
	\mathbf{F}^{d,s}_{\sigma^{transit},u,v}(\zeta,z) &=& \sum_{i_k\notin \tau^{d-1},i_k\in\tau^{d}}\mathbf{f}^{d,s}_{\sigma^{transit},u,v}(i_k,z) \\
	\label{glob-feature-init}
	\mathbf{F}^{d,s}_{\sigma^{init},u}(\zeta,z) &=& \mathbf{f}^{d,s}_{\sigma^{init},u,v}(1,z) + \sum_{i_k\in\tau^d}\mathbf{f}^{d,s}_{\sigma^{init},u,v}(i_k+1,z) \\
	\label{glob-feature-exit}
	\mathbf{F}^{d,s}_{\sigma^{end},u}(\zeta,z) &=& \sum_{i_k\in\tau^d}\mathbf{f}^{d,s}_{\sigma^{end},u,v}(i,z)
\end{eqnarray}

Substituting the global features into potentials in Equation.~\ref{joint-potential}
and \ref{model-dis} we obtain the following log-linear model:
\begin{eqnarray}
	\label{model-def2}
	\Pr(\zeta|z) = \frac{1}{Z(z)}\exp(\sum_{c \in C}\w_{\sigma^{c}}^{\top}\mathbf{F}_{\sigma^{c}}(\zeta,z))
\end{eqnarray}
where $C = \{persist,transit,init,exit\}$.

Again, for clarity of presentation we will drop the notion of
$z$ but implicitly assume that it is still in the each quantity.

%--------------------------
%
%\subsection{Feature Expectation}

\subsection{ESS for State-Persistence Features} 
%\textbf{ESS for state-persistence features:} 
%\\
%--------
Recall from Section~\ref{sec:param} that
the feature function for the state-persistence
$\mathbf{f}^{d,s}_{\sigma^{persist}}(i,j)$ 
is active only in the context where $\Pi^{d,s}_{i:j}\in\zeta$.
Thus, Equation~\ref{glob-feature-persist} can be rewritten as
\begin{eqnarray}
	\label{state-ESS1}
	\mathbf{F}^{d,s}_{\sigma^{persist}}(\zeta) = \sum_{i\in[1,T]}\sum_{j\in[i,T]}\mathbf{f}^{d,s}_{\sigma^{persist}}(i,j)\delta[\Pi^{d,s}_{i:j}\in\zeta]
\end{eqnarray}
The indicator function in the RHS ensures that 
the feature $\mathbf{f}^{d,s}_{\sigma^{persist}}(i,j)$ is only active
if there exists a symmetric Markov blanket 
$\Pi^{d,s}_{i:j}$ in the assignment of $\zeta$. Consider the following expectation
\begin{eqnarray}
	\mathbb{E}[\mathbf{f}^{d,s}_{\sigma^{persist}}(i,j)\delta[\Pi^{d,s}_{i:j}\in\zeta]] 
		&=&\sum_{\zeta}\Pr(\zeta)\mathbf{f}^{d,s}_{\sigma^{persist}}(i,j)\delta[\Pi^{d,s}_{i:j}\in\zeta]\\
		&=&\frac{1}{Z}\sum_{\zeta}\Phi[\zeta]\mathbf{f}^{d,s}_{\sigma^{persist}}(i,j)\delta[\Pi^{d,s}_{i:j}\in\zeta]
\end{eqnarray}
Using the factorisation in Equation~\ref{factorise} we can rewrite
\begin{eqnarray}
	\mathbb{E}[\mathbf{f}^{d,s}_{\sigma^{persist}}(i,j)\delta[\Pi^{d,s}_{i:j}\in\zeta]]
		= \frac{1}{Z}\sum_{\zeta}\Phi[\hat{\zeta}^{d,s}_{i:j}]\Phi[\hat{\underline{\zeta}}^{d,s}_{i:j}]R^{d,s}_{i:j}\mathbf{f}^{d,s}_{\sigma^{persist}}(i,j)\delta[\Pi^{d,s}_{i:j}\in\zeta]
\end{eqnarray}
Note that the elements inside the sum of the RHS are only non-zeros
for those assignment of $\zeta$ that respect
the persistent state $s^d_{i:j}$ and the factorisation in 
Equation~\ref{factorise}, i.e. 
$\zeta = (\zeta^{d,s}_{i:j}, \underline{\zeta}^{d,s}_{i:j}, \Pi^{d,s}_{i:j})$.
Thus, the equation can be simplified to
\begin{eqnarray}
	\mathbb{E}[\mathbf{f}^{d,s}_{\sigma^{persist}}(i,j)\delta[\Pi^{d,s}_{i:j}\in\zeta]]
		&=& \frac{1}{Z}\sum_{\zeta^{d,s}_{i:j}}\sum_{\underline{\zeta}^{d,s}_{i:j}}
			\Phi[\hat{\zeta}^{d,s}_{i:j}]\Phi[\hat{\underline{\zeta}}^{d,s}_{i:j}]R^{d,s}_{i:j}
			\mathbf{f}^{d,s}_{\sigma^{persist}}(i,j) \\
		&=& \frac{1}{Z}\Delta^{d,s}_{i:j}\Lambda^{d,s}_{i:j}R^{d,s}_{i:j}\mathbf{f}^{d,s}_{\sigma^{persist}}(i,j)
\end{eqnarray}

Using Equation~\ref{state-ESS1} we obtain the ESS for the state-persistence features
\begin{eqnarray}
	\label{data-ESS}
	\mathbb{E}[F^{d,s}_k(\zeta)] &=&
		\sum_{i \in [1,T]}\sum_{j \in [i,T]} \mathbb{E}[\mathbf{f}^{d,s}_{\sigma^{persist}}(i,j)\delta[\Pi^{d,s}_{i:j}\in\zeta]] \nonumber\\
		&=& \frac{1}{Z}\sum_{i \in [1,T]}\sum_{j \in [i,T]}\Delta^{d,s}_{i:j}\Lambda^{d,s}_{i:j}R^{d,s}_{i:j}\mathbf{f}^{d,s}_{\sigma^{persist}}(i,j)
\end{eqnarray}

There are two special cases: (1) when $d=1$, we do not sum over $i,j$ but fix
$i=1,j=T$, and (2) when $d=D$ then we keep $j=i$.
%-------------------------
\subsection{ESS for Transition Features}
%\\\\
%\textbf{ESS for transition features:} 
%\\
%---
Recall that in Section~\ref{sec:param} we define
$\mathbf{f}^{d,s}_{\sigma^{transit},u,v}(t)$ as a function that 
is active in the context $c^{transit}= (e^{d-1}_t=0,e^{d}_t=1)$, 
in which the child state $u^{d}$ finishes its job at time $t$ and 
transits to the child state $v^{d}$ under the
same parent $s^{d-1}$ (that is $s^{d-1}$ is still running).
Thus Equation~\ref{glob-feature-transit} can be rewritten as
\begin{eqnarray}
	\label{ESS-transit1}
	\mathbf{F}^{d,s}_{\sigma^{transit},u,v}(\zeta) = \sum_{t\in [1,T-1]}\mathbf{f}^{d,s}_{\sigma^{transit},u,v}(t)\delta[c^{transit} \in \zeta]
\end{eqnarray}

We now consider the following expectation
\begin{eqnarray}
	\mathbb{E}[\mathbf{f}^{d,s}_{\sigma^{transit},u,v}(t)\delta[c^{transit} \in \zeta]] &=& \sum_{\zeta}\Pr(\zeta)\mathbf{f}^{d,s}_{\sigma^{transit},u,v}(t)\delta[c^{transit} \in \zeta]\\
	\label{ESS-transit2}
		&=& \frac{1}{Z}\sum_{\zeta}\Phi[\zeta]\mathbf{f}^{d,s}_{\sigma^{transit},u,v}(t)\delta[c^{transit} \in \zeta]
\end{eqnarray}

Assume that the parent $s$ starts at $i$. Since
$e^{d}_t = 1$, the child $v$ must starts at $t+1$ and ends some time
later at $j \ge t+1$. We have the following decomposition of the
configuration $\zeta$ that respects this assumption
\begin{eqnarray}
	\zeta = (\hat{{\underline{\zeta}}}^{d-1,s}_{i:j}(v),
			\hat{{\zeta}}^{d-1,s}_{i:t}(u),
			\hat{\zeta}^{d,v}_{t+1:j})
\end{eqnarray}
and the following factorisation of the joint potential
\begin{eqnarray}
	\Phi[\zeta] = \Phi[\hat{{\underline{\zeta}}}^{d-1,s}_{i:j}(v)]
			\Phi[\hat{{\zeta}}^{d-1,s}_{i:t}(u)]
			\Phi[\hat{\zeta}^{d,v}_{t+1:j}]
			R^{d,v}_{t+1:j}A^{d,s}_{u,v,t}
\end{eqnarray}
The state persistent potential $R^{d,v}_{t+1:j}$ is enabled
in the context $c = (e^{d}_{t}=1,e^{d}_{t+1:j-1}=0,e^{d}_j=1)$
and the state transition potential $A^{d,s}_{u,v,t}$
in the context $c^{transit}$.

Substituting this factorisation into the RHS of Equation~\ref{ESS-transit2} gives us
\begin{eqnarray}
	\frac{1}{Z}\sum_{i\in [1,t]}\sum_{j \in [t+1,T]}
			\sum_{{{\zeta}}^{d-1,s}_{i:t}(u)}
			\sum_{{{\underline{\zeta}}}^{d-1,s}_{i:j}(v)}
			\sum_{\zeta^{d,v}_{t+1:j}}
			\Phi[\hat{{\underline{\zeta}}}^{d-1,s}_{i:j}(v)]
			\Phi[\hat{{\zeta}}^{d-1,s}_{i:t}(u)]
			\Phi[\hat{\zeta}^{d,v}_{t+1:j}]R^{d,v}_{t+1:j}A^{d,s}_{u,v,t}
			\mathbf{f}^{d,s}_{\sigma^{transit},u,v}(t) \nonumber
\end{eqnarray}
which can be simplified to
\begin{eqnarray}
	\label{ESS-transit3}
		\frac{1}{Z}\sum_{i\in [1,t]}\sum_{j \in [t+1,T]}
			\lambda^{d-1,s}_{i:j}(v)\alpha^{d-1,s}_{i:t}(u)\hat{\Delta}^{d,v}_{t+1:j}
			A^{d,s}_{u,v,t}\mathbf{f}^{d,s}_{\sigma^{transit},u,v}(t)
\end{eqnarray}

Using Equations~\ref{ESS-transit1} and~\ref{ESS-transit3} we
obtain the ESS for the state-transition features
\begin{eqnarray}
	&& \mathbb{E}[\mathbf{F}^{d,s}_{\sigma^{transit},u,v}(\zeta)]
	 = \sum_{t\in [1,T-1]}	\mathbb{E}[\mathbf{f}^{d,s}_{\sigma^{transit},u,v}(t)\delta[c^{transit} \in \zeta]] \nonumber\\
	\label{ESS-transit}
	&&= \frac{1}{Z}\sum_{t\in [1,T-1]}A^{d,s}_{u,v,t}\mathbf{f}^{d,s}_{\sigma^{transit},u,v}(t)
			\sum_{i\in [1,t]}\sum_{j \in [t+1,T]}
			\alpha^{d-1,s}_{i:t}(u)\lambda^{d-1,s}_{i:j}(v)\hat{\Delta}^{d,v}_{t+1:j}
\end{eqnarray}
When $d=2$ we must fix $i=1$ since $\alpha^{1,s}_{i:t}(u)$
and $\lambda^{1,s}_{i:j}(v)$ are only defined at $i=1$.
%-------------------------
%\\\\
\subsection{ESS for Initialisation Features}
%\textbf{ESS for emission features:}
%\\
%-------
Recall that in Section~\ref{sec:param} we define
$\mathbf{f}^{d,s}_{\sigma^{init},u}(i)$ as a function at level $d$ that is triggered at
time $i$ when a parent $s$ at level $d$ initialises a child $u$ at level $d+1$.
In this event, the context $c^{init} = (e^d_{i-1}=1)$ must be activated for $i > 1$.
Thus, Equation~\ref{glob-feature-init} can be rewritten as
\begin{eqnarray}
	\label{ESS-emits}
	\mathbf{F}^{d,s}_{\sigma^{init},u}(\zeta) = \sum_{i\in[1,T]}\mathbf{f}^{d,s}_{\sigma^{init},u}(i)\delta[c^{init} \in \zeta]
\end{eqnarray}

Now we consider the following feature expectation
\begin{eqnarray}
	\label{ESS-emit1}
	\mathbb{E}[\mathbf{f}^{d,s}_{\sigma^{init},u}(i)\delta[c^{init} \in \zeta]] &=& \sum_{\zeta}\Pr(\zeta)\mathbf{f}^{d,s}_{\sigma^{init},u}(i)\delta[c^{init} \in \zeta]\nonumber\\
		&=& \frac{1}{Z}\sum_{\zeta}\Phi[\zeta]\mathbf{f}^{d,s}_{\sigma^{init},u}(i)
				\delta[c^{init} \in \zeta]
\end{eqnarray}
For each assignment of $\zeta$ that enables $\mathbf{f}^{d,s}_{\sigma^{init},u}(i)$, we have
the following decomposition
\begin{eqnarray}
	\zeta = (\hat{{\underline{\zeta}}}^{d,s}_{i:j}(u),\hat{{\zeta}}^{d+1,u}_{i:j})
\end{eqnarray}
where the context $c^{init}$ activates the emission from $s$ to $u$ and
the feature function $\mathbf{f}^{d,s}_{\sigma^{init},u}(i)$.
Thus the joint potential $\Phi[\zeta]$ can be factorised as
\begin{eqnarray}
	\Phi[\zeta] = \Phi[\hat{{\underline{\zeta}}}^{d,s}_{i:j}(u)]
		\Phi[\hat{{\zeta}}^{d+1,u}_{i:j}]R^{d+1,u}_{i:j}\pi^{d,s}_{u,i}
\end{eqnarray}
Using this factorisation and noting that the elements within
the summation in the RHS of Equation~\ref{ESS-emit1}
are only non-zeros with such assignments, we can simplify 
the RHS of Equation~\ref{ESS-emit1} to 
\begin{eqnarray}
	&&\frac{1}{Z}\sum_{j \in [i,T]}\sum_{{\underline{\zeta}}^{d,s}_{i:j}(u)}
			\sum_{{\zeta}^{d+1,u}_{i:j}}
			\Phi[\hat{{\underline{\zeta}}}^{d,s}_{i:j}(u)]
			\Phi[\hat{{\zeta}}^{d+1,u}_{i:j}]
			R^{d+1,u}_{i:j}\pi^{d,s}_{u,i} \mathbf{f}^{d,s}_{\sigma^{init},u}(i) \nonumber\\
	\label{ESS-emit2}
		& & =\frac{1}{Z}\sum_{j\in[i,T]}\lambda^{d,s}_{i:j}(u)\hat{\Delta}^{d+1,u}_{i:j}\pi^{d,s}_{u,i}\mathbf{f}^{d,s}_{\sigma^{init},u}(i)
\end{eqnarray}
The summation over $j \in [i,T]$ is due to the fact that
we do not know this index. 

Using Equation~\ref{ESS-emits} and \ref{ESS-emit2} we obtain the ESS for the initialisation features
\begin{eqnarray}
	\mathbb{E}[\mathbf{F}^{d,s}_{\sigma^{init},u}(\zeta)] &=& \sum_{i\in [1,T]}\mathbb{E}[\mathbf{f}^{d,s}_{\sigma^{init},u}(i)\delta[c^{init} \in \zeta]] \nonumber\\
	\label{ESS-emit}	
		&=& \frac{1}{Z}\sum_{i\in [1,T]}\pi^{d,s}_{u,i} \mathbf{f}^{d,s}_{\sigma^{init},u}(i)
					\sum_{j\in[i,T]}\lambda^{d,s}_{i:j}(u)\hat{\Delta}^{d+1,u}_{i:j}
\end{eqnarray}

There are two special cases: (1) when $d=1$, there must be no scanning of $i$ but fix
$i=1$ since there is only a single initialisation at the beginning of sequence,
(2) when $d=D-1$, we fix $j=i$ for $\hat{\Delta}^{D,u}_{i:j}$ is only defined
at $i=j$.
%-------------------------
%\\\\
\subsection{ESS for Ending Features}
%\textbf{ESS for exiting features:} 
%\\
%-----
Recall that in Section~\ref{sec:param} we define
$\mathbf{f}^{d,s}_{\sigma^{end},u}(j)$ as a function that is activated
when a child $u$ at level $d+1$ returns the control
to its parent $s$ at level $d$ and time $j$. 
This event also enables the context $c^{end} = (e^{d}_{j}=1)$.
Thus Equation~\ref{glob-feature-exit} can be rewritten as
\begin{eqnarray}
	\label{ESS-exit1}
	\mathbf{F}^{d,s}_{\sigma^{end},u}(\zeta) = \sum_{j\in [1,T]}\mathbf{f}^{d,s}_{\sigma^{end},u}(j)\delta[c^{end} \in \zeta]
\end{eqnarray}

Now we consider the following feature expectation
\begin{eqnarray}
	\mathbb{E}[\mathbf{f}^{d,s}_{\sigma^{end},u}(j)\delta[c^{end} \in \zeta]] 
		&=& \sum_{\zeta}\Pr(\zeta)\mathbf{f}^{d,s}_{\sigma^{end},u}(j)\delta[c^{end} \in \zeta]\nonumber\\
	\label{ESS-exit2}
		&=& \frac{1}{Z}\sum_{\zeta}\Phi[\zeta]\mathbf{f}^{d,s}_{\sigma^{end},u}(j)\delta[c^{end} \in \zeta]
\end{eqnarray}
Assume that the state $s$ starts at $i$ and ends at $j$.
For each assignment of $\zeta$ that enables $\mathbf{f}^{d,s}_{\sigma^{end},u}(j)$ and respects
this assumption, we have the following decomposition
\begin{eqnarray}
	\zeta = (\hat{{\underline{\zeta}}}^{d,s}_{i:j},
					\hat{{\zeta}}^{d,s}_{i:j}(u))
\end{eqnarray}
This assignment has the context $c^{end}$ that activates the ending of $u$.
Thus the joint potential $\Phi[\zeta]$ can be factorised as
\begin{eqnarray}
	\Phi[\zeta] = \Phi[\hat{{\underline{\zeta}}}^{d,s}_{i:j}]
		\Phi[\hat{{\zeta}}^{d,s}_{i:j}(u)]
		R^{d,s}_{i:j}E^{d,s}_{u,j}
\end{eqnarray}

Substituting this factorisation into the summation of the RHS of Equation~\ref{ESS-exit2} yields
{\small
\begin{eqnarray}
	\label{ESS-exit3}			
	\sum_{i\in[1,j]}\sum_{{{\underline{\zeta}}}^{d,s}_{i:j}}
			\sum_{{{\zeta}}^{d,s}_{i:j}(u)}
			\Phi[\hat{{\underline{\zeta}}}^{d,s}_{i:j}]
			\Phi[\hat{{\zeta}}^{d,s}_{i:j}(u)]
			R^{d,s}_{i:j}E^{d,s}_{u,j}\mathbf{f}^{d,s}_{\sigma^{end},u}(j)
		= \sum_{i\in[1,j]}\hat{\Lambda}^{d,s}_{i:j}\alpha^{d,s}_{i:j}(u)
			E^{d,s}_{u,j}\mathbf{f}^{d,s}_{\sigma^{end},u}(j)
\end{eqnarray}}

Using Equations~\ref{ESS-exit1} and \ref{ESS-exit3} we obtain
the ESS for the exiting features
\begin{eqnarray}
	\label{ESS-exit}
	\mathbb{E}[\mathbf{F}^{d,s}_{\sigma^{end},u}(\zeta)] &=& \sum_{j\in [1,T]}\mathbb{E}[\mathbf{f}^{d,s}_{\sigma^{end},u}(j)\delta[e^d_{i-1} \in \zeta]] \nonumber\\
		&=& \frac{1}{Z}\sum_{j\in [1,T]}E^{d,s}_{u,j}\mathbf{f}^{d,s}_{\sigma^{end},u}(j)
			\sum_{i\in[1,j]}\hat{\Lambda}^{d,s}_{i:j}\alpha^{d,s}_{i:j}(u)
\end{eqnarray}

There is a special case: when $d=1$ there must be no scanning of $i,j$ but fix
$i=1,j=T$.

%----------------------------
\section{Partially Observed Data in Learning and Inference}
\label{sec:partial}
So far we have assumed that training data is fully labeled,
and that testing data does not have any labels.
In this section we extend the AIO to handle the cases in which
these assumptions do not hold. Specifically,
it may happen that the training data is 
not completely labeled, possibly due to lack of labeling resources. In this case, 
the learning algorithm should
be robust enough to handle missing labels. On the other hand,
during inference, we may partially obtain high quality labels from
external sources. This requires the inference algorithm to be responsive
to that data. 
%In this section, we provide a general treatment and leave
%the two important special cases in Appendix~\ref{sec:apdx-partial}.

%---
\subsection{The Constrained AIO algorithm}
\label{sec:CAIO}
%--
In this section we consider the general case
when $\zeta = (\vartheta,h)$, where
$\vartheta$ is the visible set labels,
and $h$ the hidden set. Since
our {\HCRF} is also an exponential model it shares
the same computation required for general CRFs
(Equations~\ref{log-ll-hidden} and~\ref{log-ll-hidden-grad}).
We have to compute four quantities: the partial
log-partition function $Z(\vartheta,z)$, the partition
function $Z(z)$, the
`constrained' ESS $\mathbb{E}_{h|\vartheta,z}[\F(\vartheta,h,z)]$,
and the `free' ESS $\mathbb{E}_{\zeta|z}[\F(\zeta,z)]$.
The partition function and the `free' ESS has been computed  in
Sections~\ref{sec:AIO-all} and~\ref{sec:learning}, respectively. 
This section describes the other two quantities.

Let the set of visible labels be $\vartheta = (\widetilde{x},\widetilde{e})$ 
where $\widetilde{x}$ is the visible set of state variables
and $\widetilde{e}$ is the visible set of ending indicators.
The basic idea is that we have to modify procedures for 
computing the building blocks such as
$\Delta^{d,s}_{i:j}$ and $\alpha^{d,s}_{i:j}(u)$, to 
address constraints imposed by the labels.
For example, $\Delta^{d,s}_{i:j}$ implies that the state $s$ at level $d$
starts at $i$ and persists till terminating at $j$.
Then, if any labels (e.g. there is an $\widetilde{x}^d_k \ne s$ for $k \in [i,j]$)
are seen, causing this assumption to be inconsistent, $\Delta^{d,s}_{i:j}$ will be zero.
Therefore, in general, the computation of each building block is multiplied
by an identity function that enforces the consistency between these
labels and the required constraints for computation of that block.
As an example, we consider the computation of $\Delta^{d,s}_{i:j}$
and $\alpha^{d,s}_{i:j}(u)$.

The symmetric inside mass $\Delta^{d,s}_{i:j}$ is consistent only if
all of the following conditions are satisfied:
\begin{enumerate}
\item If there are state labels $\widetilde{x}^d_k$ at level $d$ within the interval $[i,j]$,
		then $\widetilde{x}^d_k = s$,
\item If there is any label of ending indicator
	$\widetilde{e}^{d}_{i-1}$, then $\widetilde{e}^{d}_{i-1} = 1$,
\item If there is any label of ending indicator $\widetilde{e}^d_{k}$
		for some  $ k \in [i,j-1]$,
		then $\widetilde{e}^d_{k} = 0$, and
\item If any ending indicator $\widetilde{e}^{d}_j$
		is labeled, then $\widetilde{e}^{d}_j = 1$.
\end{enumerate}

These conditions are captured by
using the following identity
function:
\begin{eqnarray}
	\label{partial-inside}
	\mathbb{I}[\Delta^{d,s}_{i:j}] = \delta[\widetilde{x}^d_{k \in [i,j]} = s]\delta[\widetilde{e}^{d}_{i-1} =1]
		 \delta[\widetilde{e}^d_{k \in [i:j-1]} = 0]\delta[\widetilde{e}^{d}_{j} = 1]
\end{eqnarray}
When labels are observed, Equation~\ref{I-IF} is thus replaced by
\begin{eqnarray}
	\label{partial-I-IF}
	\Delta^{d,s}_{i:j} = \mathbb{I}[\Delta^{d,s}_{i:j}]
			\bigg(\sum_{u \in S^{d+1}}\alpha^{d,s}_{i:j}(u)E^{d,s}_{u,j}\bigg)
\end{eqnarray}
Note that we do not need to explicitly enforce the
state consistency in the summation over $u$ since in the bottom-up and left-right
computation, $\alpha^{d,s}_{i:j}(u)$ is already computed and
contributes to the sum only if it is consistent.

Analogously, the asymmetric inside mass  $\alpha^{d,s}_{i:j}(u)$
is consistent if all of the following conditions
are satisfied:
\begin{enumerate}
\item The first three conditions for the symmetric inside  mass $\Delta^{d,s}_{i:j}$ hold,
\item If the state at level $d$ at time $j$ is labeled, it must be $u$, and
\item If any ending indicator $\widetilde{e}^{d+1}_j$ is labeled, then $\widetilde{e}^{d+1}_j = 1$.
\end{enumerate}

These conditions are captured by the identity function
\begin{eqnarray}
	\label{partial-asym-inside}
	\mathbb{I}[\alpha^{d,s}_{i:j}(u)] = 
					\delta[\widetilde{x}^d_{k \in [i,j]} = s]
					\delta[\widetilde{e}^{d}_{i-1} =1]
					\delta[\widetilde{e}^d_{k \in [i:j-1]} = 0]
					\delta[\widetilde{x}^{d+1}_{j} = u]
					\delta[\widetilde{e}^{d+1}_{j} = 1]
\end{eqnarray}
Thus Equation~\ref{IF-detail} becomes
\begin{eqnarray}
	\label{partial-IF-detail}
	\alpha^{d,s}_{i:j}(u) = \mathbb{I}[\alpha^{d,s}_{i:j}(u)]\Bigg(\sum_{k=i+1}^{j}\sum_{v \in S^{d+1}}\alpha^{d,s}_{i:k-1}(v)
		\hat{\Delta}^{d+1,u}_{k:j}A^{d,s}_{v,u,k-1}
		 + \hat{\Delta}^{d+1,u}_{i:j}\pi^{d+1,s}_{u,i} \Bigg)
\end{eqnarray}
Note that we do not need to explicitly enforce the
state consistency in the summation over $v$
and time consistency in the summation over $k$
since in bottom-up
computation, $\alpha^{d,s}_{i:j}(u)$
and $\Delta^{d+1,u}_{k:j}$ are already computed and
contribute to the sum only if they are consistent.
Finally, the constrained partition function $Z(\vartheta,z)$ is
computed using Equation~\ref{Z-I} given that the inside mass is consistent with
the observations.

Other building blocks, such as the symmetric outside mass $\Lambda^{d,s}_{i:j}$
and the asymmetric outside mass $\lambda^{d,s}_{i:j}(u)$, are computed
in an analogous way. Since $\Lambda^{d,s}_{i:j}$
and $\Delta^{d,s}_{i:j}$ are complementary and they share
$(d,s,i,j)$, the same indicator function  $\mathbb{I}[\Delta^{d,s}_{i:j}]$
can be applied. Similarly,
the pair asymmetric inside mass $\alpha^{d,s}_{i:j}(u)$
and asymmetric outside mass $\lambda^{d,s}_{i:j}(u)$
are complementary and they share $d,s,i,j,u$, thus
the same indicator function $\mathbb{I}[\alpha^{d,s}_{i:j}(u)]$
can be applied. 

Once all constrained building blocks have been
computed they can be used to calculate constrained ESS as in Section~\ref{sec:learning}
without any further modifications.
The only difference is that we need to replace
the partition function $Z(z)$ by the constrained version $Z(\vartheta,z)$.

%----
\subsection{The Constrained Viterbi Algorithm}
\label{sec:CViterbi}
%--
Recall that in the Generalised Viterbi Algorithm described in
Section~\ref{sec:MAP} we want to find the most
probable configuration $\zeta^{MAP} = \arg\max_{\zeta} \Pr(\zeta|z)$.
When some variables $\vartheta$ of $\zeta$ are labeled, it is
not necessary to estimate them. The task is now 
to estimate the most probable configuration of
the hidden variables $h$ given the labels:
\begin{eqnarray}
	h^{MAP} &=& \arg\max_{h} \Pr(h|\vartheta,z) \nonumber \\
			&=& \arg\max_{h} \Pr(h,\vartheta|z) \nonumber \\
			&=& \arg\max_{h} \Phi[h,\vartheta,z]
\end{eqnarray}
It turns out that the constrained MAP estimation is identical to the standard
MAP except that we have to respect the labeled variables $\vartheta$.

Since the Viterbi algorithm is just the max-product version of
the AIO, the constrained Viterbi can be modified in the same manner
as in the constrained AIO (Section~\ref{sec:CAIO}). Specifically,
for each auxiliary quantities such as $\Delta^{max,s}_{i:j}$ and
$\alpha^{max,s}_{i:j}(u)$, we need
to maintain a set of indicator functions that ensures the
consistency with labels. 
Equations~\ref{partial-inside} and~\ref{partial-I-IF}
become
\begin{eqnarray}
	\mathbb{I}[\Delta^{max,d,s}_{i:j}] &=& 
			\delta[\widetilde{x}^d_{k \in [i,j]} = s]
			\delta[\widetilde{e}^{d}_{i-1} =1]
		 	\delta[\widetilde{e}^d_{k \in [i:j-1]} = 0]
		 	\delta[\widetilde{e}^{d}_{j} = 1] \nonumber\\
	\Delta^{max,d,s}_{i:j} &=& \mathbb{I}[\Delta^{max,d,s}_{i:j}]
			\bigg(\max_{u \in S^{d+1}}\alpha^{max,d,s}_{i:j}(u)E^{d,s}_{u,j}\bigg)
\end{eqnarray}
Likewise, we have the modifications to Equation~\ref{partial-asym-inside}
and Equation~\ref{partial-IF-detail}, respectively.
\begin{eqnarray}
	\mathbb{I}[\alpha^{max,d,s}_{i:j}(u)] &=& 
				\delta[\widetilde{x}^d_{k \in [i,j]} = s]
				\delta[\widetilde{e}^{d}_{i-1} =1]
				\delta[\widetilde{e}^d_{k \in [i:j-1]} = 0]
		 		\delta[\widetilde{x}^{d+1}_{j} = u]
		 		\delta[\widetilde{e}^{d+1}_{j} = 1]\nonumber\\
	\alpha^{max,d,s}_{i:j}(u) &=& \mathbb{I}[\alpha^{max,d,s}_{i:j}(u)]
		\max\Bigg\{\max_{k\in [i+1,j]}\max_{v \in S^{d+1}}\alpha^{max,d,s}_{i:k-1}(v)
		\hat{\Delta}^{max,d+1,u}_{k:j}A^{d,s}_{v,u,k-1}; \Bigg. \nonumber\\
		 && \q\q\q\q\q\q\q\q\Bigg.\hat{\Delta}^{max,d+1,u}_{i:j}\pi^{d+1,s}_{u,i} \Bigg\}
\end{eqnarray}

Other tasks in the Viterbi algorithm including bookkeeping and backtracking
are identical to those described in Section~\ref{sec:MAP}.

%--
\subsection{Complexity Analysis}
\label{sec:complexity-partial}
The complexity of the constrained AIO
and constrained Viterbi has an upper bound of $\mathcal{O}(T^3)$,
when no labels are given. It also has a lower bound of  $\mathcal{O}(T)$ 
when all ending indicators are known and the 
model reduces to the standard tree-structured graphical model.
In general, the complexity decreases as more labels are available,
and we can expect a sub-cubic time behaviour.

%----------------------------
\section{Numerical Scaling}
\label{sec:scaling}
%--

In previous sections, we have derived AIO-based inference and 
learning algorithms for both unconstrained
and constrained models. The quantities computed by these algorithms
like the inside/outside masses often involve summation over exponentially many positive 
potentials. The potentials, when estimated from data, are
often not upper-bound, leading to the fact that
the magnitude of the masses increases exponentially fast in the sequence
length $T$, thus goes beyond the numerical
capacity of most machines for moderate $T$.

In this section we present a scaling method to reduce this 
\emph{numerical overflow} problem.
The idea can be traced back to the Pearl's message-passing procedure 
\citep{Pearl88,Yedidia-et-al-IEEETIT05}. Our AIO algorithms can
be considered as generalisation of the message-passing, in which
the inside masses play the role of the inside-out messages.
In Pearl's method, we reduce the messages' magnitude by normalising
them at each step. In the context of HHMMs 
with which the numerical \em underflow \em problem is associated,
the similar idea has been proposed in \citep{Bui-et-al04}, which
we adapt to our overflow problem.

%--------------------
\subsection{Scaling the Symmetric/Asymmetric Inside Masses}
%--------------
Before proceeding to algorithmic details let us revisit
Equation~\ref{I-IF}. If we scale down
the asymmetric inside mass $\alpha^{d,s}_{i:j}(u)$ by a factor $\kappa_j>1$, i.e.
\begin{eqnarray}
	\alpha^{'d,s}_{i:j}(u) \leftarrow \frac{\alpha^{d,s}_{i:j}(u)}{\kappa_j}
\end{eqnarray}
then the symmetric inside mass $\Delta^{d,s}_{i:j}$ is also scaled down by the same factor.
Similarly, as we can see from Equation~\ref{IF-detail} that
\begin{eqnarray}
	\alpha^{d,s}_{i:j}(u) = \sum_{t=i+1}^{j}\sum_{v \in S^{d+1}}\alpha^{d,s}_{i:t-1}(v)
							\hat{\Delta}^{d+1,u}_{t:j}A^{d,s}_{v,u,t-1} + \hat{\Delta}^{d+1,u}_{i:j}\pi^{d,s}_{u,i}	\nonumber
\end{eqnarray}
where $\hat{\Delta}^{d+1,u}_{t:j} = {\Delta}^{d+1,u}_{t:j}R^{d+1,u}_{t:j}$,
if $\Delta^{d+1,u}_{t:j}$ for $t \in [1,j]$ is reduced by $\kappa_j$, then
$\alpha^{d,s}_{i:j}$ is also reduced by the same factor. 
In addition, using the set of recursive 
relations in Equations~\ref{IF-detail} and~\ref{I-IF}, any reduction at the bottom level
of $\Delta^{D,s}_{j:j}$ will result in the reduction
of the symmetric inside mass $\Delta^{d,s}_{i:j}$ and
of the asymmetric inside mass $\alpha^{d,s}_{i:j}(u)$, for $d < D$,
by the same factor.

Suppose $\Delta^{D,s}_{i:i}$ for all $i \in [1,j]$ is reduced
by a factor of $\kappa_i > 1$, the quantities
$\Delta^{d,s}_{1:j}$ and $\alpha^{d,s}_{1:j}(u)$ will be
reduced by a factor of $\prod_{i=1}^j\kappa_i$.
That is
\begin{eqnarray}
	\hat{\Delta}^{'d,s}_{1:j} &\leftarrow& \frac{\hat{\Delta}^{d,s}_{1:j}}{\prod_{i=1}^j\kappa_i} \\
	\alpha^{'d,s}_{1:j}(u) &\leftarrow& \frac{\alpha^{d,s}_{1:j}(u)}{\prod_{i=1}^j\kappa_i}
\end{eqnarray}
It follows immediately from Equation~\ref{Z-I} that
the partition function is scaled down by a factor
of $\prod_{i=1}^T\kappa_i$
\begin{eqnarray}
	Z' = \sum_{s\in S^1}\hat{\Delta}^{'1,s}_{1:T} = \frac{Z}{\prod_{j=1}^T\kappa_j}
\end{eqnarray}
where $\hat{\Delta}^{'1,s}_{1:T} = {\Delta}^{'1,s}_{1:T}B^{1,s}_{1:T}$.
Clearly, we should deal
with the log of this quantity to avoid numerical overflow. Thus, the
log-partition function can be computed as
\begin{eqnarray}
	\label{Z-scale}
	\log(Z) = \log\sum_{s\in S^1}\hat{\Delta}^{'1,s}_{1:T} + \sum_{j=1}^T\log\kappa_j
\end{eqnarray}
where ${\Delta}^{'1,s}_{1:T}$ has been scaled appropriately.

One question is how to choose the set of
meaningful scaling factors $\{\kappa_j\}_{1}^{T}$.
The simplest way is to choose a relatively large number
for all scaling factors but making the right choice is not straightforward.
Here we describe a more natural way to do so.
Assume that we have chosen all the scaling factors
$\{\kappa_i\}_{1}^{j-1}$. 
Using the original Equations~\ref{IF-detail}, \ref{IF-special}, and \ref{IF-special2},
where all the sub-components have been scaled appropriately,
we compute the \em partially-scaled \em
inside mass $\Delta^{''d,s}_{i:j}$ for $d \in [2,D]$ and asymmetric inside mass $\alpha^{''d,s}_{i:j}(u)$, 
for $d \in [1,D-1]$ and $i\in [1,j]$.
Then the scaling factor at time $j$ is computed as
\begin{eqnarray}
	\label{scaling-factor}
	\kappa_j = \sum_{s,u}\alpha^{''1,s}_{1:j}(u)
\end{eqnarray}

The next step is to rescale all the partially-scaled variables:
\begin{eqnarray}
	\label{IF-rescale}
	\alpha^{'d,s}_{i:j}(u) &\leftarrow& \frac{\alpha^{''d,s}_{i:j}(u)}{\kappa_j}
		\mbox{ for } s \in S^d, d \in [1,D-1] \\
	\label{I-rescale}
	\Delta^{'d,s}_{i:j} &\leftarrow& \frac{\Delta^{''d,s}_{i:j}}{\kappa_j}
		\mbox{ for } s \in S^d, d \in [2,D-1] \\
	\label{I-rescale2}
	\Delta^{'D,s}_{j:j} &\leftarrow& \frac{\Delta^{''D,s}_{j:j}}{\kappa_j}
		\mbox{ for } s \in S^D
\end{eqnarray}
where $i \in [1,j]$.

%--------------------
\subsection{Scaling the Symmetric/Asymmetric Outside Masses}
%--------------
In a similar fashion we can work out
the set of factors from the derivation of symmetric/asymmetric outside masses since
these masses solely depend on the inside masses
as building blocks. In other words, after we finish scaling
the inside masses we can compute the scaled outside masses
directly, using the same set of equations described in
Section~\ref{sec:OB}. 

The algorithm is summarised in Figure~\ref{alg:scaling}. Note that
the order of performing the loops in this case is different from that
in Figure~\ref{alg:inside}.

%----------the algorithm
\begin{figure}[htb]
\begin{center}
%\begin{algorithm}
\begin{tabular}{l}\hline
\textbf{Input}:  $D,T$ and all the contextual potentials.\\
\textbf{Output}: Scaled quantities: inside/asymmetric inside masses,\\
\q\q 				outside/asymmetric outside masses.\\
\hline
\textbf{For} $j=1,2,..,T$ \\
\q	Compute $\alpha^{d,s}_{1:j}(u)$, $d \in [1,D-1]$ using Equations~\ref{IF-detail}, \ref{IF-special} and \ref{IF-special2} \\
\q	Compute $\kappa_j$ using Equation~\ref{scaling-factor} \\
\q	Rescale $\alpha^{1,s}_{1:j}(u)$  using Equation~\ref{IF-rescale}\\
\q	\textbf{For} $i=1,2,..,j$ \\
\q\q		\textbf{For} $d=2,3,..,D-1$ \\
\q\q\q			 Rescale $\alpha^{d,s}_{i:j}(u)$ using Equation~\ref{IF-rescale}\\
\q\q\q			 Rescale $\Delta^{d,s}_{i:j}$ using Equation~\ref{I-rescale}\\
\q\q		\textbf{EndFor}\\
\q	\textbf{EndFor}\\
\q	Rescale $\Delta^{D,s}_{j:j}$ using Equation~\ref{I-rescale2}\\
\textbf{EndFor}\\
Compute true log-partition function using Equation~\ref{Z-scale}.\\
Compute the outside/asymmetric outside masses using the \\
\q	scaled inside/asymmetric inside masses instead of the original \\
\q	inside/asymmetric inside in Equations~\ref{OB-OB} and~\ref{O-OB}.\\
\hline
\end{tabular}
\end{center}
\caption{Scaling algorithm to avoid numerical overflow.} 
\label{alg:scaling}
\end{figure}
%------------------
%--

%----------------------------
\section{Applications}
\label{sec:exp}
%Experiments

%-----
\subsection{Recognising Indoor Activities}
In this experiment, we evaluate the {\HCRF}s with a relatively small
dataset from the domain of indoor video surveillance. The task is to recognise
indoor trajectories and activities of a person from his noisy positions extracted from video.
The data, which was captured in \citep{Nam-et-alCVPR05}, and subsequently
used to evaluate DCRFs in \citep{Truyen:2006},
has 90 sequences, each of which corresponds
to one of 3 the persistent activities: (1) \em preparing short-meal\em,
(2) \em having snack \em and (3) \em preparing normal-meal\em.
The persistent activities share the some of 12 sub-trajectories. 
Each sub-trajectory is a sub-sequence
of discrete positions. Thus naturally, the data has a state hierarchy of depth 3:
the dummy root for each position sequence, the persistent activities,
and the sub-trajectories. The input observations to the model are simply
sequences of discrete positions.

We split the data into two sets of equal size for training
and testing, respectively. For learning, labels for each sequence are provided fully
for the case of fully observed state data, and partially
for the case of missing state data. For testing, no labels are given to the decoder,
and decoded labels obtained from the max-product algorithm are compared against the ground-truth.

In designing features, we assume that
state features (i.e. between nodes) such as initialisation, transition and exiting
are indicator functions. For
the data-associations (i.e. embedded in state-persistence potentials) at the bottom level,
we use the same features as in \citep{Truyen:2006}.
At the second level, we use average velocities and a vector
of positions visited in the state duration.
To encode the duration
into the state-persistence potentials, we employ the sufficient statistics of
the \em gamma \em distribution as features
$f_{k}(s,\Delta t) = \mathbb{I}(s)\log(\Delta t)$
and $f_{k+1}(s,\Delta t)  = \mathbb{I}(s)(\Delta t)$.

At each level $d$ and time $t$
we count an error if the predicted state is not the
same as the ground-truth. Firstly, we examine the fully observed case where the {\HCRF}
is compared against the DCRF at both data levels,
and against the flat-CRF at bottom level.
Table~\ref{tab:perf-full} (the left half) shows that (a) both the multilevel
models significantly outperform the flat model
and (b) the {\HCRF} outperforms the DCRF.

%-------
\begin{table}[htb]
\begin{center}
\begin{tabular}{|l|c|c||l|c|c|} \hline
Alg.  		& $d=2$ & $d=3$ & Alg.  	& $d=2$ & $d=3$ \\ \hline\hline
{\HCRF}		& 100	& 93.9  & PO-{\HCRF}	& 80.2	& 90.4 \\ \hline
DCRF		& 96.5	& 89.7  & PO-CRF	&  -	& 83.5 \\ \hline
flat-CRF	& -	& 82.6  & 	-	&  -	& - 	\\ \hline
\end{tabular}
\end{center}
\caption{Accuracy (\%) for fully observed data (left), and partially observed (PO) data (right).}
\label{tab:perf-full}
\end{table}
%-------

\begin{figure}[htb]
\begin{center}
 \begin{tabular}{c}
\includegraphics[width=0.60\linewidth]{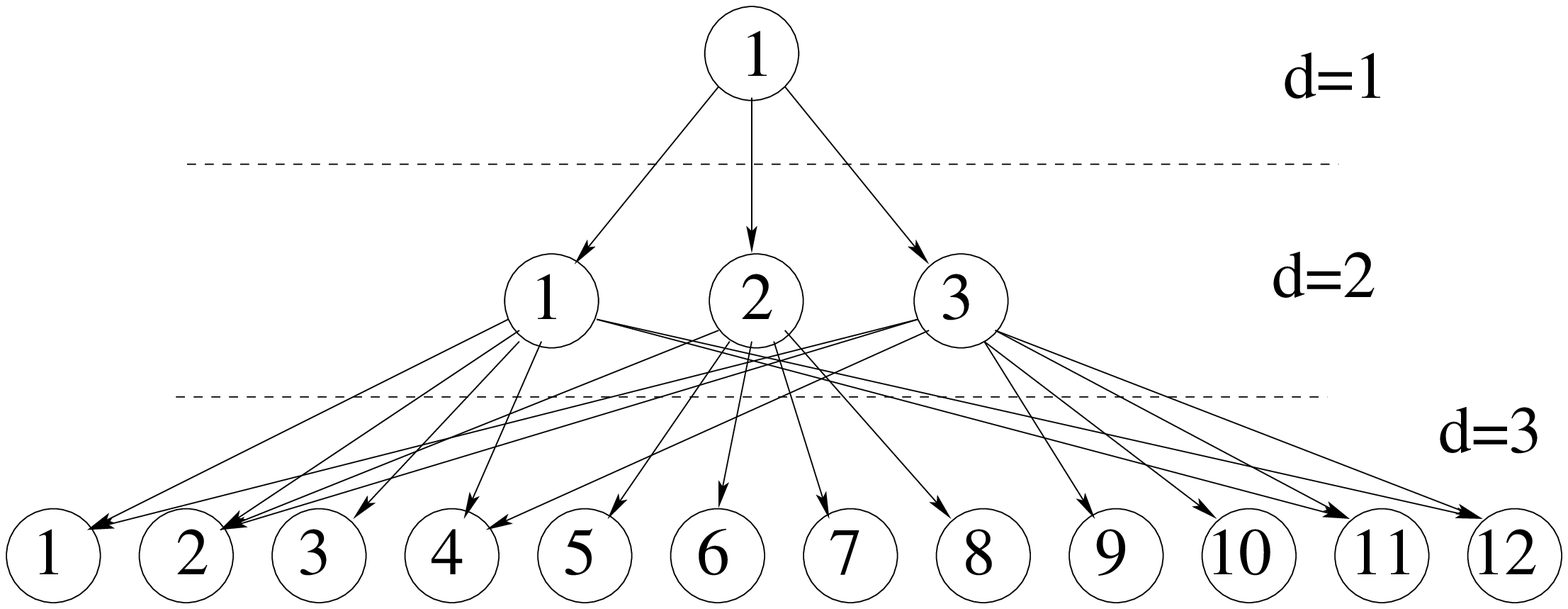}
\end{tabular}
\end{center}
\caption{The topo learned from data}
\label{fig:learnt-topo}
\end{figure}

\begin{figure}[htb]
\begin{center}
 \begin{tabular}{c}
\includegraphics[width=0.60\linewidth]{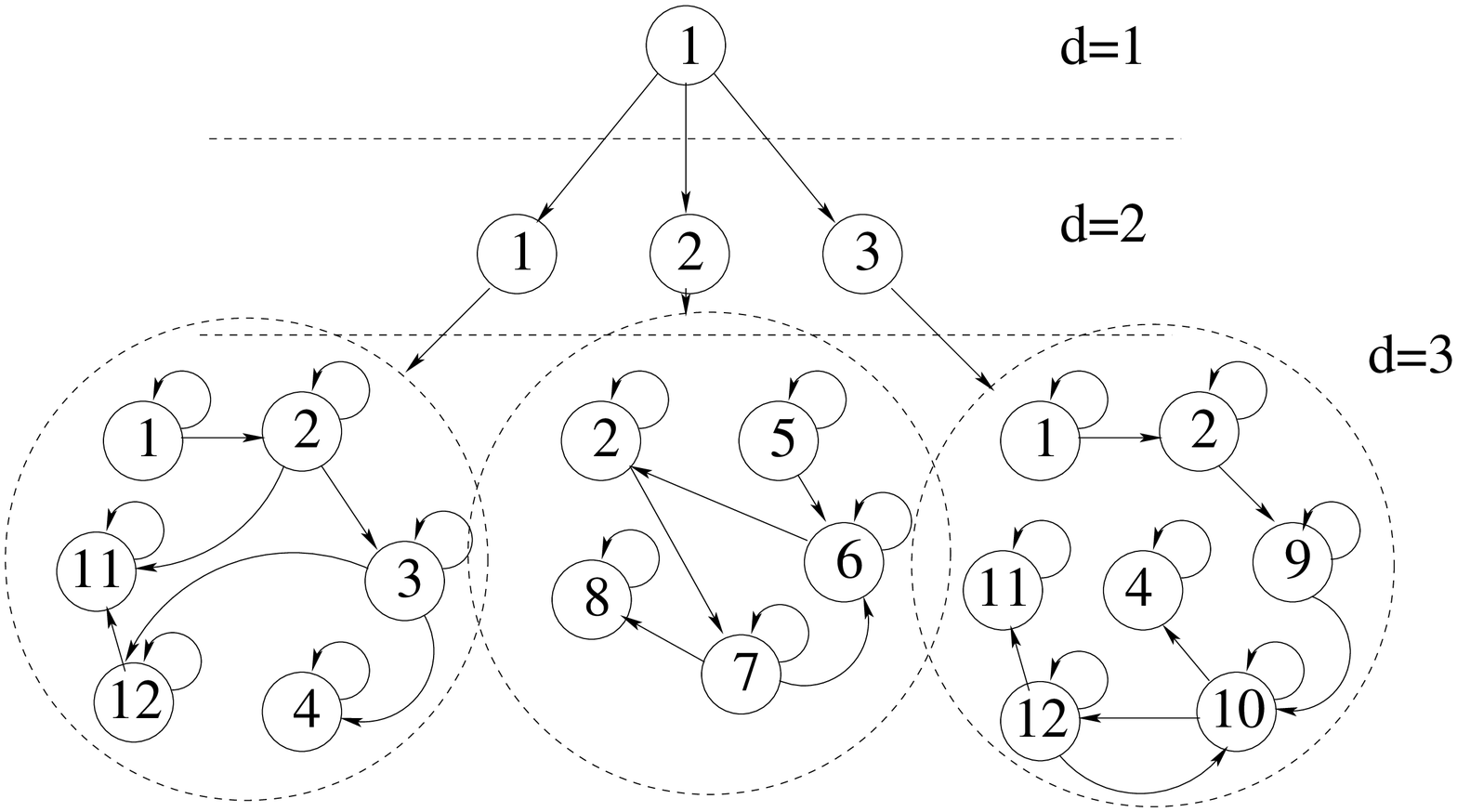}
\end{tabular}
\end{center}
\caption{The state transition model learned from data. Primitive states
	are duplicated for clarity only. They are shared
	among complex states}
\label{fig:learnt-transit}
\end{figure}

We also test the ability of the model to learn
the hierarchical topology and state transitions.
We find the it is very informative to examine
parameters which correspond to the state transition
features. Typically, negative entries in the transition
parameter matrix means that the transition is improbable.
This is because state features are non-negative, so negative
parameters mean the probabilities of these transitions
are very small (due to the exponential), compared to the positive ones.
For the transition at the second level (the complex activity
level), we obtain all negative entries. This clearly match the
training data, in which each sequence already belongs to one
of three complex activities. With this method, we
are able to construct the correct hierarchical topology
as in Figure~\ref{fig:learnt-topo}. The state transition model
is presented in Figure~\ref{fig:learnt-transit}.
There is only one wrong transition, from state 12 to state 10, which
is not presented in the training data. The rest is correct.

%-------

%-------

Next we consider partially-supervised learning
in that about 50\% of start/end times of a segment and segment labels
are observed at the second level.
All ending indicators are known at the bottom level.
The results are reported in Table~\ref{tab:perf-full} (the right half).
As can be seen, although only 50\% of the state labels and
state start/end times are observed, the model learned is still performing well
with accuracy of 80.2\% and 90.4\% at levels 2 and 3, respectively.

\begin{figure}[htb]
\begin{center}
 \begin{tabular}{c}
\includegraphics[width=0.70\linewidth]{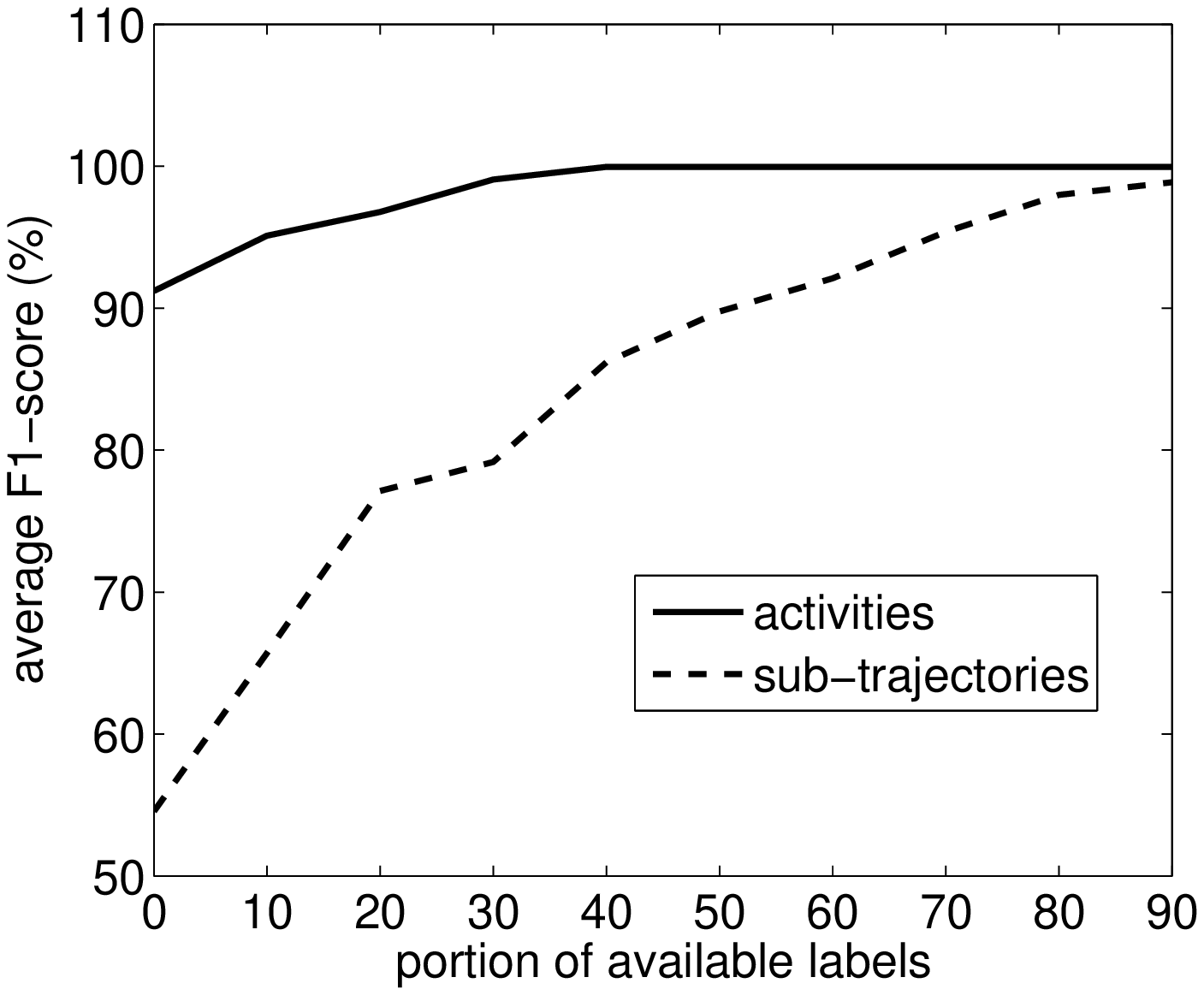}
\end{tabular}
\end{center}
\caption{Performance of the constrained max-product algorithm
		as a function of available information on label/start/end time.}
\label{fig:max-semi3-perf}
\end{figure}

%-----
We now consider the issue of using partial observed
labels during decoding to improve prediction accuracy of poorly estimated models. 
We extract the parameters from the 10th iteration of the fully observed data case.
The labels are provided at random time indexes.
Figure~\ref{fig:max-semi3-perf}a shows the decoding
accuracy as a function of available state labels. It is interesting
to observe that a moderate amount of observed labels (e.g. $20-40$\%)
causes the accuracy rate to go up considerably.

%-------

%---------------------------
\subsection{POS Tagging and Noun-Phrase Chunking}
In this experiment we apply the {\HCRF} to the task of noun-phrase chunking.
The data is from the CoNLL-2000 shared task \citep{Sang-Buchholz-CoNLL00},
in which 8926 English sentences from the Wall Street Journal corpus
are used for training and 2012 sentences are for testing. Each word in a pre-processed
sentence is labeled by two labels: the part-of-speech (POS) and the 
noun-phrase (NP). There are 48 POS different labels and 3 NP labels
(B-NP for beginning of a noun-phrase, I-NP for inside a noun-phrase
or O for others). Each noun-phrase generally has more than one word.
To reduce the computational burden, we reduce
the POS tag-set to 5 groups: \em noun, verb, adjective, adverb and others\em.
Since in our {\HCRF}s we do not have to explicitly indicate
which node is at the beginning of a segment, the NP label set can be reduced
further into NP for noun-phrase, and O for anything else.

The POS tags are actually the output of the Brill's tagger
\citep{Brill-CL95}, while the NPs are manually labeled. 
We extract raw features from the text in the way similar to
that in \citep{Sutton-et-alJMLR07}. However, we consider
only a limited vocabulary extracted from the training data
in that we only select words with more than 3 occurrences.
This reduces the vocabulary and the feature size significantly.
We also make use of bi-grams with similar selection criteria.
Furthermore, we use the contextual window of 5 instead of 7
as in \citep{Sutton-et-alJMLR07}. This setting gives rise to
about 32K raw features. The model feature is factorised as
$f(x_c,z) = \mathbb{I}(x_c)g_c(z)$, where $\mathbb{I}(x_c)$
is a binary function on the assignment of
the clique variables $x_c$, and $g_c(z)$ are the raw features.

We build an {\HCRF} topology of 3 levels where the root
is just a dummy node, the second level has 2 NP states
and the bottom level has 5 POS states. For comparison,
we implement a DCRF, a simple sequential CRF (SCRF), 
and a semi-Markov CRF (SemiCRF) \citep{sarawagi04}. 
The DCRF has grid structure of depth 2, one for modelling the
NP process and another for the POS process. Since the state spaces
are relatively small, we are able to run exact inference
in the DCRF by collapsing both the NP and POS state spaces
to a combined state space of size $3 \times 5 = 15$.
The SCRF and SemiCRF model only the NP process,
taking the POS tags as input. 

The raw feature set used in the DCRF is identical to those 
in our {\HCRF}. However, the set shared by the SCRF and the SemiCRF
is a little more elaborate since it takes the POS tags into account
\citep{Sutton-et-alJMLR07}. 

Although both the {\HCRF} and the SemiCRF are capable of modelling
arbitrary segment durations, we use a simple exponential
distribution as it can be processed sequentially and thus
is very efficient. For learning, we
use a simple online stochastic gradient ascent method
since it has been shown to work relatively well and fast
in CRFs \citep{Vishwanathan-ICML06}. At test time, as
the SCRF and the SemiCRF are able to use the Brill's POS tags
as input, it is not fair for the DCRF and {\HCRF}
to predict those labels during inference. Instead, we
also give the POS tags to the DCRF and {\HCRF} and
perform constrained inference to predict \em only \em
the NP labels. This boosts the performance of the
two multi-level models significantly.

%Let's look at the difference between the flat setting of SCRF and
%SemiCRF and the multi-level setting
%of DCRF and {\HCRF}. Let $x = (x_{np},x_{pos})$. 
%Essentially, we are about to model the distribution
%$\Pr(x|z) = \Pr(x_{np}|x_{pos},z)\Pr(x_{pos}|z)$
%in the multi-level models while we ignore
%the $\Pr(x_{pos}|z)$ in the flat models. During
%test time of the multi-level models, we predict only the $x_{np}$ based on
%by finding the maximiser of $\Pr(x_{np}|x_{pos},z)$.
%The $\Pr(x_{pos}|z)$ seems to be a waste because we do not
%make use of it at test time. However,
%$\Pr(x_{pos}|z)$ does give extra information in the joint
%$\Pr(x|z)$, that is, modelling the POS process may help
%to get smoother estimate of the NP distribution.
%It may help to improve the prediction of the NP if the training
%data is small.

\begin{figure}[htb]
\begin{center}
 \begin{tabular}{c}
\includegraphics[width=0.70\linewidth]{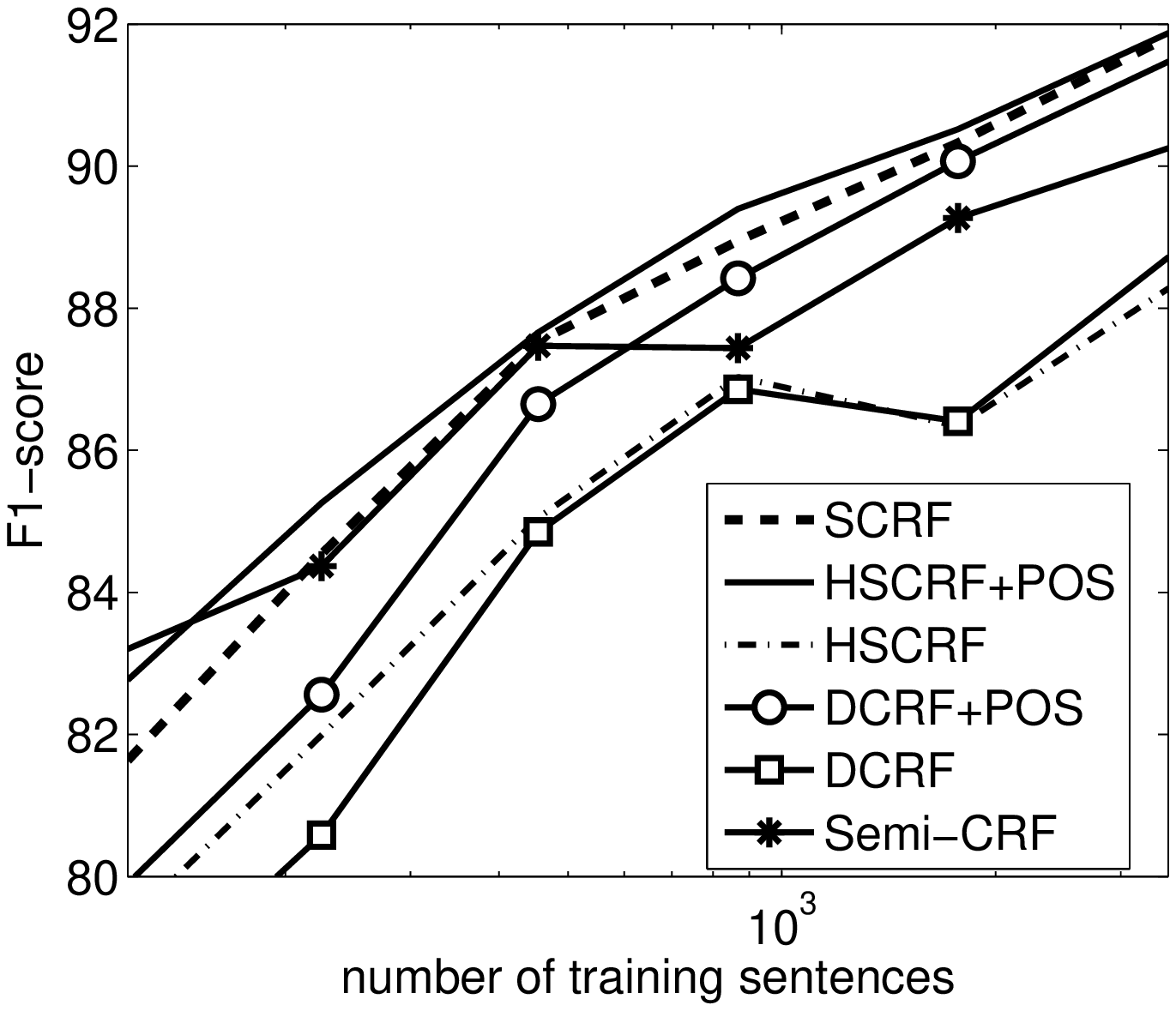}
\end{tabular}
\end{center}
\caption{Performance of various models on Conll2000 noun-phrase chunking.
			{\HCRF}+POS and DCRF+POS mean {\HCRF} and DCRF with POS given at test time,
			respectively.}
\label{fig:conll2000-perf}
\end{figure}

The performance of these models is depicted in Figure~\ref{fig:conll2000-perf}
and we are interested in only the prediction of the noun-phrases since
this data has Brill's POS tags.
Without Brill's POS tags given at test time, both the {\HCRF} and the DCRF
perform worse than the SCRF. This is not surprising because the Brill's POS
tags are always given in the case of SCRF. However, with POS tags the {\HCRF}
consistently works better than all other models. 
The DCRF does worse than the SCRF, even with POS tags given. This
does not share the observation made in \citep{Sutton-et-alJMLR07}.
However, we use a much smaller POS tag set than \citep{Sutton-et-alJMLR07} does.
Our explanation is that the SCRF is able to make use of wider context
of the given POS tags (here, within the window of 5 tags) than the DCRF
(limited to 1 POS tag per NP chunk). The SemiCRF, although in theory it is
more expressive than the SCRF, does not show any advantage under
current setting. Recall that the SemiCRF is a special case of {\HCRF}
in that the POS level is not modelled, it is possible to conclude that
joint modelling of NP and POS levels is important.

%-------

\section{Conclusions}
\label{sec:con}
% Discussion
In this paper, we have presented a novel model
called Hierarchical Semi-Markov Conditional Random Field which
extends the standard CRFs to incorporate hierarchical
and multilevel semantics.  We have developed
a graphical model-like dynamic representation
of the {\HCRF}. This appears similar to the DBN representation
of the HHMMs in \citep{Murphy-Paskin01}, and somewhat resembles
a dynamic factor graph \citep{Kschischang-et-al01}. 
However, it is not exactly the standard graphical model 
because the contextual cliques in {\HCRF}s are not fixed
during inference. 

We have derived efficient algorithms
for learning and inference, especially the ability
to learn and inference with partially given labels. 
We have demonstrated the capacity of the {\HCRF}s
on home video surveillance data and the shallow parsing of
English text, in which  the hierarchical information inherent in the context helps
to increase the recognition.

In future work we plan to attack the computational 
bottleneck in large-scale settings. 
Although the AIO family has cubic time complexity, it is still expensive
in large-scale application, especially those with long sequences.
It is therefore desirable to introduce approximation
methods that can provide speed/quality trade-offs.

We also need to make a choice between pre-computing
all the potentials prior to inference and learning, and
computing them on-the-fly. The first choice requires $\BigO(D|S|^3T^2)$ space,
which is very significant with typical real-world problems, even with
today's computing power. The second choice, however, will slow
the inference and learning very significantly due to repeated
computation at every step of the AIO algorithm.

Perhaps one of the most interesting point is that how
good the {\HCRF}s can be an approximation to general multilevel processes,
which are not necessarily recursive. For example, it is interesting
to see if any data which is naturally represented as a DCRF
can be approximately represented by an {\HCRF}. This is
important because {\HCRF}s are tractable while DCRFs are
generally not. Some data is intrinsically sequential in the sense that
there is no really `exiting' point. The {\HCRF}s force
some transitions at the edge of segments to be broken, 
so the best {\HCRF}s can do is to model quite long segments.

\bibliographystyle{natbib}
\bibliography{D:/IMPCA/doc/bibs/ME,D:/IMPCA/doc/bibs/truyen}

\appendix
%-----
\section{Proofs}
\label{apdx:proof}
%%
%%

%----
\subsection{Proof of Propositions~\ref{conj:sym-CI} and~\ref{conj:asym-CI}}
\label{apdx:proof-CI}
Before proving Proposition~\ref{conj:sym-CI} and~\ref{conj:asym-CI}
let us introduce a lemma.
\begin{lemma}
	\label{CI-lemma}
	Given a distribution of the form
	\begin{eqnarray}
		\Pr(x) = \frac{1}{Z}\Phi[x]	
	\end{eqnarray}
	where $x = (x_a,x_s,x_b)$, if there exists a factorisation
	\begin{eqnarray}
		\label{CI-lemma-factorise}
		\Phi[x] = \Phi[x_a,x_s]\Phi[x_s]\Phi[x_s,x_b]
	\end{eqnarray}
	then $x_a$ and $x_b$ are conditionally independent given $x_s$.
\end{lemma}

\textbf{Proof}: We want to prove that
\begin{eqnarray}
	\label{cond-ind}
	\Pr(x_a,x_b|x_s) = \Pr(x_a|x_s)\Pr(x_b|x_s)
\end{eqnarray}

Since $\Pr(x_a,x_b|x_s) =  \Pr(x_a,x_b,x_s)/\sum_{x_a,x_b}\Pr(x_a,x_b,x_s)$, the LHS
of Equation~\ref{cond-ind} becomes
\begin{eqnarray}
	\label{CI-lemma-LHS}
	\Pr(x_a,x_b|x_s) &=& \frac{\Phi[x_a,x_s]\Phi[x_s]\Phi[x_s,x_b]}{\sum_{x_a,x_b}\Phi[x_a,x_s]\Phi[x_s]\Phi[x_s,x_b]} \nonumber\\
					&=& \frac{\Phi[x_a,x_s]}{\sum_{x_a}\Phi[x_a,x_s]}\frac{\Phi[x_s,x_b]}{\sum_{x_b}\Phi[x_s,x_b]}
\end{eqnarray}
where we have used the following fact
\begin{eqnarray}
	\sum_{x_a,x_b}\Phi[x_a,x_s]\Phi[x_s]\Phi[x_s,x_b] = 
		\Phi[x_s]\bigg(\sum_{x_a}\Phi[x_a,x_s]\bigg)\bigg(\sum_{x_b}\Phi[x_s,x_b]\bigg)
\end{eqnarray}
and canceled out the normalisation factor $Z$ and $\Phi[x_s]$.

To prove $\Pr(x_a|x_s) = \Phi[x_a,x_s]/\sum_{x_a}\Phi[x_a,x_s]$, 
we need only to show $\Pr(x_a|x_s) \propto \Phi[x_a,x_s]$ since
the normalisation over $x_a$ is due to $\sum_{x_a}\Pr(x_a|x_s) = 1$.
Using the  Bayes rule, we have
\begin{eqnarray}
	\label{CI-lemma-RHS1}
 	\Pr(x_a|x_s) &\propto& \Pr(x_a,x_s)  \nonumber\\
 				 &=& \sum_{x_b}\Pr(x_a,x_s,x_b) \nonumber\\
 				 &=& \frac{1}{Z}\Phi[x_a,x_s]\Phi[x_s]\sum_{x_b}\Phi[x_s,x_b]\nonumber\\
				 &\propto& \Phi[x_a,x_s]
\end{eqnarray}
where we have ignored all the factors that do not depend on $x_a$.

A similar proof gives  $\Pr(x_b|x_s) = \Phi[x_s,x_b]/\sum_{x_b}\Phi[x_s,x_b]$.
Combining this result and Equation~\ref{CI-lemma-RHS1}
with Equation~\ref{CI-lemma-LHS} gives us Equation~\ref{cond-ind}. This
completes the proof $\blacksquare$

In fact, $x_s$ acts as a separator between $x_a$ and $x_b$. In standard
Markov networks there are no paths from $x_a$ to $x_b$ that do not
go through $x_s$. Now we proceed to proving
Proposition~\ref{conj:sym-CI} and~\ref{conj:asym-CI}.

Given the symmetric Markov blanket
$\Pi^{d,s}_{i:j}$, there are no potentials that are associated with variables
belonging to both $\zeta^{d,s}_{i:j}$ and $\underline{\zeta}^{d,s}_{i:j}$. 
The blanket completely separates the  $\zeta^{d,s}_{i:j}$ and $\underline{\zeta}^{d,s}_{i:j}$.
Therefore, Lemma~\ref{CI-lemma} ensures the conditional independence
between $\zeta^{d,s}_{i:j}$ and $\underline{\zeta}^{d,s}_{i:j}$.

Similarly, the asymmetric Markov blanket $\Gamma^{d,s}_{i:j}(u)$ separates
${\zeta}^{d,s}_{i:j}(u)$ and $\underline{\zeta}^{d,s}_{i:j}(u)$
and thus these two variable sets are conditionally independent due to
Lemma~\ref{CI-lemma} $\blacksquare$

%----
\subsection{Proof of Proposition~\ref{conj:SIO-conds}}
\label{apdx:proof-SIO-conds}
Here we want to derive Equations~\ref{SIO-cond}, \ref{SIO-cond2} and \ref{sym-blanket-prob}.
With the same conditions as in Lemma~\ref{CI-lemma}, in
Equation~\ref{CI-lemma-RHS1} we have shown that
$\Pr(x_a|x_s) \propto \Phi[x_a,x_s]$.
Similarly, this extends to
\begin{eqnarray}
	\Pr(\zeta^{d,s}_{i:j}|\Pi^{d,s}_{i:j}) 
			&\propto& \Phi[\zeta^{d,s}_{i:j},\Pi^{d,s}_{i:j}] \nonumber\\
			&=& \Phi[\hat{\zeta}^{d,s}_{i:j}]
\end{eqnarray}
which is equivalent to 
\begin{eqnarray}
	\Pr(\zeta^{d,s}_{i:j}|\Pi^{d,s}_{i:j})  
		&=& \frac{1}{\sum_{\zeta^{d,s}_{i:j}}\Phi[\hat{\zeta}^{d,s}_{i:j}]}\Phi[\hat{\zeta}^{d,s}_{i:j}]  \nonumber\\
		&=& \frac{1}{\Delta^{d,s}_{i:j}}\Phi[\hat{\zeta}^{d,s}_{i:j}]
\end{eqnarray}
The last equation follows from the definition of the symmetric inside mass
in Equation~\ref{inside-def}. Similar procedure will yield Equation~\ref{SIO-cond2}.

To prove Equation~\ref{sym-blanket-prob}, notice 
the Equation~\ref{sym-CI-factorise} that says
\begin{eqnarray}
	\Pr(\zeta) = \Pr(\Pi^{d,s}_{i:j})\Pr(\zeta^{d,s}_{i:j}|\Pi^{d,s}_{i:j})\Pr(\underline{\zeta}^{d,s}_{i:j}|\Pi^{d,s}_{i:j})
\end{eqnarray}
or equivalently
\begin{eqnarray}
	\Pr(\Pi^{d,s}_{i:j}) 
		&=& \Pr(\zeta) \frac{1}{\Pr(\zeta^{d,s}_{i:j}|\Pi^{d,s}_{i:j})} \frac{1}{\Pr(\underline{\zeta}^{d,s}_{i:j}|\Pi^{d,s}_{i:j})} \\
		&=& \frac{1}{Z}\Phi[\zeta]
				\frac{\Delta^{d,s}_{i:j}}{\Phi[\hat{\zeta}^{d,s}_{i:j}]}
				\frac{\Lambda^{d,s}_{i:j}}{\Phi[\hat{\underline{\zeta}}^{d,s}_{i:j}]} \\
		&=& \frac{1}{Z}\Phi[\hat{\zeta}^{d,s}_{i:j}]R^{d,s}_{i:j}\Phi[\hat{\underline{\zeta}}^{d,s}_{i:j}]
				\frac{\Delta^{d,s}_{i:j}}{\Phi[\hat{\zeta}^{d,s}_{i:j}]}
				\frac{\Lambda^{d,s}_{i:j}}{\Phi[\hat{\underline{\zeta}}^{d,s}_{i:j}]} \\ 
		&=& \frac{1}{Z}\Delta^{d,s}_{i:j}R^{d,s}_{i:j}\Lambda^{d,s}_{i:j}
\end{eqnarray}
In the proof proceeding, we have made use of the relation in Equation~\ref{factorise}.
This completes the proof $\blacksquare$

\section{Computing the State Marginals of {\HCRF}}
\label{apdx:margin}
%
%\begin{figure}[htb]
%\begin{center}
%\begin{tabular}{c}
%\includegraphics[width=0.40\linewidth]{figs/marginal.eps}
%\end{tabular}
%\end{center}
%\caption{The marginal at given index $l$ and level $d$. Here we
%	scan the left index $1\le i \le l$ and the right index
%	$l \le j \le T$.}
%\label{fig:maginal}
%\end{figure}

We are interested in computing the marginals of
state variables $\Pr(x^d_t)$. 
We have
\begin{eqnarray}
	\label{margin1}
	\Pr(x^d_t) &=& \sum_{\zeta \backslash x^d_t}\Pr(x^d_t,\zeta \backslash x^d_t) \nonumber\\
				&=& \sum_{\zeta}\Pr(\zeta)\delta(x^d_t \in \zeta) \nonumber\\
				&=& \frac{1}{Z}\sum_{\zeta}\Phi[\zeta]\delta(x^d_t \in \zeta)
\end{eqnarray}

Let $s = x^d_t$ and assume that the state $s$ starts 
at $i$ and end at $j$, and $ t \in [i,j]$.
For each configuration $\zeta$ that respects this assumption,
we have the factorisation of Equation~\ref{factorise}
that says
\begin{eqnarray}
	\Phi[\zeta] = \Phi[\hat{\zeta}^{d,s}_{i:j}]\Phi[\hat{\underline{\zeta}}^{d,s}_{i:j}]R^{d,s}_{i:j}
\end{eqnarray}
Then Equation~\ref{margin1} becomes
\begin{eqnarray}
	\label{marginal}
	\Pr(x^d_t=s) &=&  \frac{1}{Z}\sum_{\zeta}\Phi[\hat{\zeta}^{d,s}_{i:j}]\Phi[\hat{\underline{\zeta}}^{d,s}_{i:j}]R^{d,s}_{i:j}\delta(t \in [i,j])\nonumber\\
				 &=& \frac{1}{Z}\sum_{i\in [1,t]}\sum_{j\in [t,T]}
							\Delta^{d,s}_{i:j}\Lambda^{d,s}_{i:j}R^{d,s}_{i:j}
\end{eqnarray}
The summing over $i$ and $j$ is due to the fact that we do not know these indices.

There are two special cases, (1) when $d=1$ we cannot scan the left and right indices,
the marginals are simply
\begin{eqnarray}
	\Pr(x^{1}_t=s) = \frac{1}{Z}\hat{\Delta}^{1,s}_{1:T}
\end{eqnarray}
since $\Lambda^{1,s}_{1:T}=1$ for all $s \in S^1$;
and (2) when $d=D$, the start and end times must be the same ($i=j$), thus
\begin{eqnarray}
	\Pr(x^{D}_t=s) = \frac{1}{Z}\hat{\Lambda}^{D,s}_{t:t}
\end{eqnarray}
since $\Delta^{D,s}_{t:t} = 1$ for all $t \in [1,T]$ and $s \in S^D$.

Since $\sum_{s\in S^d}\Pr(x^d_t=s) = 1$, it follows from
Equation~\ref{marginal} that 
\begin{eqnarray}
	\label{Z-general2}
	Z = \sum_{s \in S^d}\sum_{i \in [1,t]}\sum_{j \in [t,T]}\Delta^{d,s}_{i:j}\Lambda^{d,s}_{i:j}R^{d,s}_{i:j}
\end{eqnarray}
This turns out to be the most general way of computing the partition
function. Some special cases have been shown earlier.
For example, when $d=1$, $i=1$ and $j=T$,
Equation~\ref{Z-general2} becomes Equation~\ref{Z-I} since $\Lambda^{1,s}_{1:T} = 1$. 
Similarly, when $d=D$, $i=j=t$,
Equation~\ref{Z-general2} recovers Equation~\ref{Z-O} since $\Delta^{D,s}_{i:i} = 1$.

\section{Semi-Markov CRFs as Special Case of {\HCRF}s}
\label{apdx:SemiCRF}
%File {\HCRF}-apdx-HSCRF.tex
%Section: Hierarchical Semi-Markov Conditional Random Fields
%\rem{
In this Appendix we first describe the semi-Markov CRF (SemiCRF)
\citep{sarawagi04} in our {\HCRF} framework
and show how to convert a SemiCRF into an {\HCRF}. Then under the
light of {\HCRF} inference we show how to modify the original SemiCRF 
to handle (a) partial supervision and constrained inference, and (b) numerical
scaling to avoid overflow. The modifications are of
interest in their own right.

%---
\subsection{SemiCRF as an {\HCRF}}
%--
%}
\begin{figure}[htb]
\begin{center}
\begin{tabular}{c}
\includegraphics[width=0.70\linewidth]{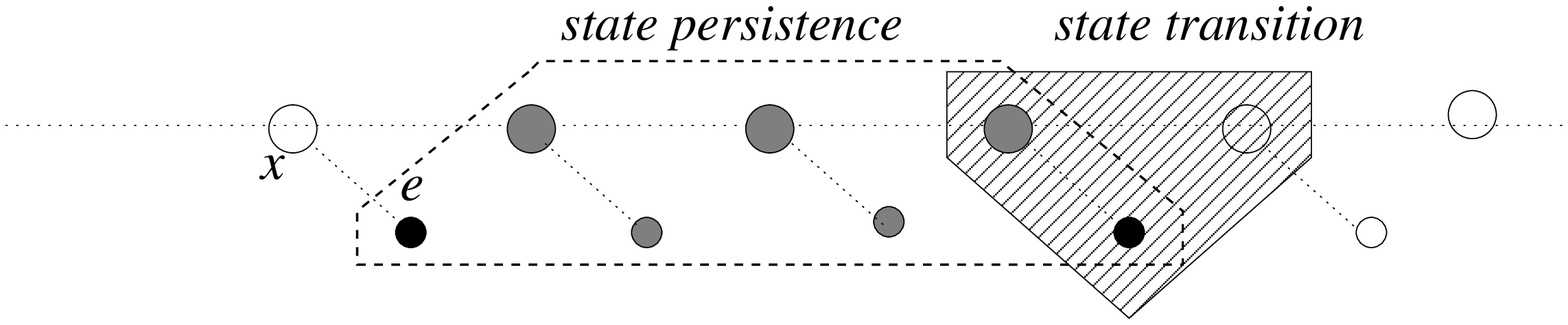}
\end{tabular}
\end{center}
\caption{The SemiCRFs in our contextual clique framework.}
\label{fig:semi-CRF}
\end{figure}

SemiCRF is an interesting flat segmental
undirected model that generalises the chain CRF. In the SemiCRF framework
the Markov process operates at the segment level, where a segment
is a non-Markovian chain of nodes. 
A chain of segments is a Markov chain. However, since each segment
can potentially have arbitrary length,
inference in SemiCRFs is more involved than the chain CRFs.

Represented in our {\HCRF} framework (Figure~\ref{fig:semi-CRF}),
each node $x_t$ of the SemiCRF is associated with an ending indicator $e_t$,
with the following contextual cliques
\begin{itemize}
\item \em Segmental state\em, which corresponds to a single segment $s_{i:j}$
	 and is essentially the \em state persistence \em contextual clique 
	 in the context $c = (e_{i-1:j}=(1,0,..,0,1))$ in the 
	 {\HCRF}'s terminology.
\item \em State transition\em, which is similar to the state transition
	contextual clique in the {\HCRF}s, corresponding to the context $c = (e_t=1)$.
\end{itemize}
Associated with the segmental state clique is 
the potential $R^s_{i:j}$, and with the state transition
is the potential $A_{s',s,t}$, where $s,s' \in S$, and  $S = \{1,2,...,|S|\}$.

A SemiCRF is a three-level {\HCRF}, where the root and bottom are dummy states.
This gives a simplified way to compute
the partition function, ESS, and the MAP assignment using the AIO algorithms.
Thus, techniques developed in this paper for numerical scaling
and partially observed data can be applied to the SemiCRF. 
%We omit
%the details here due to space restriction.
%\rem{
To be more consistent with the literature of flat models such as
HMMs and CRFs, we call the asymmetric inside/outside masses by
the \em forward/backward\em, respectively. Since the model is flat,
we do not need the inside and outside variables. 

%--------------------------
\subsubsection*{Forward}
With some abuse of notation, let 
$\zeta^s_{1:j} = (x_{1:j-1},e_{1:j-1},x_j=s, e_{j} = 1)$.
In other words, there is a segment of state $s$ ending at $j$.
We write the forward $\alpha_t(s)$ as
\begin{eqnarray}
	\label{semiCRF-forward}
	\alpha_j(s) &=& \sum_{\zeta^s_{1:j}}\Phi[\zeta^s_{1:j},z]
\end{eqnarray}

As a result the partition function can be written in term of the
forward as
\begin{eqnarray}
	Z(z) &=& \sum_{\zeta_{1:T}}\Phi[\zeta_{1:T},z] =  \sum_s\sum_{\zeta^s_{1:T}}\Phi[\zeta^s_{1:T},z] \nonumber\\
		&=& \sum_s \alpha_T(s)
\end{eqnarray}

We now derive a recursive relation for the forward. Assume that
the segment ending at $j$ starts somewhere at $i \in [1,j]$.
Then for $i > 1$, there exists the decomposition
$\zeta^s_{1:j}  = (\zeta^{s'}_{1:i-1},x_{i:j}=s,e_{i:j-1}=0)$ for some $s'$, which leads
to the following factorisation
\begin{eqnarray}
	\Phi[\zeta^s_{1:j},z] = \Phi[\zeta^{s'}_{1:i-1}] A_{{s'},s,i-1} R^s_{i:j}
\end{eqnarray}
The transition potential $A_{{s'},s,i-1}$ occurs in the context
$c = (e_{i-1}=1)$, and the segmental potential $R^s_{i:j}$
in the context $c = (x_{i:j}=s,e_{i-1} = 1,e_{i:j-1}=0)$.

For $i=1$, the factorisation reduces to $\Phi[\zeta^s_{1:j},z] = R^s_{1:j}$. Since
we do not know the starting $i$, we must consider all possible values
in the interval $[1,j$. Thus, Equation~\ref{semiCRF-forward} can be rewritten as
\begin{eqnarray}
	\label{semiCRF-forward2}
	\alpha_j(s) &=& \sum_{i \in [2,j]}\sum_{s'}\sum_{\zeta^{s'}_{1:i-1}}\Phi[\zeta^{s'}_{1:i-1}] A_{{s'},s,i-1} R^s_{i:j} + R^s_{1:j}\\
				&=& \sum_{i \in [2,j]}\sum_{s'}\alpha_{i-1}(s') A_{{s'},s,i-1} R^s_{i:j} + R^s_{1:j}
\end{eqnarray}

%--------------------------
\subsubsection*{Backward}
The backward is the `mirrored' version of the forward.
In particular, let
$\underline{\zeta}^s_{j:T} = (x_{j+1:T},e_{j:T},x_j = s, e_{j-1} = 1)$.
and we define the backward $\beta_t(s)$ as
\begin{eqnarray}
	\label{semiCRF-backward}
	\beta_j(s) &=& \sum_{\underline{\zeta}^s_{j:T}}\Phi[\underline{\zeta}^s_{j:T},z]
\end{eqnarray}

Clearly, the partition function can be written in term of the backward as
\begin{eqnarray}
	Z(z) &=&  \sum_s \beta_1(s)
\end{eqnarray}

The recursive relation for the backward
\begin{eqnarray}
	\label{SemiCRF-backward2}
	\beta_i(s) = \sum_{j\in [i,T-1]}\sum_{s'}R^s_{i:j}A_{s,{s'},j}\beta_{j+1}(s')
					+ R^s_{i:T}
\end{eqnarray}

Typically, we want to limit the segment to the maximum length of $L \in [1,T]$. 
This limitation introduces some special cases when 
performing recursive computation of the the forward and backward.
Equation~\ref{semiCRF-forward2} and~\ref{SemiCRF-backward2} are rewritten as follows
\begin{eqnarray}
	\label{SemiCRF:forward}
	\alpha_j(s) &=& \sum_{i\in [j-L+1,j], i > 1}\sum_{s'}\alpha_{i-1}(s')A_{{s'},s,i-1}R^s_{i:j}
													+ R^s_{1:j} \\
	\label{SemiCRF:backward}
	\beta_i(s) &=& \sum_{j\in [i,i+L-1], j < T}\sum_{s'}R^s_{i:j}A_{s,{s'},j}\beta_{j+1}(s')
													+ R^s_{i:T}
\end{eqnarray}

%----------
Since it is a bit clumsy to represent a SemiCRF as a three-level
{\HCRF}, we can extend the {\HCRF} straightforwardly by
allowing the bottom level states to persist. With this relaxation
we have a \emph{nested SemiCRF model} in the sense that each segment
in a Markov chain is also a Markov chain of sub-segments. 

%----------
\subsection{Partially Supervised Learning and Constrained Inference}
Following the intuition in Section~\ref{sec:CAIO}, we require
that all the forward and backward quantities
and the potentials $R^s_{i:j}$ used in 
Equations~\ref{SemiCRF:forward} and~\ref{SemiCRF:backward}
must be \em consistent \em with the labels
in the case of partial supervision and constrained inference.

Specifically, any quantities that are not consistent are set to zero.
Let the labels be $\vartheta = (\widetilde{x},\widetilde{e})$. Then the potential $R^s_{i:j}$
is consistent if it satisfies the following requirements:
\begin{itemize}
\item if there are any labeled states in the interval $[i,j]$,
			they must be $s$, 
\item if there is any labeled ending indicator
			$\widetilde{e}_{i-1}$, then $\widetilde{e}_{i-1} = 1$,
\item if there is any labeled ending indicator $\widetilde{e}_{k}$
			for some  $ k \in [i,j-1]$,	then $\widetilde{e}_{k} = 0$, and
\item if any ending indicator $\widetilde{e}_j$
		is labeled, then $\widetilde{e}_j = 1$.
\end{itemize}

These conditions are captured by
using the following identity function:
\begin{eqnarray}
	\label{SemiCRF:partial-poten}
	\mathbb{I}[R^s_{i:j}] = 
				\delta[\widetilde{x}_{k \in [i,j]} = s]
				\delta[\widetilde{e}_{i-1} =1]
		 		\delta[\widetilde{e}_{k \in [i:j-1]} = 0]
		 		\delta[\widetilde{e}_{j} = 1]
\end{eqnarray}
Notice how these conditions and equation resembles those in 
the Equation~\ref{partial-inside}.
This is because a SemiCRF is just a simplified version of an {\HCRF}
where the potential $R^s_{i:j}$ plays the role of the inside $\Delta^{2,s}_{i:j}$.

Similarly, the forward $\alpha_j(s)$ is consistent if 
the following conditions are satisfied: 
\begin{itemize}
\item if there is a labeled ending indicator at $j$, then $\widetilde{e}_j = 1$, and
\item if there is a labeled state at $j$, then $\widetilde{x}_j = s$.
\end{itemize}
The consistency is captured in the following identity function:
\begin{eqnarray}
	\label{SemiCRF:partial-forward}
	\mathbb{I}[\alpha_j(s)] = 	\delta[\widetilde{e}_j = 1]
								\delta[\widetilde{x}_j = s]
\end{eqnarray}

Furthermore, the backward $\beta_i(s)$ is consistent where:
\begin{itemize}
\item if there is a labeled ending indicator at $i-1$, then $\widetilde{e}_{i-1} = 1$, and
\item if there is a labeled state at $i$ then $\widetilde{x}_i = s$.
\end{itemize}
And again, we have the following identity function
\begin{eqnarray}
	\label{SemiCRF:partial-backward}
	\mathbb{I}[\beta_i(s)] = 	\delta[\widetilde{e}_{i-1} = 1]
								\delta[\widetilde{x}_i = s]
\end{eqnarray}

By installing the consistency identity functions
in Equations~\ref{SemiCRF:partial-poten}, \ref{SemiCRF:partial-forward} and \ref{SemiCRF:partial-backward}
into Equations~\ref{SemiCRF:forward} and~\ref{SemiCRF:backward}, we now arrive at
\begin{eqnarray}
	\label{SemiCRF:partial-forward2}
	\alpha_j(s) = \mathbb{I}[\alpha_j(s)]\left(\sum_{i\in [j-L+1,j], i > 1}
					\sum_{s'}\alpha_{i-1}(s') A_{{s'},s,i-1}\mathbb{I}[R^s_{i:j}]R^s_{i:j} + \mathbb{I}[R^s_{1:j}]R^s_{1:j}\right)
\end{eqnarray}
\begin{eqnarray}
	\label{SemiCRF:partial-backward2}
	\beta_i(s) = \mathbb{I}[\beta_i(s)]\left(\sum_{j\in [i,i+L-1], j < T}
					\sum_{s'}\mathbb{I}[R^s_{i:j}]R^s_{i:j}A_{s,{s'},j}\beta_{j+1}(s') + \mathbb{I}[R^s_{i:j}]R^s_{i:T}\right)
\end{eqnarray}

%----------
\subsection{Numerical Scaling}
We have already shown that a SemiCRF is indeed a 3-level {\HCRF} where
the top and the bottom levels are dummy states, that is, the state size is one and
all the potentials associated with them have a value of one. To apply the scaling
method described in Section~\ref{sec:scaling}, we notice that
\begin{itemize}
\item $\alpha_t(s)$ plays the role of the asymmetric inside mass $\alpha^{1,1}_{1:j}(s)$
\item $\beta_t(s)$ plays the role of the asymmetric outside mass $\lambda^{1,1}_{1:j}(s)$
\end{itemize}

What we do not have here is the explicit notion of inside mass $\Delta^{2,s}_{i:j}$,
but it can be considered as having a value of one. So to apply
the scaling algorithm in Figure~\ref{alg:scaling} we may scale
the state-persistence potential $R^s_{i:j}$ instead. The simplified
version of Figure~\ref{alg:scaling} is given in Figure~\ref{alg:scaling-semiCRF}.

%----------the algorithm
\begin{figure}[htb]
\begin{center}
%\begin{algorithm}
\begin{tabular}{l}\hline
\textbf{Input}:  $T$, the transition potentials and the state-persistence potentials.\\
\textbf{Output}: Scaled quantities: state-persistence potentials, forward/backward.\\
\hline
\textbf{For} $j=1,2,..,T$ \\
\q	\emph{/*Partial scaling*/} \\
\q	\textbf{For} $i=j-L+1,..,j-1$ \\
\q\q	Rescale $R^s_{i:j-1} \leftarrow R^s_{i:j-1}/\prod_{k=i}^{j-1}\kappa_k$\\
\q	\textbf{EndFor}\\
\q	Compute $\alpha_j(s)$ using Equation~\ref{semiCRF-forward} \\
\q	Compute $\kappa_j = \sum_s \alpha_j(s)$ \\
\q	\emph{/*Full scaling*/} \\
\q	Rescale $\alpha_j(s) \leftarrow \alpha_j(s)/\kappa_j$\\
\q	\textbf{For} $i=j-L+1,..,j$ \\
\q\q	Rescale $R^s_{i:j} \leftarrow R^s_{i:j}/\kappa_j$\\
\q	\textbf{EndFor}\\
\textbf{EndFor}\\
Compute true log-partition function using Equation~\ref{Z-scale}.\\
Compute the backward/ESSes using the scaled potentials.\\
\hline
\end{tabular}
\end{center}
\caption{Scaling SemiCRF.} 
\label{alg:scaling-semiCRF}
\end{figure}

Of course, the partial scaling step can be the source of numerical overflow
with $\prod_{k=i}^{j-1}\kappa_k$. The trick here is to realise that
$b/\prod_ka_k = \exp(\log b-\sum_k\log a_k)$ so that we never compute
$b/\prod_ka_k$ directly but the equivalence $\exp(\log b-\sum_k\log a_k)$.
%}

%------
\end{document}